%% file: main.tex
\documentclass[
11pt, 
english, 
singlespacing, 
headsepline, 
]{MastersDoctoralThesis} 

\usepackage[utf8]{inputenc} 
\usepackage[T1]{fontenc} 

\usepackage{mathpazo} 

\usepackage[backend=bibtex,style=authoryear,natbib=true]{biblatex} 

\usepackage[autostyle=true]{csquotes} 

\input{preamble}

\thesistitle{Interpolation between Convolution and Attention via $K$-Nearest Neighbors} 
\supervisor{Dr. Jeová \textsc{Farias} Sales Rocha Neto} 
\examiner{} 
\degree{Bachelor of Arts} 
\author{Mingi Kang} 
\addresses{2141 Smith Union, Brunswick, ME 04011} 

\subject{Computer Science} 
\university{Bowdoin College} 
\department{Department of Computer Science} 
\group{\empty} 
\faculty{\empty} 

\AtBeginDocument{
\hypersetup{pdftitle=\ttitle} 
\hypersetup{pdfauthor=\authorname} 
\hypersetup{pdfkeywords=\keywordnames} 
}

\begin{document}

\frontmatter 

\pagestyle{plain} 




 
 

 

 

\begin{titlepage}
\begin{center}

\vspace*{.3\textheight}

{\Large \ttitle}
 
\vspace{4cm}

\Large An Honors Paper for the \deptname\\[0.3cm] 
by \authorname
 
\vfill

{\Large \univname, \the\year}\\[0.3cm] 
{\Large \copyright \the\year\hspace{1pt} \authorname}
\end{center}
\end{titlepage}


\begin{abstract}
\addchaptertocentry{\abstractname} 
The shift from Convolutional Neural Networks to Transformers has reshaped computer vision, yet these two architectural families are typically viewed as fundamentally distinct. Convolutional Neural Networks are defined by spatially local convolution operations, while Transformers rely on global self-attention. We argue that convolution and self-attention, despite their apparent differences, can be unified within a single $k$-nearest neighbor aggregation framework. The critical insight is that both operations are special cases of neighbor selection and weighted aggregation. Convolution selects neighbors by spatial proximity while self-attention selects by feature similarity, revealing that they lie on a continuous spectrum rather than representing categorically different computations. 

We introduce \textit{Convolutional Nearest Neighbors} (\convnn), a unified framework that formalizes this connection. \convnn\ exactly recovers standard and depthwise convolution by restricting neighbor selection to normalized spatial coordinates, and exactly recovers self-attention and its sparse variants, including KVT-attention, by replacing spatial proximity with scaled dot-product similarity. Beyond these special cases, \convnn\ serves as a drop-in replacement for both convolution and attention layers, enabling systematic exploration of the intermediate spectrum between local and global aggregation through configurable similarity functions, neighbor selection strategies, positional encodings, and aggregation kernels. 

We validate the framework on CIFAR-10, CIFAR-100, and ImageNet-1K image classification tasks across two complementary architectures. First, a hybrid branching layer that combines convolution and \convnn\ branches in VGG and ResNet backbones demonstrates that blending spatial-proximity and feature-similarity selection improves accuracy and reduces loss relatively to purely convolutional baselines. Second, replacing standard self-attention with \convnn\ in Vision Transformers outperforms both full attention and sparse attention variants, suggesting that learnable neighbor selection provides provides implicit regularization through the joint optimization of neighbor indices and aggregation kernel weights during training. Together, these results establish \convnn\ as a principled and interpretable framework that dissolves the apparent boundary between convolution and self-attention, with broader implications for the design of unified vision architectures. 

\end{abstract}


\begin{acknowledgements}
\addchaptertocentry{\acknowledgementname} 

Throughout my four years at Bowdoin College, I had the opportunity to build relationships with colleagues, professors, and friends who have shaped who I am today. This work would not exist without them. 

I would like to thank Professor Abhilasha Kumar for welcoming a first-year student into her Lexicon Lab to work on memory search and retrieval. That experience was my introduction to research, and it set me on the path I have been on ever since. Her early belief in my potential gave me the confidence to pursue not only research in psychology, but eventually to pivot toward computer science and computer vision. 

To my honors thesis advisor, academic mentor, and research father figure, Professor Jeová Farias Sales Rocha Neto, I would not be where I am without your support. From the moment I started research with you during sophomore year, you have pushed me to grow in ways I did not expect. I am genuinely excited to see what we continue to build together even after Bowdoin. Beyond the research and the accomplishments, the personal conversations over lunch about life, work, and everything in between are the moments I will carry with me the longest. 

I am also deeply grateful to Professor Stephen Houser and DJ Merrill for their support with Bowdoin's HPC system and access to NVIDIA GPUs. Professor Houser's introduction to the JetStream2 platform for multi-GPU training proved to be an invaluable resource during the final stages of the experimental work in this thesis. 

To my friends and colleagues at Bowdoin, thank you for making this journey what it was. The conversations between classes, the late nights working on projects, and everything in between made these four years genuinely memorable. I wish you all the very best in wherever life takes you next. 

Finally, to my sister and my mother, thank you for everything. My sister, just a year older than me, has been by my side through every high and every low, and I could not ask for a better person to walk through life with. To have the two of you as my support system has been the greatest gift of my life. 

\end{acknowledgements}


\dedicatory{For my family, friends, faculty, and colleagues who supported my journey\ldots} 

\vfill

\noindent\enquote{\itshape The privilege of being part of Bowdoin College's Computer Science Department is something I will always be grateful for.}\bigbreak

\hfill Mingi Kang


\tableofcontents 



\begin{symbols}{ll}

$x$                                         & scalar variable \\
$\mathbf{x}$                                & vector \\
$X$                                         & matrix \\
$\mathbf{X}$                                & higher-order tensor \\
$\mathbf{X} \in \mathbb{R}^{a \times b \times c}$ & tensor with shape $[a, b, c]$ \\
$\mathbf{1}_k$                              & vector of ones in $\mathbb{R}^k$ \\
\addlinespace

$A[i, :]$                                   & $i$-th row of matrix $A$ \\
$A[I, :]$                                   & rows of $A$ selected by index set $I$ \\
\addlinespace

$\|\mathbf{x}\|_2$                          & $\ell_2$ norm of $\mathbf{x}$ \\
$\operatorname{softmax}(\mathbf{x})_i = \exp(x_i) / \sum_j \exp(x_j)$ & softmax, normalizes $\mathbf{x}$ to a probability distribution \\
$\operatorname{diag}(\mathbf{v})$           & diagonal matrix with entries of $\mathbf{v}$ on its main diagonal \\
\addlinespace

$\mathbf{y} \in \mathbb{R}^{H \times W \times C}$ & 2D input feature map: height $H$, width $W$, channels $C$ \\
$\mathbf{y} \in \mathbb{R}^{N \times C}$    & 1D input: sequence length $L$, channels $C$ \\
$X \in \mathbb{R}^{N \times C}$             & input feature matrix: $N$ feature vectors of dimension $C$ \\
\addlinespace

$R$                                         & spatial kernel size \\
$C'$                                        & number of output channels \\
$W \in \mathbb{R}^{R \times R \times C \times C'}$ & standard convolution kernel (2D) \\
$W_d,\, W_p$                                & depthwise and pointwise kernels \\
\addlinespace

$Q, K, V$                                   & query, key, and value matrices \\
$W^Q, W^K, W^V$                             & query, key, and value projection matrices \\
$d_k$                                       & query/key projection dimension \\
$d_{\text{model}}$                          & model embedding dimension \\
$a_{i,j}$                                   & attention weight between position $i$ and $j$ \\
\addlinespace

$N$                                         & number of feature vectors in the input \\
$K$                                         & number of nearest neighbors \\
$C_{qk}$                                    & projected query/key dimension \\
$C_v$                                       & projected value dimension \\
$S \in \mathbb{R}^{N \times N}$             & pairwise similarity matrix \\
$S_{ij} = \operatorname{sim}(q_i, k_j)$    & similarity score between query $q_i$ and key $k_j$ \\
$\operatorname{sim}(\cdot, \cdot)$          & similarity function (e.g., dot product, cosine) \\
$\rho : \mathbb{R}^K \to \mathbb{R}^K$     & aggregation weight function (e.g., uniform, softmax) \\
$\mathbf{s}_i = \kmax_k(S)[i,:]$           & top-$K$ similarity values for query $\mathbf{x}_i$ \\
$I_i = \argkmax_k(S)[i,:]$                 & top-$K$ neighbor indices for query $\mathbf{x}_i$ \\
$X_{\text{nn},i} \in \mathbb{R}^{K \times C_v}$ & neighborhood matrix for query $\mathbf{x}_i$ \\
$X_{\text{nn}} \in \mathbb{R}^{(KN) \times C_v}$ & concatenation of all neighborhood matrices \\
$Y \in \mathbb{R}^{N \times C'}$            & \convnn\ output feature matrix \\
\addlinespace

$\kmax_k : \mathbb{R}^{N \times N} \to \mathbb{R}^{N \times K}$               & returns top-$K$ values per row \\
$\argkmax_k : \mathbb{R}^{N \times N} \to \{0,\dots,N-1\}^{N \times K}$       & returns column indices of top-$K$ values per row \\
\addlinespace

$\mathbf{v} = (i, j)$                       & 2D spatial coordinate normalized to $[-1,1]^2$ \\
$B \in \mathbb{R}^{d \times 2}$    & frozen random Gaussian matrix for Fourier encoding \\
$\gamma(\mathbf{v}) \in \mathbb{R}^{2d}$    & Fourier feature positional encoding of coordinate $\mathbf{v}$ \\
$\Delta_{ij} = (\Delta x, \Delta y)$        & relative spatial offset between positions $i$ and $j$ \\
$\mathbf{B} \in \mathbb{R}^{(2H-1)\times(2W-1)}$ & learnable relative positional bias table \\

\end{symbols}


\mainmatter 

\pagestyle{thesis} 


\include{Chapters/Chapter0}
\include{Chapters/Chapter1}

\include{Chapters/Chapter2} 
\include{Chapters/Chapter3}
\include{Chapters/Chapter4} 
\include{Chapters/Chapter5} 
\include{Chapters/Chapter6}

\include{Chapters/Chapter7}

\printbibliography[heading=bibintoc]


\end{document}

%% file: preamble.tex
\usepackage{newtxtext}
\usepackage{enumitem} 

\usepackage{float}
\usepackage{listings}
\usepackage{fancybox}

\usepackage{multirow}
\usepackage{algorithm}
\usepackage{algorithmic}
\usepackage{threeparttable}
\usepackage{cuted}
\usepackage{wrapfig}
\usepackage{xspace}
\usepackage{amsmath}
\usepackage{subcaption}
\usepackage{tabularx}
\usepackage{booktabs}
\usepackage{caption}

\usepackage{colortbl}
\usepackage{xcolor}

\usepackage{pgfplots}
\pgfplotsset{compat=1.18} 

\def\convnn{\textsc{ConvNN}\xspace}

\def\argkmax{\text{$k$-argmax}}
\def\kmax{\text{$k$-max}}

\definecolor{vsc-bg}{HTML}{FFFFFF}
\definecolor{vsc-keyword}{HTML}{0000FF}
\definecolor{vsc-identifier}{HTML}{001080}
\definecolor{vsc-string}{HTML}{A31515}
\definecolor{vsc-comment}{HTML}{008000}
\definecolor{vsc-number}{HTML}{098658}
\definecolor{vsc-builtin}{HTML}{795E26}
\definecolor{vsc-frame}{HTML}{E0E0E0}

\lstdefinestyle{vscpythonlight}{
    backgroundcolor=\color{vsc-bg},
    basicstyle=\ttfamily\scriptsize\color{black},
    commentstyle=\itshape\color{vsc-comment},
    keywordstyle=\color{vsc-keyword}\bfseries,
    stringstyle=\color{vsc-string},
    numberstyle=\tiny\color{gray},
    frame=lines,
    rulecolor=\color{black},
    framerule=1pt,
    showstringspaces=false,
    tabsize=4,
    numbersep=6pt,
    breaklines=true,
    language=Python,
    alsoletter={0123456789_}, 
    emph={self,True,False,None}, emphstyle=\color{vsc-builtin},
    emph={[2]torch,nn,F}, emphstyle={[2]\color{vsc-builtin}},
    morekeywords={class,def,return,for,if,elif,else,import,from,with,as,while,try,except,raise,in,is},
    moredelim=[is][\color{vsc-builtin}]{@@}{@@}, 
}

%% file: Chapters/Chapter0.tex

\chapter{Introduction} 

\label{Chapter0} 


\newcommand{\keyword}[1]{\textbf{#1}}
\newcommand{\tabhead}[1]{\textbf{#1}}
\newcommand{\code}[1]{\texttt{#1}}
\newcommand{\file}[1]{\texttt{\bfseries#1}}
\newcommand{\option}[1]{\texttt{\itshape#1}}

\label{sec:intro}
Deep learning architectures for computer vision have undergone a fundamental transformation over the past two decades. Convolutional Neural Networks (CNNs) dominated through pioneering architectures like AlexNet (\cite{DBLP:journals/cacm/KrizhevskySH17}), VGG (\cite{DBLP:journals/corr/SimonyanZ14a}), and ResNet (\cite{he2016deep}). More recently, Vision Transformers (ViT) (\cite{DBLP:conf/iclr/DosovitskiyB0WZ21}) have reshaped the landscape with attention-based models achieving state-of-the-art performance across numerous vision tasks. Yet despite these successes, convolution and attention are typically treated as fundamentally distinct mechanisms suited for different situations. 

Traditionally, convolution operates through local spatial aggregation, where each output is computed from a fixed spatial neighborhood. In contrast, self-attention (\cite{vaswani2017attention}) enables global feature aggregation by computing similarity-based weighted combinations across all spatial positions. However, beneath these surface differences lies a common principle: both operations aggregate features from selected neighbors. The distinction is in neighbor selection--convolution selects by spatial proximity, while attention selects by learned feature similarity. 

Prior work has hinted at this connection. \citet{DBLP:conf/iclr/CordonnierLJ20} demonstrated that attention with relative positional encoding can act like convolution, suggesting these operations may be related. Similarly, Non-Local Neural Networks (\cite{DBLP:conf/cvpr/0004GGH18}) introduced global aggregation mechanisms into CNNs. However, these approaches treat convolution and attention as separate mechanisms to be bridged externally. They do not provide a unified framework that reveals how both operations are instances of the same underlying principle. 

We propose \textit{Convolutional Nearest Neighbors (\convnn)}, a unified framework that formalizes the interconnection between convolution and attention through $k$-nearest neighbor ($k$-NN) selection. Our key insight is that both operations are specialized instances of neighbor aggregation, differing in their selection strategy. Crucially, \convnn can be exactly configured to recover standard convolution and exactly recover standard attention, proving both are points on a continuous spectrum. 

The flexibility of \convnn enables systematic exploration of this spectrum. Rather than choosing between convolution or attention, we can design hybrid configurations that combine spatial and feature-based neighbor selection. Our experiments on CIFAR-10, CIFAR-100, and ImageNet-1K demonstrate that these configurations achieve consistent improvements in accuracy when integrated into both CNN (VGG and ResNet) and Transformer (ViT) architectures. Beyond performance gains, hybrid configurations provide interpretability by explicitly balancing local spatial features with global feature dependencies. This unified framework provides a foundation for designing next-generation operations and architectures that leverage the strengths of both mechanisms. 

\section{Related Works}
\label{sec:related}
\paragraph{Convolution and Attention in Vision}
Convolution has dominated computer vision through architectures like AlexNet (\cite{DBLP:journals/cacm/KrizhevskySH17}), VGG (\cite{DBLP:journals/corr/SimonyanZ14a}), and ResNet (\cite{he2016deep}), providing efficient local feature extraction with spatial inductive biases. Recently, Vision Transformers (ViT) (\cite{DBLP:conf/iclr/DosovitskiyB0WZ21}) and other attention-based models (\cite{DBLP:conf/nips/ParmarRVBLS19, DBLP:conf/iccv/LiuL00W0LG21}) have achieved comparable or superior performance through self-attention (\cite{vaswani2017attention}), which captures long-range dependencies without spatial constraints, but at a larger computational requirement.

Several variants of the self-attention mechanism have been proposed to reduce its quadratic computational and memory complexity. \textit{Local Attention} (\cite{aguilera2024local}) restricts each query to attend only to nearby tokens within a fixed window, thereby exploiting locality while lowering the attention cost from $O(N^2)$ to $O(NW)$, where $W$ is the window size. \textit{Sparse Attention} (\cite{child2019generating}), further generalizes this idea by enforcing structured sparsity patterns in the attention matrix, allowing each token to attend to a predefined subset of positions. \textit{Linear Attention} (\cite{katharopoulos2020transformers}) reduces the quadratic cost of standard attention to $O(N)$ by replacing the softmax with a kernel feature map $\phi$, allowing the key-value produce $\phi(K)^\top V$ to be precomputed independently of the query. These approaches preserve most of the modeling capacity of global attention while making Transformer architectures more scalable for long sequences.

\paragraph{Connections Between Convolution and Attention}
Recent work began exploring the theoretical relationships between convolution and attention. \citet{DBLP:conf/iclr/CordonnierLJ20} provided evidence that self-attention with relative positional encoding can express any convolutional operation, suggesting convolution and attention may exist on a spectrum. \textit{Non-Local Neural Networks} (\cite{DBLP:conf/cvpr/0004GGH18}) and channel-based attention mechanisms (\cite{DBLP:journals/pami/HuSASW20}) introduced global aggregation mechanisms, demonstrating that operations beyond local convolution can enhance feature learning. Furthermore, recent architectures have explicitly combined both operations. \textit{Attention Augmented Convolutional Networks} (\cite{DBLP:conf/iccv/BelloZLVS19}) augment convolutional layers with self-attention. \textit{CoAtNet} (\cite{DBLP:conf/iccv/YuanG0ZYW21}) systematically studies depthwise convolutions followed by attention blocks, showing that this combination achieves strong performance. \textit{ConViT} (\cite{DBLP:conf/icml/dAscoliTLMBS21}) introduces gated positional self-attention that gradually transitions from convolutional-like to attention-like behavior. \textit{Neighborhood Attention} (\cite{DBLP:conf/cvpr/000100LS23}) employs sliding windows to reduce attention's computational cost while preserving spatial locality. While these hybrid architectures demonstrate the benefits of combining local and global operations, they treat convolution and attention as distinct operations combined through architectural composition rather than unified through a common theoretical principle. Notably, \citet{katharopoulos2020transformers} observe that the linear attention formulation is equivalent to a recurrent neural network in the sequential setting. \citet{han2021connection} further showed that local attention in Vision Transformers is structurally equivalent to depthwise convolution under two shared properties of sparse local connectivity and per-channel weight sharing. The key distinction is that local attention uses dynamically predicted weights conditioned on each input instance, whereas depthwise convolution uses fixed spatial weights, with empirical results confirming that the two operations achieve comparable performance when this dynamic weighting is accounted for.

\paragraph{$k$-Nearest Neighbors in Deep Learning}
The $k$-nearest neighbor ($k$-NN) principle has recently reemerged as a valuable design component in neural network architectures. 
\textit{Neural Nearest Neighbors Networks} (\cite{DBLP:conf/nips/Plotz018}) first explored the integration of $k$-NN reasoning primarily for few-shot learning. 
More recently, $k$-NN mechanisms have been incorporated into Transformers to mitigate the quadratic computational and memory costs of self-attention. In particular, \textit{$k$-NN Attention for Boosting Vision Transformers (KVT)} (\cite{DBLP:conf/eccv/WangWWLCLJ22}) enhances Vision Transformers by introducing a $k$-nearest-neighbor attention mechanism that restricts each token’s attention to its most relevant local neighbors rather than the entire sequence. 
Formally, KVT introduces a $k$-NN masking function that constrains each query to attend only to its $K$ most similar keys, according to the traditional scaled dot-product similarity. The selected neighbor values are then normalized via a softmax operation, and the overall attention is computed with a reduced complexity of $O(N K \log K)$ which is lower than the traditional $O(N^2 d)$ of full self-attention, where $N$ denotes the number of tokens and $d$ the embedding dimension. By adaptively selecting the $K$ most similar keys for each query in feature space, KVT effectively balances local and global context modeling, leading to better performance and faster convergence compared to standard ViTs.

Similarly, \textit{Unlimiformer} (\cite{DBLP:conf/nips/Bertsch0NG23}) addresses attention computation complexity by using $k$-NN indices to compute only $k$-NN distances. \textit{Routing Transformers} (\cite{roy-etal-2021-efficient}) rely on $k$-means clustering to model sparse attention matrices, allowing each token to attend to its $k$ closest tokens. Furthermore, in $k$-NN Attention Demystified (\cite{DBLP:conf/iclr/Haris25}), a theoretical framework for $k$-NN attention is presented, reframing self-attention as expectations and leveraging lazy Gumbel sampling for efficient approximation. \textit{Memory-efficient Transformers} (\cite{gupta2021memoryefficient}) also addresses this complexity through a highly accurate approximation for vanilla attention with top-$k$ scores in chunks of queries. These works demonstrate the value of neighbor selection as a design principle, though it treats $k$-NN as an enhancement to existing architectures rather than a unifying framework for understanding different operations.

\section{Proposal}
In this thesis, we propose \textit{Convolutional Nearest Neighbors} (\convnn), a unified framework that formalizes convolution and self-attention as two points on a continuous operational spectrum, connected through the shared principle of neighbor selection and weighted aggregation. Rather than treating these operations as architecturally distinct mechanisms that must be bridged externally, \convnn\ reveals them as special cases of the same underlying formulation, differing only in how neighbors are selected and how their contributions are weighted. 

The \convnn\ framework is organized around three configurable components: similarity computation, neighbor selection and modulation, and weighted aggregation. By varying these components, \convnn\ exactly recovers standard and depthwise convolution by restricting neighbor selection to normalized spatial coordinates with negative exponential Euclidean distance, and exactly recovers self-attention and its sparse variants, including KVT-attention, by replacing spatial proximity with scaled dot-product similarity. Beyond these special cases, \convnn\ enables systematic exploration of the intermediate spectrum, supporting configurable similarity functions, positional encodings, neighbor selection strategies, and aggregated kernels. 

We validate the framework on CIFAR-10, CIFAR-100, and ImageNet-1K image classification across two complementary architectures. In convolutional networks (VGG-11 and ResNet-50), a hybrid branching layer that allocates a fraction $\lambda$ of output channels to a \convnn\ branch and the remainder to a standard convolutional branch consistently outperforms both branches individually. In vision transformers (ViT-Tiny and ViT-Base), replacing standard multi-head self-attention with \convnn-attention surpasses full self-attention, KVT-attention, and other sparse attention variants and achieves the best overall result of 81.64\% top-1 accuracy on ImageNet-1K with a Triton-accelerated implementation. 

The thesis is organized as follows. Chapter~\ref{Chapter1} reviews the background on data representations, model training, Convolutional Neural Networks and convolutions, self-attention, and Transformers. Chapter~\ref{Chapter2} introduces the \convnn\ framework, establishes its connection to convolution and attention, and describes its extensions including positional encodings, similarity functions, sparse candidate search, and hybrid branching. Chapter~\ref{Chapter3} presents the Fast-\convnn\ implementation with custom Triton kernels. Chapters~\ref{Chapter4} and~\ref{Chapter5} report experimental results on CIFAR and ImageNet, respectively. Chapter~\ref{Chapter6} concludes with a summary of findings and directions for future work. 

%% file: Chapters/Chapter1.tex

\chapter{Background} 

\label{Chapter1} 


\section{Introduction}
This chapter establishes the technical background necessary to understand Convolutional Nearest Neighbors (\convnn). We begin with the data representations and model training procedures, family of convolution operations and landmark Convolutional Neural Networks architectures, then turn to self-attention, the Transformer, and its extension to vision. These foundations are revisited and unified in subsequent chapters, where \convnn is developed as a framework that bridges convolutional and attention-based computation.

\section{Tensors and Data Representations}
\begin{figure}[H]
\vspace{10pt}
    \centering
    \includegraphics[width=0.85\linewidth]{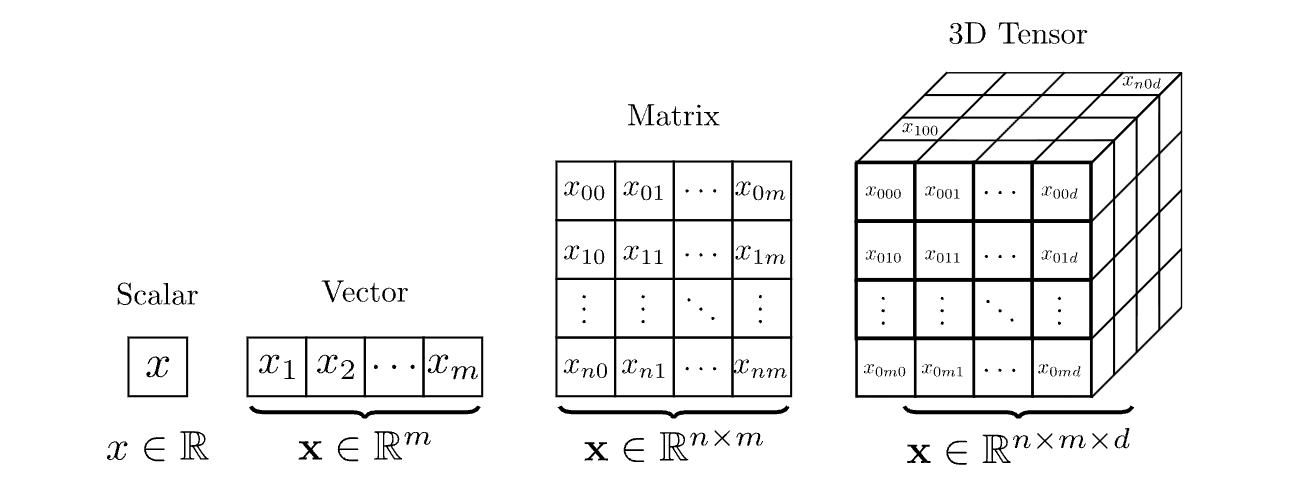}
    \caption{Scalar, vector, matrix, and 3D tensor representations
             used in deep learning (\cite{detorakis2024practical}).}
    \label{fig:tensors}
\end{figure}

Throughout this thesis, inputs, outputs, and model weights are all represented as tensors, a multi-dimensional array of real numbers. Figure~\ref{fig:tensors} illustrates the four levels of this hierarchy. 

A \textit{scalar} $x \in \mathbb{R}$ is a single real-valued number. A \textit{vector} $\mathbf{x} \in \mathbb{R}^m$ is a one-dimensional array of $m$ scalars, used to represent a single feature embedding or signal. A \textit{matrix} $X \in \mathbb{R}^{n \times m}$ is a two-dimensional array with $n$ rows and $m$ columns. In sequence modeling, $n$ denotes the number of sequence elements and $m$ the feature dimension of each element. A \textit{3D tensor} $\mathbf{X} \in \mathbb{R}^{n \times m \times d}$ extends the matrix with a third dimension, used throughout this thesis to represent 2D feature maps with height, width, and channels for the three dimensions in order. 

In image processing, inputs are naturally represented as $\mathbb{R}^{H \times W \times C}$ tensors. For RGB color images $C = 3$, corresponding to the red, green, and blue channels. For grayscale images $C = 1$. In sequence and language modeling, inputs take the matrix form $\mathbb{R}^{N \times C}$, where $N$ is the sequence length and $C$ is the embedding dimension. For tensors of more than three dimensions, such as batched inputs of shape $\mathbb{R}^{B \times H \times W \times C}$, the additional leading dimensions index independent samples in a batch and can be interpreted as a collection of 3D tensor processed in parallel.

\section{Training Deep Learning Models}
\label{sec:optimization}
Deep learning models are parameterized by a set of learnable weights $\theta$, which includes convolutional kernel filters, linear projection matrices, and bias vectors. These weights are not hand-designed but learned automatically from data by minimizing a scalar objective called a \emph{loss function}. The training procedure iteratively adjusts $\theta$ so that the model's predictions become increasingly accurate on the training set, using the gradient of the loss with respect to the weights to determine the direction of each update. 

\subsection{Loss Functions}
\label{subsec:loss_functions}
A loss function $\mathcal{L}(\hat{y}, y)$ measures the discrepancy between the model's prediction $\hat{y}$ and the ground truth label $y$. For a dataset of $n$ training examples, the empirical loss is: 

\begin{equation}
    \mathcal{L}(\theta)
    = \frac{1}{n} \sum_{i=1}^{n} \ell\!\left(f_\theta(x^{(i)}),\; y^{(i)}\right),
\end{equation}

\noindent where $f_\theta(x^{(i)})$ is the model output for input $x^{(i)}$ and $\ell$ is a per-sample loss. For multi-class classification, the standard choice is \emph{cross-entropy loss}, defined for a predicted probability vector $\hat{y} \in \mathbb{R}^C$ over $C$ classes and a one-hot target $y$ as:

\begin{equation}
    \ell_{\text{CE}}(\hat{y}, y)
    = -\sum_{c=1}^{C} y_c \log \hat{y}_c.
\end{equation}

When using data mixing strategies such as MixUp (\cite{zhang2017mixup} or CutMix (\cite{yun2019cutmix}), which produce soft label targets $y \in [0, 1]^C$ rather than one-hot vectors, the same cross-entropy formula applies directly and is referred to as \emph{soft-target cross-entropy}. Label smoothing (\cite{muller2019does} is a related technique that replaces the hard one-hot target with a smoothed distribution, assigning a small probability $\epsilon / C$ to all non-target classes and $1 - \epsilon + \epsilon/C$ to the true class, improving calibration and generalization. 

\subsection{Backpropagation}
\label{subsec:backprop}
Training requires computing the gradient $\nabla_\theta \mathcal{L}$ of the loss with respect to every learnable parameter in the model. Backpropagation (\cite{rumelhart1986learning}) is the algorithm that computes these gradients efficiently by applying the chain rule of calculus through the computational graph of the model in reverse order. For a composed function $f = f_L \circ \cdots \circ f_1$, the gradient of the loss $\mathcal{L}$ with respect to the parameters $\theta_l$ of layer $l$ is: 

\begin{equation}
    \frac{\partial \mathcal{L}}{\partial \theta_l}
    = \frac{\partial \mathcal{L}}{\partial f_L}
      \cdot \frac{\partial f_L}{\partial f_{L-1}}
      \cdots
      \frac{\partial f_{l+1}}{\partial f_l}
      \cdot \frac{\partial f_l}{\partial \theta_l}.
\end{equation}

The forward pass computes and caches the intermediate activations $f_1(x), f_2(f_1(x)), \dots$ required for the backward pass. The backward pass then propagates error signals from the output layer back through each layer in sequence, accumulating gradients for each set of parameters. For deep networks, the repeated multiplication of partial derivatives can cause gradients to shrink exponentially toward zero (\emph{vanishing gradients}) or grow unboundedly (\emph{exploding gradients}), which motivates architectural choices such as residual connections in Section~\ref{subsec:residual_networks} and gradient norm clipping. 

\subsection{Gradient Descent and Optimizers}
\label{subsec:optimizers}
Given the gradient $\nabla_\theta \mathcal{L}$, the model parameters are updated by taking a step in the direction that reduces the loss. The simplest update rule is \emph{stochastic gradient descent} (SGD), which computes the gradients on a randomly sampled mini-batch $\mathcal{B} \subset \{i, \dots, n\}$ of size $B$ and updates: 

\begin{equation}
    \theta \leftarrow \theta
    - \eta \cdot \nabla_\theta
      \frac{1}{|\mathcal{B}|} \sum_{i \in \mathcal{B}}
      \ell\!\left(f_\theta(x^{(i)}),\; y^{(i)}\right),
\end{equation}

\noindent where $\eta > 0$ is the \emph{learning rate}, a scalar hyperparameter that controls the step size. Choosing $\eta$ too large causes the optimization to overshoot and diverge. Too small causes slow convergence. 

Modern deep learning predominantly uses adaptive optimizers that maintain per-parameter learning rates based on gradient history. The \emph{Adam} optimizer (\cite{kingma2014adam}) maintains exponential moving averages of the gradient $m_t$ and its squared magnitude $v_t$: 

\begin{equation}
    m_t = \beta_1 m_{t-1} + (1 - \beta_1)\, g_t, \qquad
    v_t = \beta_2 v_{t-1} + (1 - \beta_2)\, g_t^2,
\end{equation}

\noindent where $g_t = \nabla_\theta \mathcal{L}_t$ is the gradient at step $t$ and $\beta_1, \beta_2 \in [0, 1)$ are decay coefficients, typically set to $0.9$ and $0.999$ respectively. The bias-corrected estimates $\hat{m}_t = m_t / (1 - \beta_1^t)$ and $\hat{v}_t = v_t / (1 - \beta_2^t)$ are used to compute the parameter update: 

\begin{equation}
    \theta_t = \theta_{t-1}
    - \eta \cdot \frac{\hat{m}_t}{\sqrt{\hat{v}_t} + \epsilon}.
\end{equation}

\emph{AdamW} (\cite{loshchilov2017decoupled}) decouples weight decay from the gradient update by applying it directly to the parameters rather than adding an $\ell_2$ penalty to the loss, which improves regularization in practice: 

\begin{equation}
    \theta_t = (1 - \eta \lambda)\,\theta_{t-1}
    - \eta \cdot \frac{\hat{m}_t}{\sqrt{\hat{v}_t} + \epsilon},
\end{equation}

\noindent where $\lambda$ is a weight decay coefficient. AdamW is the optimizer used throughout all experiments in this thesis. 

\subsection{Learning Rate Scheduling}
\label{subsec:lr_schedule}
The learning rate $\eta$ is rarely held constant throughout training. Learning rate scheduling adjusts $\eta$ over the course of training to balance rapid early convergence with fine-grained optimization in later stages. A common strategy is \emph{cosine annealing} (\cite{loshchilov2016sgdr}), which decays the learning rate smoothly from its initial value $\eta_{\max}$ to a minimum $\eta_{\min}$ over $T$ training steps: 

\begin{equation}
    \eta_t = \eta_{\min}
    + \frac{1}{2}(\eta_{\max} - \eta_{\min})
      \left(1 + \cos\!\left(\frac{\pi t}{T}\right)\right).
\end{equation}

In large-scale training, a linear \emph{warmup} phase is typically added before the schedule, during which the learning rate increases linearly from a small initial value to $\eta_{\max}$ over the first $T_{\text{warm}}$ steps, before the cosine decay begins. This prevents instability in the early stages of training when the model parameters are far from optimal and gradient magnitudes can be large.

\section{Convolution Operations}
\label{sec:convolution_operations}
Convolution operations are fundamental building blocks of deep learning in computer vision. Convolutions are parameterized by three quantities: kernel size, stride, and padding value. The convolutional kernel sweeps across the input spatially, computing an elementwise product between the kernel weights and the corresponding input values at each location and summing to produce a single output value per sweep position. 

\begin{figure}[H]
    \vspace{5pt}
    \centering
    \includegraphics[width=0.8\linewidth]{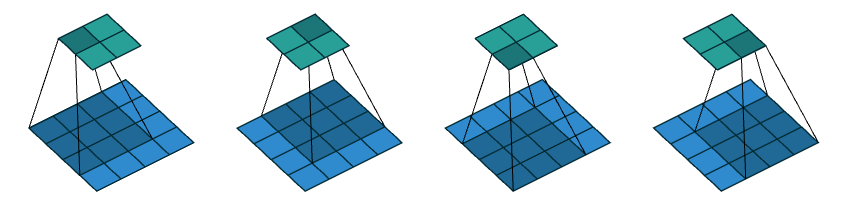}
    \caption{Convolving a $3 \times 3$ kernel over an input with unit
             stride and no padding (\cite{dumoulin2016guide}).}
    \label{fig:conv_basic}
\end{figure}

\begin{figure}[H]
    \vspace{5pt}
    \centering
    \includegraphics[width=1.0\linewidth]{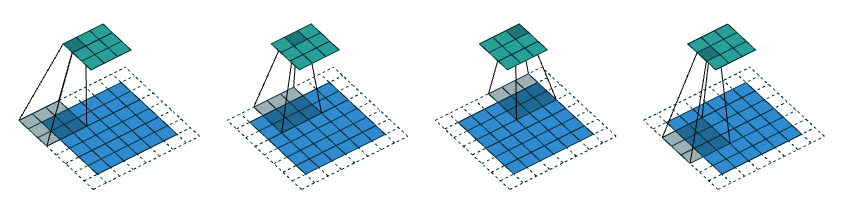}
    \caption{Convolving a $3 \times 3$ kernel over an input with
             $2 \times 2$ stride and $1 \times 1$ border
             padding (\cite{dumoulin2016guide}).}
    \label{fig:conv_stride_pad}
\end{figure}

Figures~\ref{fig:conv_basic} and~\ref{fig:conv_stride_pad} illustrate this sweeping motion in 2 dimensional representation for convolution 2D for a $3 \times 3$ kernel under two configurations: unit stride with no padding, and $2 \times 2$ stride with $1 \times 1$ border padding, respectively. Increasing the stride reduces the spatial resolution of the output, while padding controls whether boundary regions of the input are included in the sweep.

\begin{figure}[H]
    \vspace{10pt}
    \centering
    \includegraphics[width=0.7\linewidth]{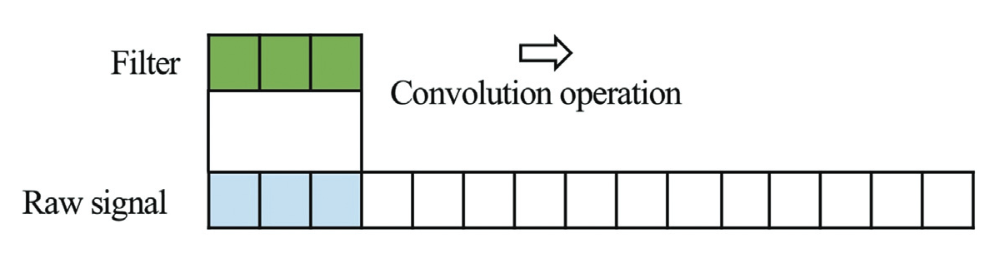}
    \caption{A kernel of size $3$ sliding over a 1D input sequence (\cite{hyeongmin2023}).}
    \label{fig:conv1d}
\end{figure}

Figure~\ref{fig:conv1d} illustrate the analogous operation in one dimension, where the kernel slides along a single spatial axis over an input sequence. The same three parameters: kernel size, stride, and padding, govern the 1D case identically to the 2D case, with the output at each position computed as the dot product between the kernel weights and the corresponding window of input values. 

\begin{figure}[H]
\vspace{10pt}
    \centering
    \includegraphics[width=0.65\linewidth]{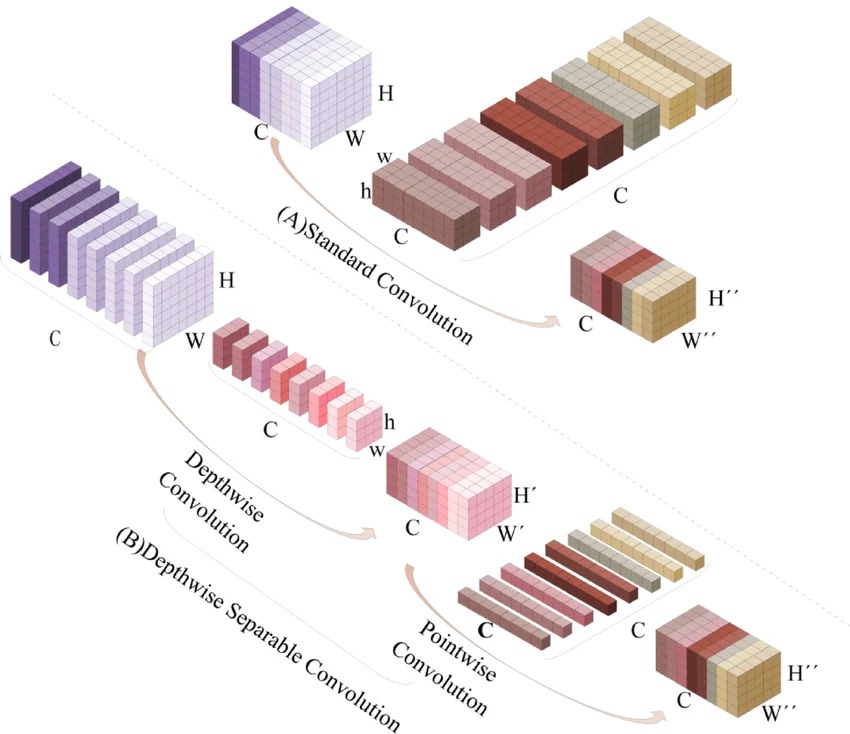}
    \vspace{10pt}
    \caption{Visualizations of standard, depthwise, pointwise, and
             depthwise separable convolutions (\cite{luo2025}).}
    \label{fig:conv_variants}
\end{figure}

We describe four variants of convolution: standard convolution, pointwise convolution, depthwise convolution, and depthwise separable convolution, illustrated in Figure~\ref{fig:conv_variants}. Throughout, we present each operation for both 2D and 1D inputs. For the 2D case, let $\mathbf{y} \in \mathbb{R}^{H \times W \times C}$ denote an input feature map with spatial dimensions $H \times W$ and $C$ channels. For the 1D case, let $\mathbf{y} \in \mathbb{R}^{N \times C}$ denote an input with sequence length $N$ and $C$ channels. In both cases, let $R$ denote the spatial kernel size and $C'$ denote the number of output channels.

\subsection{Standard Convolution}
\label{subsec:std_conv}
Standard convolution applies a learned kernel $W \in \mathbb{R}^{R \times R \times C \times C'}$ over the full spatial and channel extent of the input. For a 2D input: 
\begin{equation}
    \text{Conv}(W, \mathbf{X})_{(i,j,c')}
    = \sum_{p=1}^{R}\sum_{q=1}^{R}\sum_{c=1}^{C}
      W_{p,q,c,c'} \cdot \mathbf{X}_{i+p,\, j+q,\, c}.
\end{equation}

For a 1D input with kernel $W \in \mathbb{R}^{R \times C \times C'}$, the spatial summation reduces to a single dimension: 
\begin{equation}
    \text{Conv}(W, \mathbf{x})_{(i,c')}
    = \sum_{p=1}^{R}\sum_{c=1}^{C}
      W_{p,c,c'} \cdot \mathbf{x}_{i+p,\, c}.
\end{equation}

\subsection{Pointwise Convolution}
\label{subsec:pw_conv}
Pointwise convolution is a special case of standard convolution with a $1\times 1$ spatial kernel. It mixes information across channels at each spatial location independently, leaving the spatial dimensions unchanged. For a 2D input with kernel $W \in \mathbb{R}^{C \times C'}$: 
\begin{equation}
    \text{PointwiseConv}(W, \mathbf{X})_{(i,j,c')}
    = \sum_{c=1}^{C} W_{c,c'} \cdot \mathbf{X}_{i,\, j,\, c}.
\end{equation}

For a 1D input, the operation reduces to a channel-wise linear combination at each position: 
\begin{equation}
    \text{PointwiseConv}(W, \mathbf{x})_{(i,c')}
    = \sum_{c=1}^{C} W_{c,c'} \cdot \mathbf{x}_{i,\, c}.
\end{equation}

\subsection{Depthwise Convolution}
\label{subsec:dw_conv}
Depthwise convolution applies a separate spatial kernel to each input channel independently, with no mixing of channel information. For a 2D input with per-channel kernel $W \in \mathbb{R}^{R \times R \times C}$: 
\begin{equation}
    \text{DepthwiseConv}(W, \mathbf{X})_{(i,j,c)}
    = \sum_{p=1}^{R}\sum_{q=1}^{R}
      W_{p,q,c} \cdot \mathbf{X}_{i+p,\, j+q,\, c}.
\end{equation}

For a 1D input with kernel $W \in \mathbb{R}^{R \times C}$, the per-channel spatial filtering reduces to: 
\begin{equation}
    \text{DepthwiseConv}(W, \mathbf{x})_{(i,c)}
    = \sum_{p=1}^{R}
      W_{p,c} \cdot \mathbf{x}_{i+p,\, c}.
\end{equation}

\subsection{Depthwise Separable Convolution}
\label{subsec:dws_conv}
Depthwise separable convolution factorizes a standard convolution into two sequential steps: a depthwise convolution followed by a pointwise convolution. The same compositional form applies to both 1D and 2D inputs:
\begin{equation}
    \text{DepthwiseSepConv}(W_d, W_p, \mathbf{x})
    = \text{PointwiseConv}\!\bigl(W_p,\;
      \text{DepthwiseConv}(W_d, \mathbf{x})\bigr).
\end{equation}

This factorization decouples spatial feature extraction (depthwise step) from cross-channel mixing (pointwise step), substantially reducing parameter count and computational cost relative to standard convolution.

\section{Convolutional Neural Networks}
Convolutional Neural Networks (CNNs) have been the cornerstone architecture of computer vision for decades. They were first introduced by \citet{lecun-gradientbased-learning-applied-1998}, who demonstrated their effectiveness on handwritten digit classification using the MNIST dataset (\cite{lecun1998mnist}). CNNs leverage local connectivity, weight sharing, and spatial hierarchies to learn increasingly abstract visual representations through successive convolutional layers. 

\section{Historical Convolutional Neural Network Architectures}
Over the past two decades, CNN architectures have evolved substantially in depth, efficiency, and generalization ability. We briefly review two foundational architectures, VGG and ResNet, that shaped the trajectory of modern deep learning for vision. 

\subsection{VGG}
VGG was introduced by Simonyan and Zisserman (\cite{DBLP:journals/corr/SimonyanZ14a}) in 2014 as a systematic study of the effect of network depth on classification accuracy. Building on AlexNet (\cite{krizhevsky_imagenet_2012}), VGG demonstrated that competitive performance could be achieved using an architecture composed exclusively of small $3 \times 3$ convolutional filters, replacing larger kernels with stacks of smaller ones to increase depth while controlling parameter count. 

The architecture follows a regular structure. The input image passes through a sequence of convolutional blocks, each consisting of one or more $3 \times 3$ convolutions with stride 1 and same padding, which preserves the spatial resolution within each block. Each convolutional block is followed by a $2 \times 2$ max pooling operation with stride 2, which halves the spatial dimensions. Applied five times, this progressively reduces the spatial resolution from $224 \times 224$ down to $7 \times 7$: 
\begin{equation*}
    224 \times 224
    \;\to\; 112 \times 112
    \;\to\; 56 \times 56
    \;\to\; 28 \times 28
    \;\to\; 14 \times 14
    \;\to\; 7 \times 7
\end{equation*}
The convolutional blocks are followed by a three fully connected layers: two layers of width 4096 and a final linear layer of width 1000 matching the number of ILSVRC ImageNet classes, followed by a softmax classifier. 

VGG is defined in five configurations, A through E, which differ in the number of convolutional layers within each block. Configuration C additionally introduces $1 \times 1$ convolutions in later blocks. Table~\ref{tab:vgg_configurations} summarizes the block compositions across all five configurations and their corresponding parameter size.

\begin{table}[H]
    \vspace{10pt}
    \scriptsize
    \centering
    \setlength{\tabcolsep}{4pt}
    \resizebox{\linewidth}{!}{%
    \begin{tabular}{lccccc}
        \toprule
        \textbf{Stage} & \textbf{A} & \textbf{B} & \textbf{C} & \textbf{D} & \textbf{E} \\
        \midrule
        conv$_1$ & \texttt{conv3-64}
          & $2\times$\texttt{conv3-64}
          & $2\times$\texttt{conv3-64}
          & $2\times$\texttt{conv3-64}
          & $2\times$\texttt{conv3-64} \\
        conv$_2$ & \texttt{conv3-128}
          & $2\times$\texttt{conv3-128}
          & $2\times$\texttt{conv3-128}
          & $2\times$\texttt{conv3-128}
          & $2\times$\texttt{conv3-128} \\
        conv$_3$ & $2\times$\texttt{conv3-256}
          & $2\times$\texttt{conv3-256}
          & $2\times$\texttt{conv3-256}$+$\texttt{conv1-256}
          & $3\times$\texttt{conv3-256}
          & $4\times$\texttt{conv3-256} \\
        conv$_4$ & $2\times$\texttt{conv3-512}
          & $2\times$\texttt{conv3-512}
          & $2\times$\texttt{conv3-512}$+$\texttt{conv1-512}
          & $3\times$\texttt{conv3-512}
          & $4\times$\texttt{conv3-512} \\
        conv$_5$ & $2\times$\texttt{conv3-512}
          & $2\times$\texttt{conv3-512}
          & $2\times$\texttt{conv3-512}$+$\texttt{conv1-512}
          & $3\times$\texttt{conv3-512}
          & $4\times$\texttt{conv3-512} \\
        \midrule
        linear$_1$ & \multicolumn{5}{c}{\texttt{FCL-4096}} \\
        linear$_2$ & \multicolumn{5}{c}{\texttt{FCL-4096}} \\
        linear$_3$ & \multicolumn{5}{c}{\texttt{FCL-$N_{\text{classes}}$}} \\
        classifier & \multicolumn{5}{c}{\texttt{softmax}} \\
        \midrule
        \textbf{Parameters} & 133M & 133M & 134M & 138M & 144M \\
        \bottomrule
    \end{tabular}}
    \caption{VGG architectural configurations A–E. \texttt{conv$s$-$c$}
             denotes a convolution with kernel size $s\times s$ and $c$ output
             channels. FCL denotes a fully connected linear layer.}
    \label{tab:vgg_configurations}
    \vspace{10pt}
\end{table}

VGG is adopted as the backbone for image classification experiments with small scale datasets in this work. Its exclusive use of $3 \times 3$ convolutions with same padding preserves the spatial resolution of feature maps throughout each convolutional block, with resolution changes occurring only at the max pooling stages. This property is particularly compatible with the \convnn framework, as it ensures that the spatial structure of the feature map remains intact for neighbor search and aggregation.

\subsection{Residual Networks (ResNet)}
\label{subsec:residual_networks}

\begin{figure}[H]
    \centering
    \includegraphics[width=0.8\linewidth]{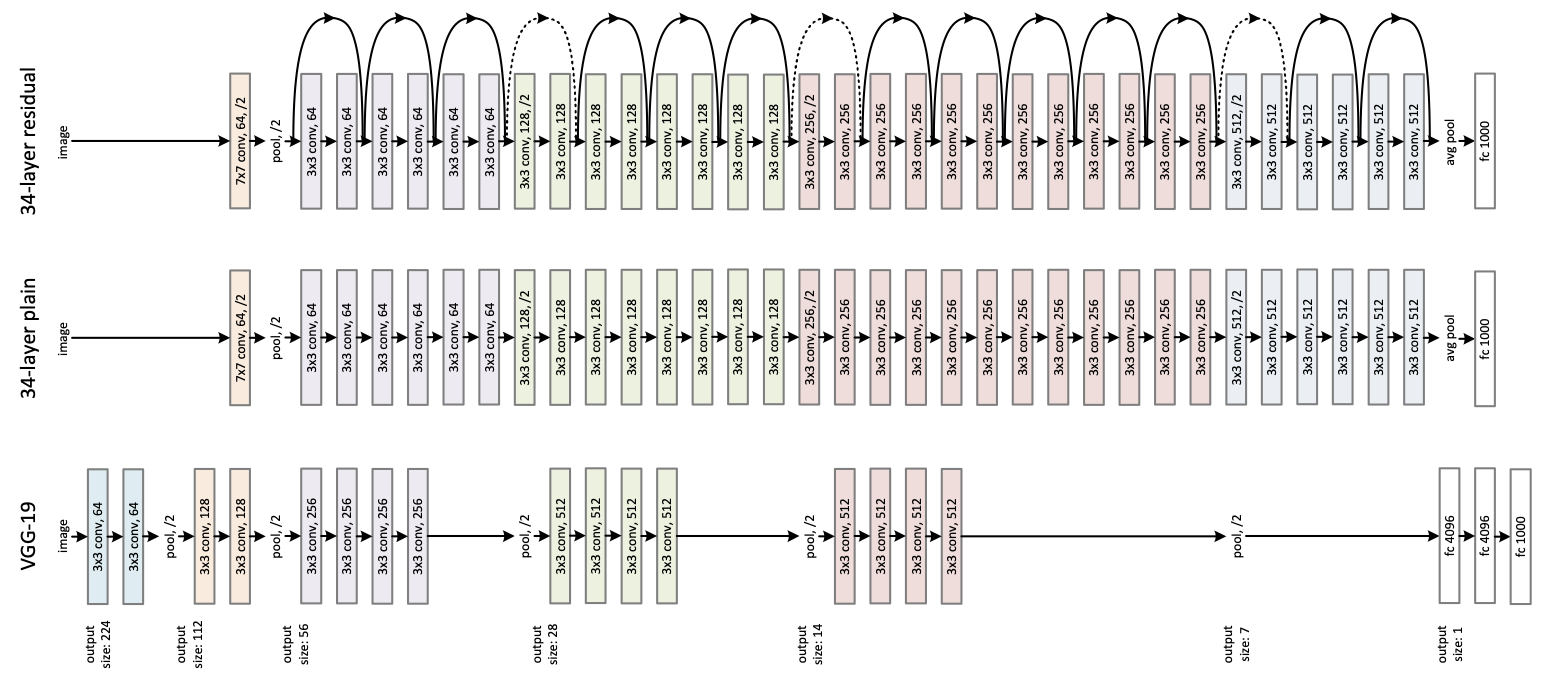}
    \caption{Comparison of VGG-19, a 34-layer plain network, and a 34-layer residual network (\cite{he2016deep}).}
    \label{fig:resnet_model}
\end{figure}

Residual Networks (ResNet) were introduced by \citet{he2016deep} to address the degrading gradient problem observed in very deep networks. As network depth increases, accuracy saturates and degrades rapidly, not due to overfitting, but due to vanishing or exploding gradients during backpropagation. The key insight is to reformulate the learning objective by introducing \emph{shortcut connections} or \emph{residual connections} that bypass one or more layers, allowing gradients to flow directly through the network without passing through every transformations. As illustrated in Figure~\ref{fig:resnet_model}, this contrast directly with purely sequential architectures such as VGG, where each layer receives only the output of the immediately preceding layer. In the plain 34-layer network (middle), layers are stacked sequentially without any shortcuts, producing a deep but unconnected computational graph. The 34-layer residual network (top) introduces curved shortcut connections that bypass pairs of convolutional layers. At stage transitions where the spatial dimensions or channel count change, the shortcut becomes a projection (shown as dashed arrows rather than a plain identity connection). 

\begin{figure}[H]
    \vspace{10pt}
    \centering
    \begin{subfigure}[t]{0.40\linewidth}
        \centering
        \includegraphics[width=\linewidth]{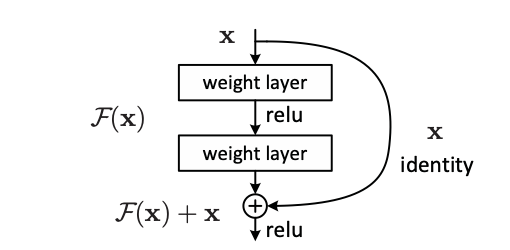}
        \caption{\centering Residual learning building block}
        \label{fig:res-block}
    \end{subfigure}%
    \hfill
    \begin{subfigure}[t]{0.60\linewidth}
        \centering
        \includegraphics[width=\linewidth]{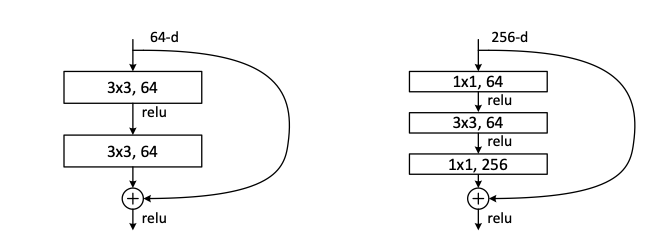}
        \caption{\centering Basic block used in ResNet-18/34 (left) and bottleneck block used in ResNet-50/101/152 (right)}
        \label{fig:resmodel-block}
    \end{subfigure}%
    \vspace{-5pt}
    \caption{ResNet building blocks (\cite{he2016deep}).}
    \label{fig:resnet_blocks}
\end{figure}

As shown in Figure~\ref{fig:res-block}, the input $x$ is added directly to the output of a stack of layers via an identity shortcut. Letting $F(x, \{W_i\})$ denotes the residual mapping learned by the stacked layers, the block output is: 
\begin{equation}
    y = F(x, \{W_i\}) + x.
\end{equation}

When the input and output have different spatial dimensions or channel counts, a linear projection $W_s$ is applied to the shortcut to match dimensions:

\begin{equation}
    y = F(x, \{W_i\}) + W_sx.
\end{equation}

The function $F$ takes one of two forms depending on the model depth, as illustrated in Figure~\ref{fig:resmodel-block}. For shallower networks (ResNet-18/34), $F$ is a basic block consisting of two $3 \times 3$ convolutional layers. For deeper networks (ResNet-50/101/152), $F$ is a bottleneck block consisting of three layers: a $1 \times 1$ convolution that reduces the channel dimension, a $3 \times 3$ convolution, and a $1 \times 1$ convolution that restores the channel dimension. The bottleneck design reduces computational cost while maintaining expressive capacity. 

All ResNet variants share the same top-level structure. The input passes through an initial $7 \times 7$ convolution with 64 channels and stride 2, followed by a $3 \times 3$ max pooling with stride 2, reducing the spatial resolution from $224 \times 224$ to $56 \times 56$. Four residual stages (conv$_{2}$-conv$_{5}$) then progressively halve the spatial dimensions while doubling the channel count, ending with a global average pooling, a fully connected layer of width 1000, and a softmax classifier. The configurations differ only in the number and type of residual blocks per stage, as detailed in Table~\ref{tab:resnet_configurations}.

\begin{table}[H]
    \scriptsize
    \centering
    \setlength{\tabcolsep}{5pt}
    \resizebox{\linewidth}{!}{%
    \begin{tabular}{llccccc}
        \toprule
        \textbf{Stage} & \textbf{Output Size} & \textbf{18} & \textbf{34}
                       & \textbf{50} & \textbf{101} & \textbf{152} \\
        \midrule
        conv$_1$ & $112\times112$
            & \multicolumn{5}{c}{$7\times7$, 64, stride 2} \\
        \midrule
        pool$_1$ & $56\times56$
            & \multicolumn{5}{c}{$3\times3$ max pool, stride 2} \\
        \midrule
        conv$_2$ & $56\times56$
            & $2\times\begin{bmatrix}3\times3,\ 64\\3\times3,\ 64\end{bmatrix}$
            & $3\times\begin{bmatrix}3\times3,\ 64\\3\times3,\ 64\end{bmatrix}$
            & $3\times\begin{bmatrix}1\times1,\ 64\\3\times3,\ 64\\1\times1,\ 256\end{bmatrix}$
            & $3\times\begin{bmatrix}1\times1,\ 64\\3\times3,\ 64\\1\times1,\ 256\end{bmatrix}$
            & $3\times\begin{bmatrix}1\times1,\ 64\\3\times3,\ 64\\1\times1,\ 256\end{bmatrix}$ \\[14pt]
        conv$_3$ & $28\times28$
            & $2\times\begin{bmatrix}3\times3,\ 128\\3\times3,\ 128\end{bmatrix}$
            & $4\times\begin{bmatrix}3\times3,\ 128\\3\times3,\ 128\end{bmatrix}$
            & $4\times\begin{bmatrix}1\times1,\ 128\\3\times3,\ 128\\1\times1,\ 512\end{bmatrix}$
            & $4\times\begin{bmatrix}1\times1,\ 128\\3\times3,\ 128\\1\times1,\ 512\end{bmatrix}$
            & $8\times\begin{bmatrix}1\times1,\ 128\\3\times3,\ 128\\1\times1,\ 512\end{bmatrix}$ \\[14pt]
        conv$_4$ & $14\times14$
            & $2\times\begin{bmatrix}3\times3,\ 256\\3\times3,\ 256\end{bmatrix}$
            & $6\times\begin{bmatrix}3\times3,\ 256\\3\times3,\ 256\end{bmatrix}$
            & $6\times\begin{bmatrix}1\times1,\ 256\\3\times3,\ 256\\1\times1,\ 1024\end{bmatrix}$
            & $23\times\begin{bmatrix}1\times1,\ 256\\3\times3,\ 256\\1\times1,\ 1024\end{bmatrix}$
            & $36\times\begin{bmatrix}1\times1,\ 256\\3\times3,\ 256\\1\times1,\ 1024\end{bmatrix}$ \\[14pt]
        conv$_5$ & $7\times7$
            & $2\times\begin{bmatrix}3\times3,\ 512\\3\times3,\ 512\end{bmatrix}$
            & $3\times\begin{bmatrix}3\times3,\ 512\\3\times3,\ 512\end{bmatrix}$
            & $3\times\begin{bmatrix}1\times1,\ 512\\3\times3,\ 512\\1\times1,\ 2048\end{bmatrix}$
            & $3\times\begin{bmatrix}1\times1,\ 512\\3\times3,\ 512\\1\times1,\ 2048\end{bmatrix}$
            & $3\times\begin{bmatrix}1\times1,\ 512\\3\times3,\ 512\\1\times1,\ 2048\end{bmatrix}$ \\[14pt]
        \midrule
        pool$_2$ & \multicolumn{6}{c}{Global average pool} \\
        linear & \multicolumn{6}{c}{\texttt{FCL-}$N_{\text{classes}}$} \\
        classifier &\multicolumn{6}{c}{\texttt{softmax}} \\
        \midrule
        \textbf{Parameters} & & 11.7M & 21.8M & 25.6M & 44.5M & 60.2M \\
        \bottomrule
    \end{tabular}}
    \caption{ResNet architectural configurations. Each entry denotes the
             number of blocks per stage, type of convolution, and channel values. Basic blocks (ResNet-18/34) consist
             of two $3\times 3$ convolutions; bottleneck blocks
             (ResNet-50/101/152) consist of $1\times 1$, $3\times 3$,
             $1\times 1$ convolutions.}
    \label{tab:resnet_configurations}
    \vspace{10pt}
\end{table}

Residual connections have since become a standard architectural component across virtually all modern deep learning architectures. Beyond ResNet, they are employed in the Transformer (\cite{DBLP:conf/cvpr/000100LS23}) after each attention and FFN sub-layer and in Vision Transformers (\cite{DBLP:conf/iclr/DosovitskiyB0WZ21}) throughout the encoder blocks. By providing a direct gradient path through deep networks, residual connections are necessary for the standard in practice. 

\section{Self-Attention and Variations}
\label{sec:attention}

\subsection{Self-Attention}
\label{subsec:self_attention}
Self-attention was introduced by \citet{vaswani2017attention} for sequence modeling in natural language processing. Prior to its introduction, sequential models such as Recurrent Neural Networks (RNNs) and Convolutional Neural Networks captured dependencies within a local context window, making it difficult to model relationships between distant elements in a sequence. Self-attention addresses this limitation directly by allowing every position in the input to attend to every other position in a single operation, regardless of their distance, enabling the modeling of global dependencies without any locality bias.

Given an input sequence $X \in \mathbb{R}^{N \times d_{\text{model}}}$, queries, keys, and values are obtained via learned linear projections $f_Q, f_K, f_V$: 
\begin{equation}
    Q = f_Q(X) = XW^Q, \quad 
    K = f_K(X) = XW^K, \quad 
    V = f_V(X) = XW^V, 
    \qquad W^Q, W^K, W^V \in \mathbb{R}^{d_{\text{model}} \times d_k}, 
\end{equation}

where $d_{\text{model}}$ is the input feature dimension and $d_k$ is the projected query, key, and value embedding dimension.

The attention output is then computed as: 
\begin{equation}
    \operatorname{Attention}(Q, K, V)
    = \operatorname{softmax}\!\left(\frac{QK^\top}{\sqrt{d_k}}\right)V.
    \label{eq:self_attention}
\end{equation}

The $\operatorname{softmax}$ is applied row-wise to the scaled score matrix $QK^\top / \sqrt{d_k} \in \mathbb{R}^{N \times N}$, converting each row into a probability distribution over positions: 

\begin{equation}
    \operatorname{softmax}(z)_j = \frac{\exp(z_j)}{\sum_{j'=1}^{N} \exp(z_{j'})}.
\end{equation}

This allows each output position $i$ attends to all $N$ positions with non-negative weights that sum to 1. The scaling factor $1/\sqrt{d_k}$ in Equation~\eqref{eq:self_attention} prevents the dot products from growing large in magnitude as $d_k$ increases, which would concentrates the softmax distribution toward a single position and produce very small gradients elsewhere. 

In elementwise form, the query, key, and value vectors for position $i$ are $q_i = W^Q x_i$, $k_j = W^K x_j$, $v_j = W^V x_j$. The attention weight and output at position $i$ are: 
\begin{equation}
    a_{i,j}
    = \frac{\exp(q_i^\top k_j / \sqrt{d_k})}
           {\sum_{j'=1}^{N} \exp(q_i^\top k_{j'} / \sqrt{d_k})},
    \qquad
    y_i = \sum_{j=1}^{N} a_{i,j}\, v_j.
    \label{eq:self_attention_partition}
\end{equation}

\paragraph{Causal (Masked) Self-Attention.}
\label{par:causal_self_attention}
Causal self-attention, also known as masked self-attention, is used in autoregressive sequence modeling where each position $i$ may only attend to positions $j \leq i$, that is to itself and all preceding elements. This constraint prevents the model from accessing future positions during training, ensuring that predictions at position $i$ depend only on the observed context ($x_1, \dots, x_i$). 

The causal attention weight is defined by restricting the softmax normalization to the valid context window: 

\begin{equation}
    a_{i,j}^{\text{causal}}
    = \frac{\exp(q_i^\top k_j / \sqrt{d_k})}
           {\sum_{j'=1}^{i} \exp(q_i^\top k_{j'} / \sqrt{d_k})},
    \qquad j \leq i,
    \label{eq:causal_attention}
\end{equation}

\noindent so that the output at each position aggregates only over preceding and current values: 

\begin{equation}
    y_i = \sum_{j=1}^{i} a_{i,j}^{\text{causal}}\, v_j.
\end{equation}

In practice, causal masking is implemented by adding a mask matrix $M \in \{0, -\infty\}^{N \times N}$ to the pre-softmax attention scores, where $M_{ij} = 0$ if $j \leq i$ and $M_{ij} = -\infty$ otherwise:

\begin{equation}
    \operatorname{CausalAttention}(Q, K, V)
    = \operatorname{softmax}\!\left(
        \frac{QK^\top}{\sqrt{d_k}} + M
      \right) V.
    \label{eq:causal_attention_matrix}
\end{equation}

Since $\exp(-\infty) = 0$, the softmax assigns exactly zero weight to all future positions, effectively zeroing out the upper right triangular entries of the attention matrix. This makes the operation fully autoregressive while remaining parallelizable across all positions during training. Unlike recurrent architectures that must process tokens sequentially, causal self-attention achieves autoregressive behavior without processing each position sequentially.

\subsection{Multi-Head Attention}
\label{subsec:multihead_attention}

Multi-head attention runs $H$ attention operations in parallel over learned subspaces of the projected representations. Formally: 

\begin{align}
    \operatorname{MultiHead}(Q, K, V)
    &= \operatorname{Concat}(\text{head}_1, \dots, \text{head}_H)\,W^O,
    \label{eq:multihead_attention} \\
    \text{head}_h
    &= \operatorname{Attention}(QW_h^Q,\; KW_h^K,\; VW_h^V),
\end{align}

\noindent where $W_h^Q, W_h^K, W_h^V \in \mathbb{R}^{d_\text{model} \times d_k/H}$ are per-head projection matrices and $W^O \in \mathbb{R}^{d_k \times d_{\text{model}}}$ is the output projection. 

In practice, multi-head attention is implemented not by running $H$ independent attention operations sequentially, but by introducing a head dimension directly into the projected tensors. Given $X \in \mathbb{R}^{N \times d_{\text{model}}}$, the full projections $Q, K, V \in \mathbb{R}^{N \times d_k}$ are first computed, then reshaped by partitioning the $d_k$ embedding dimension evenly across $H$ heads, yielding: 

\begin{equation}
    Q_H, K_H, V_H \in \mathbb{R}^{H \times N \times d_k/H},
\end{equation}

\noindent where $d_k$ must be divisible by $H$. Each head $h$ operates on a distinct $d_k/H$-dimensional subspace of the full projection. The attention computation then proceeds in parallel across all heads: 

\begin{equation}
    \operatorname{MultiHead}(Q_H, K_H, V_H)
    = \operatorname{softmax}\!\left(
        \frac{Q_H K_H^\top}{\sqrt{d_k / H}}
      \right) V_H,
    \label{eq:multihead_attention_split}
\end{equation}

\noindent where $\operatorname{softmax}\!\left(Q_H K_H^\top / \sqrt{d_k/H}\right) \in \mathbb{R}^{H \times N \times N}$ and $V_H \in \mathbb{R}^{H \times N \times d_k/H}$, with the output of $\mathbb{R}^{H \times N \times d_k/H}$. The head dimension is then collapsed by concatenating along the last dimension, recovering a tensor of shape $\mathbb{R}^{N \times d_k}$, which is finally projected by $W^O \in \mathbb{R}^{d_k \times d_{\text{model}}}$:

\begin{equation}
    Y = \operatorname{Concat}_H\!\left(
        \operatorname{softmax}\!\left(
            \frac{Q_H K_H^\top}{\sqrt{d_k/H}}
        \right) V_H
    \right) W^O \in \mathbb{R}^{N \times d_{\text{model}}}.
\end{equation}

This formulation is mathematically equivalent to the concatenation of $H$ independent attention heads in Equation~\eqref{eq:multihead_attention}, but is significantly more computationally efficient as it avoids sequential head computation and instead exploits batched matrix multiplication across the head dimension.

\subsection{Linear Attention}
Standard self-attention incurs $O(N^2)$ time and memory complexity in the sequence length $N$, which is prohibitive for long sequences. Linear attention, introduced by \citet{katharopoulos2020transformers}, reduces this to $O(N)$ by replacing the softmax with a kernel-based similarity function. 

The attention operation can be written in a general similarity form as: 

\begin{equation}
    \text{Attention}_i
    = \frac{\displaystyle\sum_{j=1}^{N} \text{sim}(q_i, k_j)\, v_j}
           {\displaystyle\sum_{j=1}^{N} \text{sim}(q_i, k_j)},
\end{equation}

\noindent where the dot-product attention corresponds to $\text{sim}(q, k) = \exp(q^\top k / \sqrt{d_k})$. By replacing $\text{sim}$ with a kernel function $\text{sim}(q, k) = \phi(q)^\top \phi(k)$ for a feature map $\phi(\cdot)$, the attention becomes: 

\begin{equation}
    \text{Attention}_i
    = \frac{\displaystyle\sum_{j=1}^{N} \phi(q_i)^\top \phi(k_j)\, v_j}
           {\displaystyle\sum_{j=1}^{N} \phi(q_i)^\top \phi(k_j)}.
\end{equation}

Factoring out $\phi(q_i)$ via the associativity of matrix multiplication: 

\begin{equation}
    \text{Attention}_i
    = \frac{\phi(q_i)^\top
            \displaystyle\sum_{j=1}^{N} \phi(k_j)\, v_j^\top}
           {\phi(q_i)^\top
            \displaystyle\sum_{j=1}^{N} \phi(k_j)}.
\end{equation}

This is equivalent to reordering the matrix product as: 

\begin{equation}
    \bigl(\phi(Q)\,\phi(K)^\top\bigr)V
    = \phi(Q)\,\bigl(\phi(K)^\top V\bigr),
\end{equation}

\noindent where the right-hand side computes $\phi(K)^\top V \in \mathbb{R}^{d \times d}$ first, reducing the overall complexity from $O(N^2 \cdot d)$ to $O(N \cdot d^2)$, which is linear in $N$. 

\section{Transformer Architecture}
\label{sec:transformers}
The Transformer was introduced by \citet{vaswani2017attention} as the first architecture to rely entirely on self-attention, turning away from recurrence and convolution altogether. Prior to its introduction, sequence modeling for tasks such as machine translation was dominated by encoder-decoder architectures built on recurrent models (\cite{sutskever_sequence_2014}) or convolutional sequence models (\cite{gehring_convolutional_2017}), both of which process input sequentially and struggle to capture long-range dependencies efficiently. The Transformer addresses both limitations through its fully parallel, attention-based design. Beyond natural language processing, the Transformer architecture has since been adapted to a wide range of domains, most notably computer vision, demonstrating its generality as a foundational building block across modalities. 

\subsection{Transformers for Language Modeling}
\label{subsec:transformers}
The original Transformer (\cite{vaswani2017attention}) was proposed for machine translation and follows an encoder-decoder structure. The encoder maps an input sequence $(x_1, \dots, x_N)$ to a sequence of continuous representations, and the decoder generates the output sequence autoregressively where each output token is conditioned on all previously generated tokens and the full encoder output. 

\begin{figure}[H]
    \vspace{10pt}
    \centering
    \includegraphics[width=0.41\linewidth]{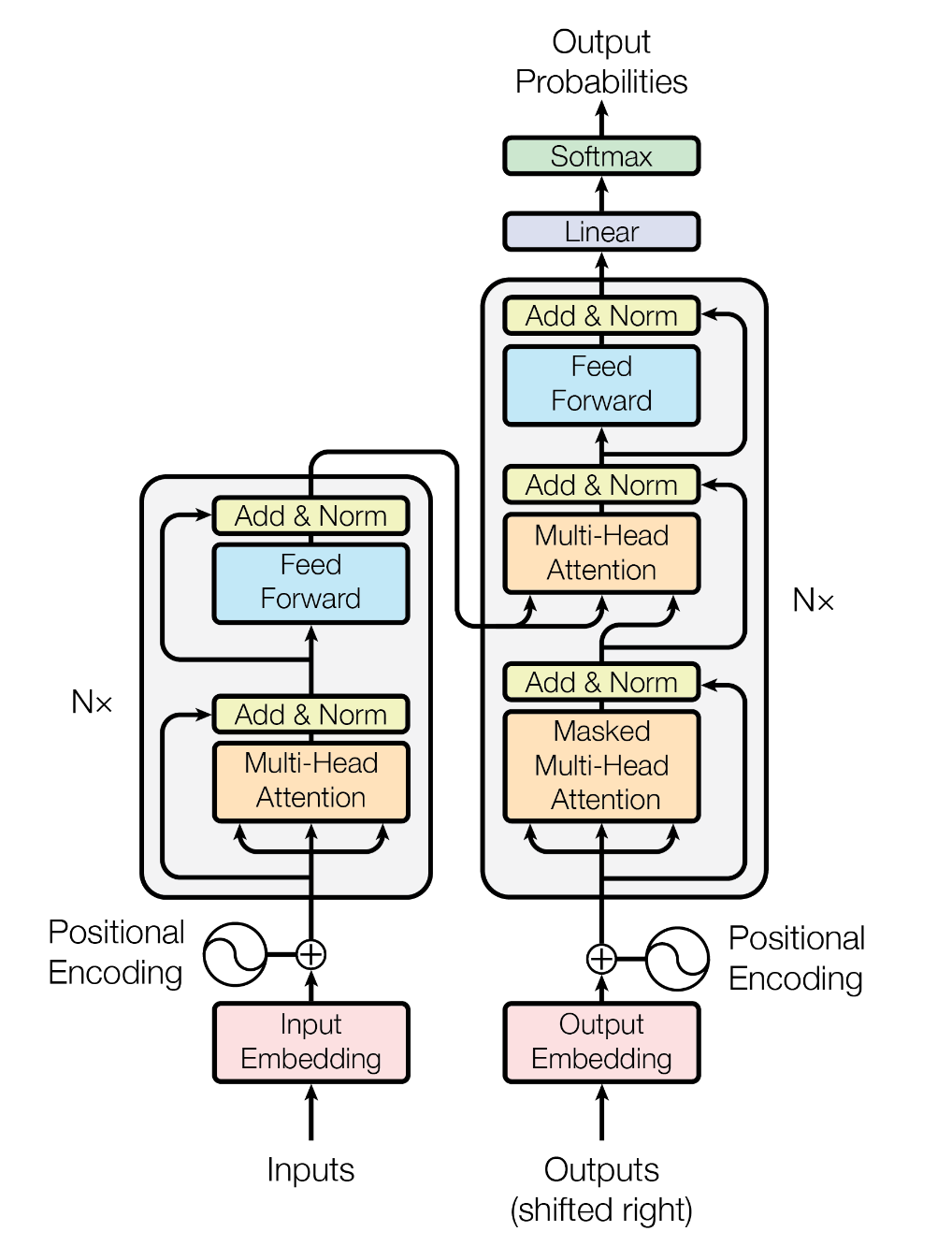}
    \caption{Transformer encoder-decoder architecture (\cite{vaswani2017attention}).}
    \label{fig:transformer_model}
\end{figure}

Figure~\ref{fig:transformer_model} depicts the full encoder-decoder architecture. The encoder stack (left) processes the input sequence through repeated self-attention and feed-forward network (FFN) blocks, producing a sequence of continuous representations. The decoder stack (right) attends to both its own previously generated outputs via masked self-attention and to the encoder representations via cross-attention, enabling autoregressive generation of the output sequence. 

\paragraph{Encoder.}
The encoder consists of $6$ identical blocks, each comprising a multi-head self-attention followed by a position-wise FFN, with residual connections and layer normalization applied after each sub-layer. All sub-layers and embedding layers produce outputs of dimension $d_{\text{model}} = 512$. 

\paragraph{Decoder.}
The decoder also consist of $6$ identical blocks. Each block contains three sub-layers: a masked multi-head self-attention layer, which prevents each position from attending to future positions during autoregressive generation, a cross-attention layer that attends to the encoder output, and a position-wise FFN. The same residual and layer normalization pattern is applied throughout. The final output is produced by a linear projection followed by a softmax over the vocabulary. 

\paragraph{Position-wise Feed-Forward Network.}
The FFN is applied independently to each position in the sequence: 

\begin{equation}
    \operatorname{FFN}(x) = \max(0,\; xW_1 + b_1)W_2 + b_2, 
\end{equation}

\noindent where the inner dimension is $d_{\text{FFN}} = 2048$ and the output dimension matches $d_{\text{model}}$. 

\paragraph{Sinusoidal Positional Encoding.}
Since the Transformer contains no recurrence or convolution, positional information must be injected explicitly. \citet{vaswani2017attention} propose fixed sinusoidal positional encoding, defined for position $pos$ and dimension index $i$ as: 

\begin{equation}
    \text{PE}_{pos,\, 2i}   = \sin\!\left(\frac{pos}{10000^{2i/d_{\text{model}}}}\right),
    \qquad
    \text{PE}_{pos,\, 2i+1} = \cos\!\left(\frac{pos}{10000^{2i/d_{\text{model}}}}\right).
    \label{eq:sinusoidal_pe}
\end{equation}

Even-indexed dimensions use sine and odd-indexed dimensions use cosine. The positional encoding is added element-wise to the input token embeddings before the first encoder or decoder block. 

Following the original Transformer, the architecture has been extended in numerous directions. Encoder-only models such as BERT (\cite{devlin2019bert}) removed the decoder and introduced masked language modeling for bidirectional pretraining, while decoder-only models such as GPT (\cite{radford2019language}) focus on causal autoregressive language modeling, forming the foundation for modern large language models (LLMs). More recent architectures such as T5 (Text-to-Text Transfer Transformer) (\cite{raffel2020exploring}) and BART (\cite{lewis2020bart}) revisited the encoder-decoder design for generalized sequence-to-sequence tasks. 

\subsection{Vision Transformers}
\label{subsec:vision_transformers}
\begin{figure}[H]
    \centering
    \includegraphics[width=1.0\linewidth]{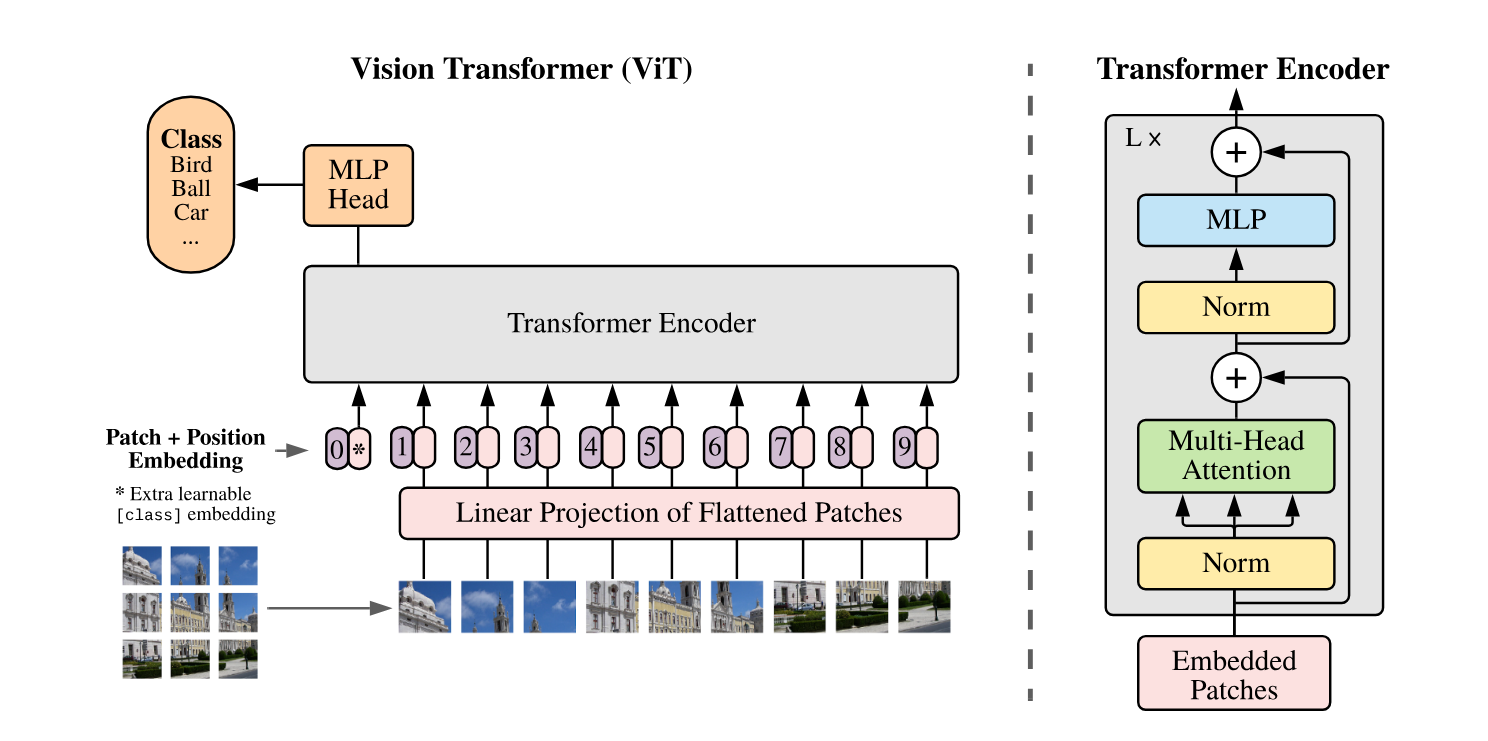}
    \caption{Vision Transformer (ViT) architecture (\cite{DBLP:conf/iclr/DosovitskiyB0WZ21}).}
    \label{fig:vit_model}
\end{figure}

Following the success of Transformer in natural language processing, \citet{DBLP:conf/iclr/DosovitskiyB0WZ21} proposed the Vision Transformer (ViT), which adapts the Transformer encoding directly to image classification by treating an image as a sequence of fixed-sized patches. Figure~\ref{fig:vit_model} shows the full architectural sequence from the input image to the classification output

\paragraph{Patch Embedding.}
Given an input image $\mathbf{X} \in \mathbb{R}^{H \times W \times C}$, the image is divided into non-overlapping patches of size $p \times p$. This is implemented as a single strided convolution with kernel size $p$, stride $p$, input channels $C$, and output channels $d_{\text{model}}$, producing a patch feature map of shape $\mathbb{R}^{H/p \times W/p \times d_{\text{model}}}$. This is then flattened along the spatial dimensions to obtain a sequence of patch embeddings: 

\begin{equation}
    X_{\text{patch}} \in \mathbb{R}^{\frac{HW}{p^2} \times d_{\text{model}}}. 
\end{equation}

\paragraph{Class Token and Positional Encoding.}
Following BERT (\cite{devlin2019bert}), a learnable classification token $\mathbf{x}_{\text{cls}}$ is prepended to the patch sequence, extending the length to $\frac{HW}{p^2} + 1$. A positional encoding is then added element-wise to the full sequence. The original ViT uses learnable 1D positional embeddings; in practice, the sinusoidal encoding of \citet{vaswani2017attention} is also widely adopted, as defined in Equation~\eqref{eq:sinusoidal_pe}. The positional encoding is added after the class token is prepended. 

\paragraph{Transformer Encoder.}
The encoder follows a pre-norm design where each block applies layer normalization before the multi-head self-attention layer, followed by a residual connection and dropout. A second layer normalization precedes the MLP sub-layer, again with a residual connection and dropout. The MLP consists of two linear layers with a GELU (\cite{hendrycks2016gaussian}) activation and a hidden dimension of $4 \times d_{\text{model}}$. 

\paragraph{Classification Head.}
After passing through all encoder blocks, the final state of the class token $\mathbf{x}_{\text{cls}}$ is extracted and passed through a linear layer mapping from $d_{\text{model}}$ to the number of output classes for the final prediction. 

The ViT model configurations are summarized in Table~\ref{tab:vit_configurations}.

\begin{table}[H]
\vspace{10pt}
    \scriptsize
    \centering
    \caption{Vision Transformer (ViT) model configurations. Tiny and Small
             configurations follow the DeiT architecture (\cite{touvron2021training}).}
    \label{tab:vit_configurations}
    \vspace{4pt}
    \setlength{\tabcolsep}{5pt}
    \resizebox{0.65\linewidth}{!}{%
    \begin{tabular}{lcccccc}
        \toprule
        \textbf{Model} & \textbf{Layers} & \textbf{$d_{\text{model}}$}
                       & \textbf{MLP Size} & \textbf{Heads $H$}
                       & \textbf{Patch Size} & \textbf{Params} \\
        \midrule
        ViT-Tiny   & 12 & 192  & 768  & 3  & $16 \times 16$ & 5.7M  \\
        ViT-Small  & 12 & 384  & 1536 & 6  & $16 \times 16$ & 22M   \\
        \midrule
        ViT-Base   & 12 & 768  & 3072 & 12 & $16 \times 16$ & 86M   \\
        ViT-Large  & 24 & 1024 & 4096 & 16 & $16 \times 16$ & 307M  \\
        ViT-Huge   & 32 & 1280 & 5120 & 16 & $14 \times 14$ & 632M  \\
        \bottomrule
    \end{tabular}}
    \vspace{10pt}
\end{table}

The ViT architecture has been extended and refined in several directions. DeiT (\cite{touvron2021training}) introduced data-efficient training strategies and a distillation token, enabling ViT-scale models to train competitively on ImageNet-1K without large-scale pretraining data. Swin Transformer (\cite{DBLP:conf/iccv/LiuL00W0LG21}) introduced hierarchical feature maps and shifted window attention, recovering the multi-scale inductive biases of CNNs while retaining global attention capability, and has become a standard backbone for dense prediction tasks. More recently, the core Transformer architecture has continued to prove its versatility in computer vision beyond classification. For image segmentation, SegFormer (\cite{xie_segformer_2021}) proposed a lightweight hierarchical Transformer encoder paired with a simple MLP decoder, achieving strong performance with reduced computational cost. DiT (Diffusion Transformer) (\cite{peebles2023scalable}) applied the Transformer backbone to diffusion-based image generation, replacing the conventional U-Net denoising architecture and demonstrating strong scalability with model size. Together, these works establish the Transformer as a general-purpose architecture competitive across recognition, segmentation, and generative tasks in computer vision.

%% file: Chapters/Chapter2.tex

\chapter{Convolutional Nearest Neighbors} 

\label{Chapter2} 

\section{Notation}
\label{sec:notation}

Scalar variables, vectors, and matrices are denoted in plain lowercase (e.g., $x$), bold lowercase (e.g., $\mathbf{x}$, and plain uppercase (e.g., $X$), respectively. High-order tensors are denoted by bold uppercase letters (e.g., $\mathbf{X}$). The shape of a tensor is written as $\mathbf{X} \in \mathbb{R}^{a \times b \times c}$, abbreviated as $[a, b, c]$ where convenient. A vector of ones in $\mathbb{R}^k$ is denoted as $\mathbf{1}_k$. 

We adopt Python-style indexing, where square brackets indicate row and column selection. Given a matrix $A \in \mathbb{R}^{n \times m}$, $A[i, :]$ selects the $i$-th row of $A$. Given an index sets $I \subset \{0, \dots, n-1\}$, $A[I, :]$ selects the rows of $A$ corresponding the indices in $I$. 

A vector $\mathbf{x}$ is $\ell_2$-normalized when $\|\mathbf{x}\|_2 = 1$. The operator $\operatorname{softmax}(\mathbf{x})$ normalizes $\mathbf{x}$ into a probability distribution with $\operatorname{softmax}(\mathbf{x})_i = \exp(x_i) / \sum_j \exp(x_j)$. The operator $\operatorname{diag}(\mathbf{v})$ denotes the diagonal matrix with entries of $\mathbf{v}$ on its main diagonal. 

\subsection{$k$-Nearest Neighbors}
\label{subsec:knn_notation}
Let $X \in \mathbb{R}^{N \times d}$ be a matrix whose $i$-th row $\mathbf{x}_i \in \mathbb{R}^d$ is the $i$-th feature vector, for $i = 1, \dots, N$. 

We make use of two operators throughout: $\kmax$ and $\argkmax$. The operator $\kmax_k$ returns the $K$ largest values in each row of a matrix: 

\begin{equation}
    \kmax_k : \mathbb{R}^{N \times N} \to \mathbb{R}^{N \times K}.
\end{equation}

The operator $\argkmax_k$ returns the column indices of these $K$ largest values for each row: 

\begin{equation}
    \argkmax_k : \mathbb{R}^{N \times N} \to \{0, \dots, N-1\}^{N \times K}.
\end{equation}

Given a similarity matrix $S \in \mathbb{R}^{N \times N}$, the global index matrix and top-$K$ similarity matrix are defined as: 

\begin{equation}
    \label{eq:knn}
    I = \argkmax_k(S) \in \{0, \dots, N-1\}^{N \times K}, \qquad
    T = \kmax_k(S) \in \mathbb{R}^{N \times K},
\end{equation}

\noindent where $I_i = I[i, :]$ and $T_i = T[i, :]$ denote the neighbor indices and similarity values for query $\mathbf{x}_i$, respectively. The set of $K$-nearest neighbors of $\mathbf{x}_i$ is then retrieved as the submatrix $X[I_i, :] \in \mathbb{R}^{K \times d}$.

\section{Core Framework}
\label{sec:core_framework}
We now present \convnn, an operational framework that generalizes both convolution and self-attention, is built upon three core steps: \textit{similarity computation}, \textit{neighbor selection and modulation}, and \textit{weighted aggregation}. 

\begin{figure}[h]
    \centering
    \includegraphics[width=1.0\linewidth]{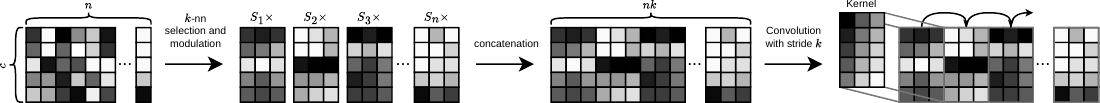}
    \caption{The \convnn operation}
    \label{fig:convnn_diagram}
\end{figure}

Let $X \in \mathbb{R}^{N \times C}$ denote the input feature matrix, where $N$ is the number of feature vectors and $C$ is the channel dimension. For inputs with spatial structure of shape $[H, W, C]$, \convnn\ first flattens the spatial dimensions to obtain a matrix of shape $[HW, C]$, applies the procedure described below, and reshapes the result back to $[H, W, C']$. An overview of the \convnn\ operation is illustrated in Figure~\ref{fig:convnn_diagram}.

\paragraph{1. Similarity Computation.}
Let $f_Q : \mathbb{R}^{N \times C} \to \mathbb{R}^{N \times C_{qk}}$ and $f_K : \mathbb{R}^{N \times C} \to \mathbb{R}^{N \times C_{qk}}$ be projection functions, and define $Q = f_Q(X)$ and $K = f_K(X)$. Similarities are computed in the projected space $\mathbb{R}^{C_{qk}}$ rather than the original feature space $\mathbb{R}^C$, via matrix multiplication: 

\begin{equation}
    S = QK^\top \in \mathbb{R}^{N \times N},
    \label{eq:sim_mat}
\end{equation}

where $S_{ij} = \text{sim}(q_i, k_j)$ is the similarity score between query $q_i$ and key $k_j$. The similarity function $\text{sim}(\cdot, \cdot)$ defaults to the dot product but can be replaced by any of the metrics described in Section~\ref{subsec:convnn_sim_extension}, including cosine similarity, euclidean distance, and others. 

\paragraph{2. Neighbor Selection and Modulation.}
Let $f_V : \mathbb{R}^{N \times C} \to \mathbb{R}^{N \times C_v}$ be a projection function, and define $V = f_V(X)$. Using the operators defined in Section~\ref{subsec:knn_notation}, we compute the top-$K$ similarity values and their corresponding indices for each query: 

\begin{equation}
    \mathbf{s}_i = T[i, :] = \kmax_k(S)[i, :] \in \mathbb{R}^{K}, 
\end{equation}

\begin{equation}
    I_i = I[i, :] = \argkmax_k(S)[i, :] \in \{0, \dots, N-1\}^{K}.
\end{equation}

We introduce a weighting function $\rho : \mathbb{R}^K \to \mathbb{R}^K$ that maps raw similarity scores to aggregation weights, instantiated either as uniform weighting ($\rho(\mathbf{z}) = \mathbf{1}_K$) or softmax normalization ($\rho(\mathbf{z}) = \operatorname{softmax}(\mathbf{z})$). Letting $S_i = \operatorname{diag}(\rho(\mathbf{s}_i))$, the \textit{neighborhood matrix} for $\mathbf{x}_i$ is defined as: 

\begin{equation}
    X_{\text{nn}, i} = S_i\, V[I_i, :] \in \mathbb{R}^{K \times C_v},
    \label{eq:neigh_mat}
\end{equation}

\noindent which collects the $K$ projected neighbors of $\mathbf{x}_i$, each scaled by its corresponding weight in $\rho(\mathbf{s}_i)$ and ordered by similarity. 

\paragraph{3. Weighted Aggregation.}
Let $X_{\text{nn}} \in \mathbb{R}^{(KN) \times C_v}$ denote the concatenation of all neighborhood matrices $\{X_{\text{nn},i}\}_{i=1}^N$ along the row dimension. \convnn\ applies a \textsc{Conv1D} operation (standard or depthwise) with $C'$ output channels, kernel size $K$, and stride $K$ to $X_{\text{nn}}$, yielding:

\begin{equation}
    Y = \text{Conv1D}(X_{\text{nn}}) \in \mathbb{R}^{N \times C'}.
\end{equation}

\subsection{PyTorch Implementation of \convnn\ Operator}
The PyTorch (\cite{pytorch2025}) implementation below (Code~\ref{code:convnn_basic}) follows the three-step \convnn\ framework illustrated in Figure~\ref{fig:convnn_diagram} directly. Throughout the code, tensor shapes are commented using bracket notation $[a, b, c]$, where each entry corresponds to a dimension of the tensor in order (batch, sequence, channel, etc.), matching the shorthand introduced in Section~\ref{sec:notation}. 

\textbf{Step 1} implements the similarity computation of Equation~\eqref{eq:sim_mat}. The projections \texttt{f\_Q}, \texttt{f\_K}, and \texttt{f\_V} are \texttt{nn.Linear} layers corresponding to $f_Q$, $f_K$, and $f_V$ respectively, and the \texttt{similarity} method computes $S = QK^\top \in \mathbb{R}^{N \times N}$ via a batched matrix multiplication.

\textbf{Step 2} implements neighbor selection and modulation. \texttt{torch.topk} corresponds to $\kmax_k$ and $\argkmax_k$: given the similarity matrix $S$. It returns the $K$ largest scores \texttt{s} $= T_i$ and their column indices \texttt{I} $= I_i$ for each query position simultaneously, as defined in Equation~\eqref{eq:knn}. Softmax is then applied to \texttt{s} to obtain the weights $\rho(\mathbf{s}_i)$. To gather the corresponding value vectors, \texttt{V} is first transposed to a channels-first layout $[B, C_v, N]$ and expanded to $[B, C_v, N, K]$ so that \texttt{torch.gather} can index along the sequence dimension (dim=2) using the neighbor index tensor \texttt{I\_exp}. \texttt{torch.gather} implements the submatrix selection $V[I_i, :]$ from Equation~\eqref{eq:neigh_mat}: for each query position and each channel, it retrieves the $K$ value vectors at the neighbor positions specified by \texttt{I\_exp}, producing $X_{\text{nn}, i} \in \mathbb{R}^{K \times C_v}$ for all queries simultaneously. 

\textbf{Step 3} implements the weighted aggregation. The neighborhood tensor \texttt{X\_nn} of shape $[B, C_v, N, K]$ is transposed and reshaped to $[B, C_v, N \cdot K]$, interleaving the $K$ neighbors of each position contiguously in memory. A \texttt{nn.Conv1d} with \texttt{kernel\_size=K} and \texttt{stride=K} then processes each $K$-length block independently, which is exactly equivalent to applying the Conv1D aggregation of Step 3 to each neighborhood matrix $X_{\text{nn}, i}$ in turn. 

\vspace{10pt}
\begin{lstlisting}[
    basicstyle=\footnotesize\ttfamily,
    caption={PyTorch implementation of the \convnn\ operation.}, 
    captionpos=b,
    label={code:convnn_basic}
]
class ConvNN(nn.Module):
    def __init__(self, C, C_qk, C_v, C_out, K):
        """
        Args:
            C     : input channel dimension
            C_qk  : projected query/key dimension (C_{qk})
            C_v   : projected value dimension
            C_out : output channel dimension (C')
            K     : number of nearest neighbors
        """
        super(ConvNN, self).__init__()
        self.K    = K
        self.f_Q  = nn.Linear(C, C_qk)
        self.f_K  = nn.Linear(C, C_qk)
        self.f_V  = nn.Linear(C, C_v)
        self.rho = nn.Softmax(dim=-1)
        self.conv = nn.Conv1d(
            in_channels=C_v,
            out_channels=C_out,
            kernel_size=K,
            stride=K
        )

    def similarity(self, Q, K):
        # S = Q K^T in R^{N x N}
        return Q @ K.transpose(-2, -1)

    def forward(self, X):
        # X: [B, N, C]
        B, N, C = X.shape

        # Step 1 - Similarity Computation
        # Q, K: [B, N, C_qk],   V: [B, N, C_v]
        Q, K, V = self.f_Q(X), self.f_K(X), self.f_V(X)
        S = self.similarity(Q, K)               # S: [B, N, N]

        # Step 2 - Neighbor Selection and Modulation
        # s_i = kmax_k(S)[i, :],  I_i = argkmax_k(S)[i, :]
        s, I = torch.topk(S, k=self.K, dim=-1)  # s, I: [B, N, K]

        # rho(s_i): weighting function - softmax or uniform
        s = self.rho(s)                         # [B, N, K]

        # Gather V[I_i, :] in R^{K x C_v}
        # V: [B, N, C_v] -> transpose to [B, C_v, N]
        # then expand to [B, C_v, N, K] for gather along dim=2
        C_v   = V.shape[-1]
        V_t   = V.permute(0, 2, 1)                           # [B, C_v, N]
        V_exp = V_t.unsqueeze(-1).expand(B, C_v, N, self.K)  # [B, C_v, N, K]
        I_exp = I.unsqueeze(1).expand(B, C_v, N, self.K)     # [B, C_v, N, K]
        V_nn  = torch.gather(V_exp, dim=2, index=I_exp)      # [B, C_v, N, K]

        # X_nn_i = diag(rho(s_i)) * V[I_i, :]
        s_exp = s.unsqueeze(1).expand(B, C_v, N, self.K)     # [B, C_v, N, K]
        X_nn  = s_exp * V_nn                                 # [B, C_v, N, K]

        # Step 3 - Weighted Aggregation
        # Reshape to [B, C_v, N*K] for Conv1D with kernel_size=K, stride=K
        X_nn = X_nn.permute(0, 1, 3, 2).reshape(B, C_v, -1)  # # X_nn: [B, C_v, N*K]
        Y    = self.conv(X_nn)                               # Y: [B, C', N]
        Y    = Y.permute(0, 2, 1)                            # Y: [B, N, C']

        return Y
\end{lstlisting}

\section{Extensions}
To enhance the expressiveness of \convnn, we introduce couple extensions that enable it to interpolate between purely feature-based and spatially-aware neighbor selection.

\subsection{Similarity Computation}
\label{subsec:convnn_sim_extension}
The similarity computation step is the first stage of the \convnn framework. Given a query vector $q$ and a candidate key vector $k$, a similarity function $\text{sim}(q, k)$ scores how closely $k$ matches $q$. This score drives neighbor selection in the subsequent step. The framework is agnostic to the choice of similarity metric; we survey the principal options below. Note that distance-based metrics $D(q, k)$ can be converted to similarities by negating and exponentiating (e.g., $\text{sim}(q,k) = \exp(-D(q, k))$) before neighbor selection.

\subsubsection{Dot-Product Similarity}
The dot-product similarity is the standard scoring function used in scaled dot-product attention (\cite{vaswani2017attention}): 

\begin{equation}
    \text{sim}(q, k) = q^\top k.
\end{equation}

It measures the projection of $k$ onto $q$, rewarding vectors that are large in magnitude and well-aligned in direction. In self-attention, this is scaled by $\frac{1}{\sqrt{d_k}}$ to stabilize gradients in high dimensions, yielding the scaled dot-product attention matrix. The dot product is computationally efficient and GPU-optimized, making it the default choice in most attention-based architectures.  

\vspace{10pt}
\begin{lstlisting}[
nolol, 
basicstyle=\footnotesize\ttfamily,
caption={Dot-product similarity. \texttt{scale} defaults to $d_k$,
         recovering scaled dot-product attention.}, 
label={code:dot_product_sim}, 
captionpos=b
]
def dot_product_similarity(Q, K, scale=None):
    # Q, K: [B, N, d_k]
    # Returns S: [B, N, N]
    if scale is None:
        scale = Q.shape[-1]
    return torch.matmul(Q, K.transpose(-2, -1)) / np.sqrt(scale)
\end{lstlisting}
\vspace{-5pt}

\subsubsection{Cosine Similarity}
Cosine similarity normalizes the dot product by the magnitude of both vectors, making it invariant to their scale:

\begin{equation}
    \text{sim}(q, k)
    = \frac{q^\top k}{\|q\|\,\|k\|}
    = \frac{q^\top k}{\sqrt{q^\top q}\,\sqrt{k^\top k}}.
\end{equation}

It measures the angle between $q$ and $k$, taking values in $[-1, 1]$. Cosine similarity is useful when the feature vectors matter more than their magnitudes, and it is commonly used in retrieval and metric learning settings. 

\vspace{10pt}
\begin{lstlisting}[
nolol, 
basicstyle=\footnotesize\ttfamily,
caption={Cosine similarity. Requires \texttt{import torch.nn.functional as F}.}, 
label={code:cosine_sim}, 
captionpos=b
]
def cosine_similarity(Q, K):
    # Q, K: [B, N, d_k]
    # Returns S: [B, N, N]
    Q_norm = F.normalize(Q, p=2, dim=-1)   # L2-normalize along feature dim
    K_norm = F.normalize(K, p=2, dim=-1)
    return torch.matmul(Q_norm, K_norm.transpose(-2, -1))
\end{lstlisting}
\vspace{-5pt}

\subsubsection{Euclidean Distance}
Euclidean distance, the $L_2$ norm of the difference vector, measures straight-line distance in feature space:

\begin{equation}
    D(q, k) = \|q - k\|_2
    = \sqrt{\sum_{i=1}^{d}(q_i - k_i)^2} = \|q\|_2.
\end{equation}

Smaller values indicate greater similarity. As a distance metric, it is converted to a similarity by negation and exponentiation before neighbor selection. Euclidean distance is sensitive to the scale of individual dimensions, so feature normalization is often applied beforehand. 

\vspace{10pt}
\begin{lstlisting}[nolol, basicstyle=\footnotesize\ttfamily,
caption={Squared Euclidean distance.}, 
label={code:euclidean_dist}, 
captionpos=b
]
def euclidean_distance(Q, K):
    # Q, K: [B, N, d_k]
    # Returns D: [B, N, N]  (squared Euclidean distance)
    # Uses the identity ||q - k||^2 = ||q||^2 + ||k||^2 - 2 q^T k
    Q_sq = torch.sum(Q ** 2, dim=-1, keepdim=True)       # [B, N, 1]
    K_sq = torch.sum(K ** 2, dim=-1, keepdim=True)       # [B, N, 1]
    dot  = torch.matmul(Q, K.transpose(-2, -1))          # [B, N, N]
    return Q_sq + K_sq.transpose(-2, -1) - 2 * dot
\end{lstlisting}

When $q$ and $k$ are $\ell_2$-normalized, the squared Euclidean distance simplifies to: 

\begin{equation}
    \|q - k\|_2^2
    = \|q\|_2^2 + \|k\|_2^2 - 2q^\top k
    = 2(1 - q^\top k),
\end{equation}

\noindent since $\|q\|_2 = \|k\|_2 = 1$. The cosine similarity between normalized vectors is $\text{sim}(q, k) = q^\top k$, so minimizing squared Euclidean distance is exactly equivalent to maximizing cosine similarity. This means that for $\ell_2$-normalized features, two metrics induce the same neighbor ordering and are therefore interchangeable within the \convnn framework. 

\subsubsection{Manhattan Distance}
Manhattan distance, the $L_1$ norm of the difference vector, sums the absolute per-dimension deviations:

\begin{equation}
    D(q, k) = \|q - k\|_1
    = \sum_{i=1}^{d} |q_i - k_i|.
\end{equation}

Compared to Euclidean distance, the $L_1$ norm is more robust to outliers in individual dimensions, as large deviations are not amplified by squaring. 

\vspace{10pt}
\begin{lstlisting}[
nolol, 
basicstyle=\footnotesize\ttfamily,
caption={Manhattan ($L_1$) distance.}, 
label={code:manhattan_dist}, 
captionpos=b
]
def manhattan_distance(Q, K):
    # Q, K: [B, N, d_k]
    # Returns D: [B, N, N]
    Q_exp = Q.unsqueeze(2)              # [B, N, 1, d_k]
    K_exp = K.unsqueeze(1)              # [B, 1, N, d_k]
    return torch.sum(torch.abs(Q_exp - K_exp), dim=-1)  # [B, N, N]
\end{lstlisting}
\vspace{-10pt}

\subsubsection{Minkowski Distance}
Minkowski distance is a parameterized family that generalizes both Euclidean and Manhattan distance via a parameter $p \geq 1$:

\begin{equation}
    D(q, k) = \|q - k\|_p
    = \left(\sum_{i=1}^{d} |q_i - k_i|^p\right)^{1/p}.
\end{equation}

Setting $p = 2$ recovers Euclidean distance and $p = 1$ recovers Manhattan distance. In the limit $p \to \infty$, the Minkowski distance converges to the Chebyshev distance $D(q, k) = \max_i |q_i - k_i|$, which is determined solely by the largest per-dimension deviation. 

\vspace{10pt}
\begin{lstlisting}[
nolol, 
basicstyle=\footnotesize\ttfamily,
caption={Minkowski distance. Setting \texttt{p=2} recovers Euclidean
         distance; \texttt{p=1} recovers Manhattan distance.}, 
label={code:minkowski_dist}, 
captionpos=b
]
def minkowski_distance(Q, K, p=2):
    # Q, K: [B, N, d_k]
    # Returns D: [B, N, N]
    Q_exp = Q.unsqueeze(2)              # [B, N, 1, d_k]
    K_exp = K.unsqueeze(1)              # [B, 1, N, d_k]
    return torch.sum(torch.abs(Q_exp - K_exp) ** p, dim=-1) ** (1.0 / p)
\end{lstlisting}
\vspace{-10pt}

\subsubsection{Mahalanobis Distance}
Mahalnobis distance accounts for correlations between dimensions by scaling with the inverse covariance matrix $\Sigma^{-1}$ of the feature distribution:

\begin{equation}
    D(q, k)
    = \sqrt{(q - k)^\top \Sigma^{-1} (q - k)}.
\end{equation}

When $\Sigma = I$, this reduces to Euclidean distance. Unlike isotropic distance metrics, Mahlanobis distance is invariant to linear transformations of the feature space and naturally handles anisotropic or correlated feature distributions. It is particularly relevant when feature dimensions have significantly different variances. 

\vspace{10pt}
\begin{lstlisting}[
nolol, 
basicstyle=\footnotesize\ttfamily,
caption={Mahalanobis distance. \texttt{S\_inv} is the precomputed
         inverse covariance matrix $\Sigma^{-1}$ of the feature space.}, 
label={code:mahalanobis_dist}, 
captionpos=b
]
def mahalanobis_distance(Q, K, S_inv):
    # Q, K:   [B, N, d_k]
    # S_inv:  [d_k, d_k]  inverse covariance matrix
    # Returns D: [B, N, N]
    diff    = Q.unsqueeze(2) - K.unsqueeze(1)      # [B, N, N, d_k]
    # diff @ S_inv @ diff^T per pair
    left    = torch.matmul(diff, S_inv)             # [B, N, N, d_k]
    dist_sq = torch.sum(left * diff, dim=-1)        # [B, N, N]
    return torch.sqrt(dist_sq.clamp(min=0))
\end{lstlisting}
\vspace{-5pt}

\subsubsection{Pearson Correlation}
Pearson correlation is a mean-centered variant of cosine similarity. Let $\bar{q} = \frac{1}{d}\sum_i q_i$ and $\bar{k} = \frac{1}{d}\sum_i k_i$ denote the per-vector means. Then: 

\begin{equation}
    \text{sim}(q, k)
    = \frac{\displaystyle\sum_{i=1}^{d}(q_i - \bar{q})(k_i - \bar{k})}
           {\sqrt{\displaystyle\sum_{i=1}^{d}(q_i - \bar{q})^2}\;
            \sqrt{\displaystyle\sum_{i=1}^{d}(k_i - \bar{k})^2}}.
\end{equation}

By subtracting the mean, Pearson correlation is invariant to both the scale and the offset of the feature vectors, measuring purely the linear co-variation in their fluctuations. 

\vspace{10pt}
\begin{lstlisting}[
nolol, 
basicstyle=\footnotesize\ttfamily,
caption={Pearson correlation. Equivalent to cosine similarity applied
         to mean-centered feature vectors.}, 
label={code:pearson_corr}, 
captionpos=b
]
def pearson_correlation(Q, K):
    # Q, K: [B, N, d_k]
    # Returns S: [B, N, N]
    Q_c = Q - Q.mean(dim=-1, keepdim=True)   # mean-center along d_k
    K_c = K - K.mean(dim=-1, keepdim=True)
    Q_n = F.normalize(Q_c, p=2, dim=-1)      # L2-normalize
    K_n = F.normalize(K_c, p=2, dim=-1)
    return torch.matmul(Q_n, K_n.transpose(-2, -1))
\end{lstlisting}
\vspace{-5pt}

\subsubsection{Bilinear Similarity}
Bilinear similarity introduces a learned weight matrix $W \in \mathbb{R}^{d \times d}$ to parameterize the interaction between $q$ and $k$:

\begin{equation}
    \text{sim}(q, k) = q^\top W k.
\end{equation}

This is strict generalization of the dot product (recovered when $W = I$) and subsumes the standard attention scoring function when $W = {W^Q}^\top W^K$ decomposes into query and key projection matrices. The learned $W$ allows the model to capture asymmetric or directional relationships between query and key features that an isotropic dot product cannot express.

\vspace{10pt}
\begin{lstlisting}[
nolol, 
basicstyle=\footnotesize\ttfamily,
caption={Bilinear similarity with learned weight matrix $W \in
         \mathbb{R}^{d_k \times d_k}$. Reduces to dot-product
         similarity when $W = I$.}, 
label={code:bilinear_sim}, 
captionpos=b
]
class BilinearSimilarity(nn.Module):
    def __init__(self, d_k):
        super().__init__()
        self.W = nn.Linear(d_k, d_k, bias=False)  # learned W in R^{d x d}

    def forward(self, Q, K):
        # Q, K: [B, N, d_k]
        # Returns S: [B, N, N]
        return torch.matmul(Q, self.W(K).transpose(-2, -1))
\end{lstlisting}
\vspace{-5pt}

\subsection{Positional Encoding}
The \convnn framework supports several positional encoding strategies that inject spatial structure into the similarity computation step. These range from lightweight coordinate concatenation to learned and input-adaptive schemes. Each variant offers a different tradeoff between expressiveness, computational cost, and translation invariance.  

\subsubsection{Raw Normalized Coordinates}
The simplest positional encoding appends normalized spatial coordinates directly to the input along the channel dimension. For a 1D input of shape $[N \times C]$ with $C$ channels and $N$ input sequence, a single channel encoding the linearized position is appended, yielding a tensor of shape $[N \times (C + 1)]$. For a 2D input of shape $[H \times W \times C]$ with spatial dimensions $H \times W$ and $C$ channels, two channels encoding the horizontal and vertical coordinates $(i, j)$ are appended yielding a tensor of shape $[H \times W \times (C + 2)]$. This lightweight encoding enables \convnn to blend feature-based and spatially-aware similarity computations with minimal overhead. 

Given an input $X \in \mathbb{R}^{N \times C}$, a single coordinate channel encoding the normalized position $p_i \in [\alpha, \beta]$ is appended along the channel dimension, yielding $X_{\text{pos}} \in \mathbb{R}^{N \times (C+1)}$, where: 

\begin{equation}
    p_i = \alpha + (\beta - \alpha) \cdot \frac{i}{N - 1},
    \qquad i = 0, \dots, N-1.
\end{equation}

For a 2D input $\mathbf{X} \in \mathbb{R}^{H \times W \times C}$, two coordinate channels encoding the normalized horizontal and vertical positions are appended, yielding $\mathbf{X}_{\text{pos}} \in \mathbb{R}^{H \times W \times (C+2)}$, where for spatial position $(i, j)$:

\begin{equation}
    p_{(i,j)}^x = \alpha + (\beta - \alpha) \cdot \frac{i}{H-1},
    \qquad
    p_{(i,j)}^y = \alpha + (\beta - \alpha) \cdot \frac{j}{W-1},
    \qquad (i,j) \in [0, H-1] \times [0, W-1].
\end{equation}

Common choices are $[\alpha, \beta] = [0, 1]$, which preserves non-negative magnitudes, and $[\alpha, \beta] = [-1, 1]$, which centers the coordinate space around the origin and is often preferred when coordinates are used by dot-product similarity functions. the choice of normalization range does not affect the relative ordering of positions and therefore does not alter the neighbor selection behavior of \convnn.

\subsubsection{Fourier Feature Positional Encoding}
Raw normalized coordinates are smooth and low-frequency but can limit the model's ability to resolve fine-grained spatial structure. To address this, projected spatial coordinates into a higher-dimensional space using random Fourier features (\cite{tancik2020fourier}) can be used. For a 2D coordinate $\mathbf{v} = (i, j)$ normalized to $[-1, 1]^2$, we compute the feature mapping $\gamma(\mathbf{v}) \in \mathbb{R}^{2d}$ as: 
\begin{equation}
    \gamma(\mathbf{v})
    = \bigl[\sin(2\pi B\mathbf{v}),\;
             \cos(2\pi B\mathbf{v})\bigr],
\end{equation}
\noindent where $B \in \mathbb{R}^{d \times 2}$ is a random Gaussian matrix sampled at initialization and kept frozen throughout training, and $d$ is the projection dimension. The resulting high-dimensional coordinate vector is concatenated to the input features along the channel dimension. By spanning a broad range of frequencies, this encoding allows the similarity computation to better resolve high-frequency spatial variations and fine-grained local structures. 

\subsubsection{Relative Positional Bias}
Rather than modifying the input features, relative positional bias injects spatial structure directly into the similarity scores. Let $\Delta_{ij} = (\Delta x, \Delta y)$ denote the 2D relative offset between the spatial positions of query $\mathbf{x}_i$ and candidate $\mathbf{x}_j$. This introduces a learnable bias table $\mathbf{B} \in \mathbb{R}^{(2H-1) \times (2W-1)}$, which covers all possible relative offsets within a feature map of spatial dimensions at $H \times W$. The raw similarity score $S_{ij}$ is the augmented as: 
\begin{equation}
    S'_{ij} = S_{ij} + \mathbf{B}[\Delta_{ij}].
\end{equation}
Because the bias depends only on the relative offset $\Delta_{ij}$ and not on absolute positions, this formulation preserves translational equivariance. Neighbor selection is thereby influenced by spatial proximity in a manner that is consistent across all locations in the feature map. 

\subsubsection{Conditional Positional Encoding}
Static positional encodings are fixed at initialization and cannot adapt to the input content or resolution. Conditional Positional Encoding (CPE) (\cite{chu2021conditional}) addresses this dynamically generating position-aware features from the input itself. Concretely, given an input feature map $\mathbf{X} \in \mathbb{R}^{H \times W \times C}$, we apply a lightweight zero-padded depthwise convolution and combine the result via a residual connection: 
\begin{equation}
    \mathbf{X}' = \mathbf{X} + \text{DWConv}(\mathbf{X}).
\end{equation}
The zero-padding of the depthwise convolution provides implicit positional anchors at the spatial boundaries, allowing the network to infer absolute position from context. CPE is resolution-agnostic and introduces minimal additional parameters, making it well-suited to a variable-resolution inputs. 



\subsection{Sparse Candidate Search} 

\begin{figure}[H]
    \vspace{5pt}
    \centering
    \begin{subfigure}{0.25\linewidth}
        \centering
        \includegraphics[width=\linewidth]{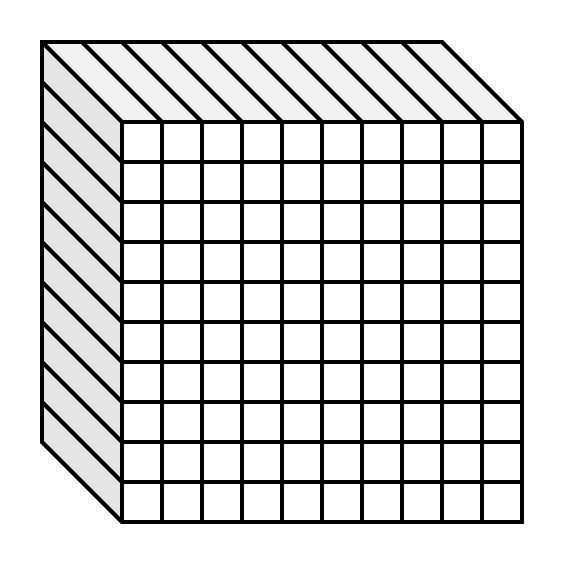}
        \caption{\centering Input Tensor}
    \end{subfigure}%
    \hfill
    \begin{subfigure}{0.25\linewidth}
        \centering
        \includegraphics[width=\linewidth]{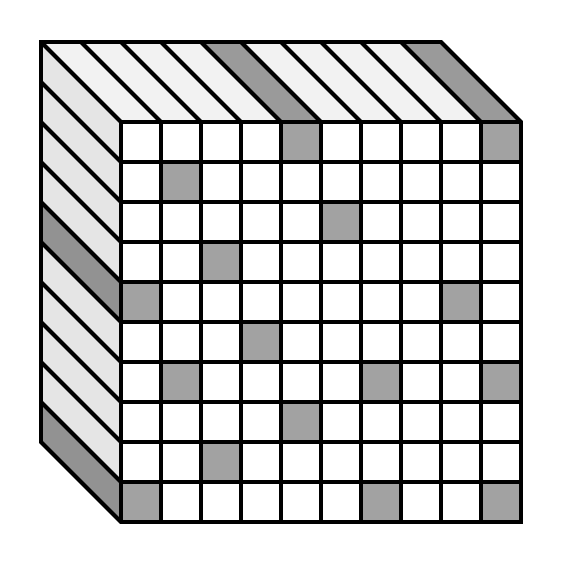}
        \caption{\centering Random}\label{fig:random}
    \end{subfigure}%
    \hfill
    \begin{subfigure}{0.25\linewidth}
        \centering
        \includegraphics[width=\linewidth]{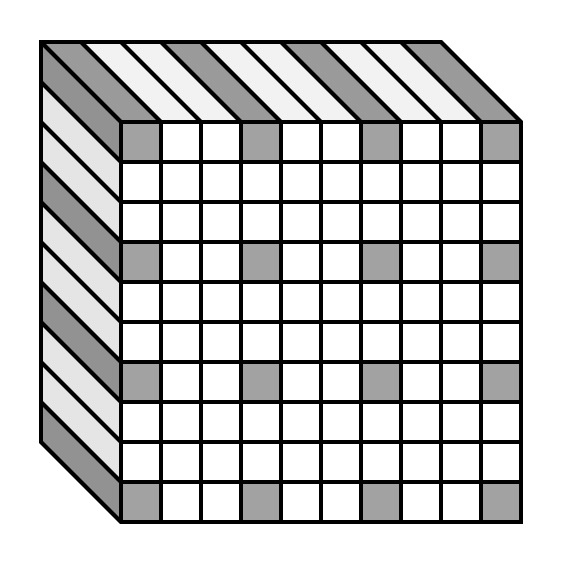}
        \caption{\centering Spatial}\label{fig:spatial}
    \end{subfigure}%
    \vspace{-5pt}
    \caption{Candidate selection strategies for sparse neighbor search with
             $r = 16$. Selected candidate features are highlighted in gray.}
    \label{fig:candidate_strategies}
\end{figure}

Exhaustive $k$-NN search over all $N$ input features incurs $O(N^2)$ similarity computations, which is prohibitive for large feature maps. To reduce this cost, we restrict the search to a reduced candidate set $\mathbf{X}^{\text{sparse}}$ of size $r \ll N$. Reducing the candidate set from $N$ to $r \ll N$ yields two complementary gains. The similarity computation cost reduces from $O(N^2)$ to $O(N \cdot r)$, since each of the $N$ queries is compared only against $r$ candidates rather than the full sequence. The subsequent $\text{top-}K$ selection over $r$ scores per query reduces from $O(N^2 \log N)$ to $O(N \cdot r \log r)$. The similarity matrix memory footprint reduces from $O(N^2)$ to $O(N \cdot r)$, which is the dominant saving in practice since the $N \times N$ is the primary source of memory pressure for large feature maps. We consider two candidate selection strategies, illustrated in Figure~\ref{fig:candidate_strategies}: random selection and spatial subsampling. 

To incorporate sparse neighbor search into \convnn, the key matrix $K \in \mathbb{R}^{N \times C_{qk}}$ is subsampled to a candidate set $K^{\text{sparse}} \in \mathbb{R}^{r \times C_{qk}}$ using indices drawn by either random or spatial selection. The query matrix $Q \in \mathbb{R}^{N \times C_{qk}}$ remains over all $N$ positions, so the similarity computation reduces from $\mathbb{R}^{N \times N}$ to $\mathbb{R}^{N \times r}$, lowering the per-query cost from $O(N)$ to $O(r)$. After applying $\text{top-}K$ over the $r$ candidate scores, the resulting indices refer to positions within the reduced candidate set and must be remapped to their original positions in the full sequence using the stored candidate indices. For positions that appear in the candidate set, their self-similarity score is set to $-\infty$ before selection to prevent a position from selecting itself as a neighbor. The remapped indices are then concatenated with each query's own index to form the final $K$-neighbor index set, and the aggregation proceeds identically to the full \convnn\ forward pass in Code~\ref{code:convnn_self_attention}. 

\vspace{10pt}
\begin{lstlisting}[
    nolol, 
    basicstyle=\footnotesize\ttfamily,
    caption={Sparse \convnn\ with random or spatial
             candidate selection. Mirrors the interface
             of \texttt{ConvNN} with $r \ll N$ candidate
             keys substituted for the full key matrix.}, 
    captionpos=b
]
class SparseConvNN(nn.Module):
    def __init__(self, C, C_qk, C_v, C_out, K,
                 num_samples, sampling_type='random', sample_padding=0):
        """
        Args:
            C             : input channel dimension
            C_qk          : projected query/key dimension
            C_v           : projected value dimension
            C_out         : output channel dimension C'
            K             : number of nearest neighbors
            num_samples   : candidate pool size r (r << N)
            sampling_type : 'random' or 'spatial'
            sample_padding: padding for spatial boundary exclusion
        """
        super().__init__()
        self.K             = K
        self.num_samples   = num_samples
        self.sampling_type = sampling_type
        self.sample_padding = sample_padding

        self.f_Q  = nn.Linear(C, C_qk)
        self.f_K  = nn.Linear(C, C_qk)
        self.f_V  = nn.Linear(C, C_v)
        self.rho  = nn.Softmax(dim=-1)
        self.conv = nn.Conv1d(
            in_channels=C_v, out_channels=C_out,
            kernel_size=K, stride=K
        )

    def similarity(self, Q, K_sparse):
        # Q: [B, N, C_qk],  K_sparse: [B, r, C_qk]
        # Returns S: [B, N, r]
        return Q @ K_sparse.transpose(-2, -1)

    def _get_sample_indices(self, N, device):
        # Returns r indices into [0, N) for sparse key selection
        if self.sampling_type == 'random':
            return torch.randperm(N, device=device)[:self.num_samples]
        elif self.sampling_type == 'spatial':
            return torch.linspace(
                self.sample_padding,
                N - self.sample_padding - 1,
                self.num_samples,
                device=device
            ).long()

    def forward(self, X):
        # X: [B, N, C]
        B, N, C = X.shape

        # Step 1 - Similarity Computation over sparse candidates
        # Q: [B, N, C_qk],  K, V: full sequence
        Q, K, V = self.f_Q(X), self.f_K(X), self.f_V(X)
        C_v = V.shape[-1]

        sample_idx = self._get_sample_indices(N, X.device)    # [r]
        K_sparse   = K[:, sample_idx, :]                      # [B, r, C_qk]
        S = self.similarity(Q, K_sparse)                      # [B, N, r]

        # Mask self-similarity for positions in the candidate set
        pos_idx = torch.arange(len(sample_idx), device=X.device)
        S[:, sample_idx, pos_idx] = float('-inf')

        # Step 2 - Neighbor Selection: top-(K-1) from r candidates
        # K-th neighbor is the query itself (self-connection)
        s, I = torch.topk(S, k=self.K - 1, dim=-1)           # [B, N, K-1]

        # Remap candidate indices to original sequence positions
        I_mapped = sample_idx[I]                              # [B, N, K-1]

        # Prepend each query's own index as the K-th neighbor
        self_idx  = torch.arange(N, device=X.device).view(1, N, 1).expand(B, N, 1)
        I_full    = torch.cat([self_idx, I_mapped], dim=-1)   # [B, N, K]

        # Prepend self-score of 1, then apply softmax over all K scores
        ones  = torch.ones(B, N, 1, device=X.device)
        s_all = torch.cat([ones, s], dim=-1)                  # [B, N, K]
        s_all = self.rho(s_all)                               # [B, N, K]

        # Gather V[I_full, :] in R^{K x C_v}
        # V: [B, N, C_v] -> [B, C_v, N]
        # expand to [B, C_v, N, K] for gather along dim=2
        V_t   = V.permute(0, 2, 1)                            # [B, C_v, N]
        V_exp = V_t.unsqueeze(-1).expand(B, C_v, N, self.K)   # [B, C_v, N, K]
        I_exp = I_full.unsqueeze(1).expand(B, C_v, N, self.K) # [B, C_v, N, K]
        V_nn  = torch.gather(V_exp, dim=2, index=I_exp)       # [B, C_v, N, K]

        # X_nn_i = diag(rho(s_i)) * V[I_i, :]
        s_exp = s_all.unsqueeze(1).expand(B, C_v, N, self.K)  # [B, C_v, N, K]
        X_nn  = s_exp * V_nn                                  # [B, C_v, N, K]

        # Step 3 - Weighted Aggregation
        # Reshape to [B, C_v, N*K] for Conv1D with kernel_size=K, stride=K
        X_nn = X_nn.permute(0, 1, 3, 2).reshape(B, C_v, -1)  # [B, C_v, N*K]
        Y    = self.conv(X_nn)                                # [B, C_out, N]
        Y    = Y.permute(0, 2, 1)                             # [B, N, C_out]
        return Y
\end{lstlisting}
\label{code:sparse_convnn}

\subsubsection{Random Selection}
The simplest strategy samples $r$ features uniformly at random without replacement. Formally, let $I_r \subset \{1, \dots, n\}$ be a set of $r$ indices drawn uniformly without replacement. The candidate set is then: 

\begin{equation}
    X^{\text{sparse}} = X[I_r, :] \in \mathbb{R}^{r \times C}.
\end{equation}

Random selection requires no spatial assumptions, making it applicable to both 1D and 2D inputs. It preserves feature diversity across the full spatial extent, though it offers no guarantee of spatial coverage. 

\subsubsection{Spatial Subsampling}
Spatial subsampling selects features at regular intervals, ensuring uniform coverage of the spatial domains. For a 1D input $X \in \mathbb{R}^{N \times C}$ with stride $s = N / r$, the candidate set $X^{\text{sparse}} \in \mathbb{R}^{r \times C}$ is obtained by: 

\begin{equation}
    X^{\text{sparse}}[i, :] = X[s \cdot i, :],
    \qquad i = 0, \dots, r-1.
\end{equation}

For a 2D input $\mathbf{X} \in \mathbb{R}^{H \times W \times C}$, a spatial stride of $s = \sqrt{r}$ is applied along both axes, yielding
$\mathbf{X}^{\text{sparse}} \in \mathbb{R}^{H/s) \times (W/s) \times C}$:

\begin{equation}
    \mathbf{X}^{\text{sparse}}[i, j, :] = \mathbf{X}[s \cdot i,\; s \cdot j, :],
    \qquad i = 0, \dots, \tfrac{H}{s}-1,\quad j = 0, \dots, \tfrac{W}{s}-1.
\end{equation}

Spatial Subsampling is deterministic and preserves the geometric layout of the feature map, but discards all features at non-sampled locations.

\subsection{Score-to-Weight Mapping}
After computing raw similarity scores $s_{ij} = \text{sim}(q_i, k_j)$ between query $q_i$ and each candidate $k_j \in X^{\text{sparse}}$, an aggregate weight function $\rho$ maps the score vector $\mathbf{s}_i = [s_{i,1}, \dots, s_{i,N}]^\top \in \mathbb{R}^N$ to a weight vector $\mathbf{w}_i = f(\mathbf{s}_i) \in \mathbb{R}^N$, which is then used before the weighted aggregation step. The choice of $\rho$ controls the sharpness, sparsity, and relative contribution of each neighbor.

\subsubsection{Softmax}
The standard softmax normalizes all scores into a valid probability distribution over the $N$ candidates: 

\begin{equation}
    \mathbf{w}_{ij} = \text{softmax}(\mathbf{s}_i)_j
    = \frac{\exp(s_{ij})}{\displaystyle\sum_{j'=1}^{N} \exp(s_{ij'})}.
\end{equation}

All neighbors receive strictly positive weight, with higher-scoring candidate contributing proportionally more to the attention weight. This is the default weight function used in standard dot-product attention (\cite{vaswani2017attention}). 

\subsubsection{Softmax with Temperature}
Introducing a temperature parameter $T > 0$ controls the sharpness of the softmax distribution: 

\begin{equation}
    \mathbf{w}_{ij} = \text{softmax}(\mathbf{s}_i / T)_j
    = \frac{\exp(s_{ij} / T)}{\displaystyle\sum_{j'=1}^{N} \exp(s_{ij'} / T)}.
\end{equation}

As $T \to 0$, the distribution concentrates on a single highest-scoring neighbor, approaching a hard $\arg\max$ selection. As $T \to \infty$, the distribution becomes uniform, assigning equal weight to all candidates. $T$ can be fixed as a hyperparameter or learned during training. 

\subsubsection{Softmin}
Softmin inverts the scoring by negating the raw scores before applying softmax, assigning larger weights to candidate with \emph{lower} similarity scores: 

\begin{equation}
    \mathbf{w}_{ij} = \text{softmin}(\mathbf{s}_i)_j
    = \frac{\exp(-s_{ij})}{\displaystyle\sum_{j'=1}^{N} \exp(-s_{ij'})}.
\end{equation}

When $\text{sim}(q, k)$ is a distance metric (e.g., Euclidean or Manhattan distance), softmin assigns higher weight to spatially or semantically closer neighbors, making it the natural counterpart to softmax for distance-based similarity computations. 

\subsubsection{Sparsemax}
Sparsemax (\cite{martins2016softmax}) projects the score vector onto the probability simplex $\Delta^{r-1}$ via a Euclidean projection, yielding \emph{sparse} weight vectors in which low-scoring candidates receive exactly zero weight: 

\begin{equation}
    \mathbf{w}_i = \text{sparsemax}(\mathbf{s}_i)
    = \underset{\mathbf{p} \,\in\, \Delta^{r-1}}{\arg\min}
      \|\mathbf{p} - \mathbf{s}_i\|_2^2.
\end{equation}

Unlike softmax, which produces dense weights, sparsemax encourages the aggregation to focus on a subset of high-scoring neighbors while ignoring the rest entirely. It remains differentiable almost everywhere, preserving end-to-end trainability. 

\subsection{Patch-Level Comparison}
Individual feature comparisons may fail to capture local texture and structural patterns. To address this limitation, \convnn supports patch-level comparisons of size $p \times p$, where $p=1$ corresponds to the standard per-feature case.

\begin{figure}[H]
\vspace{5pt}
    \begin{subfigure}[t]{0.25\linewidth}
        \centering
        \includegraphics[width=\linewidth]{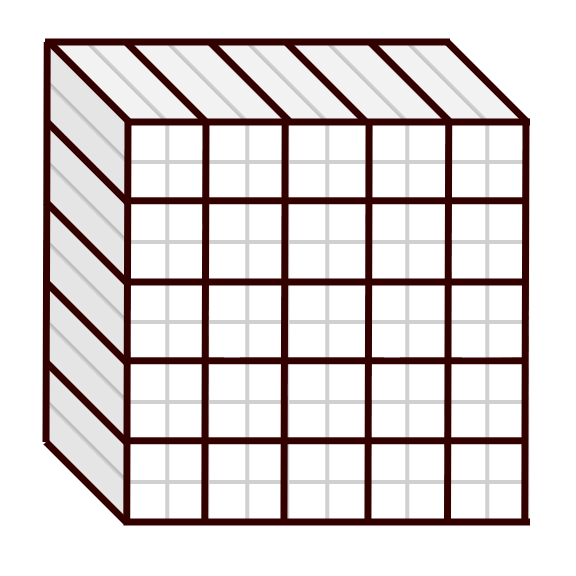}
        \caption{\centering Locality on Input Tensor}
    \end{subfigure}%
    \hfill
    \begin{subfigure}[t]{0.25\linewidth}
        \centering
        \includegraphics[width=\linewidth]{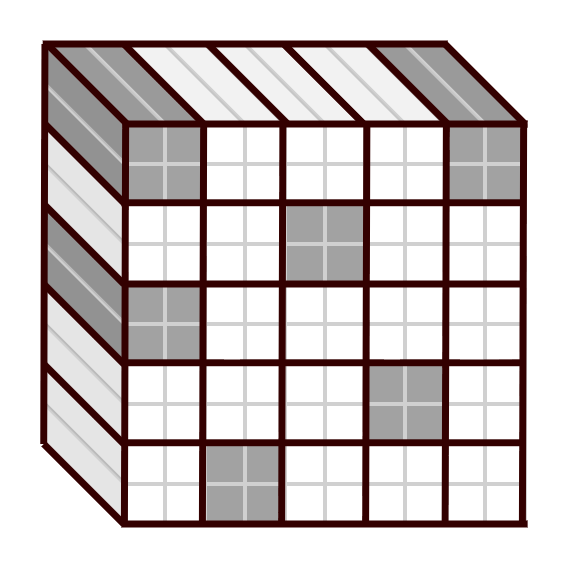}
        \caption{\centering Locality with Random Selection}\label{fig:random_loc}
    \end{subfigure}%
    \hfill
    \begin{subfigure}[t]{0.25\linewidth}
        \centering
        \includegraphics[width=\linewidth]{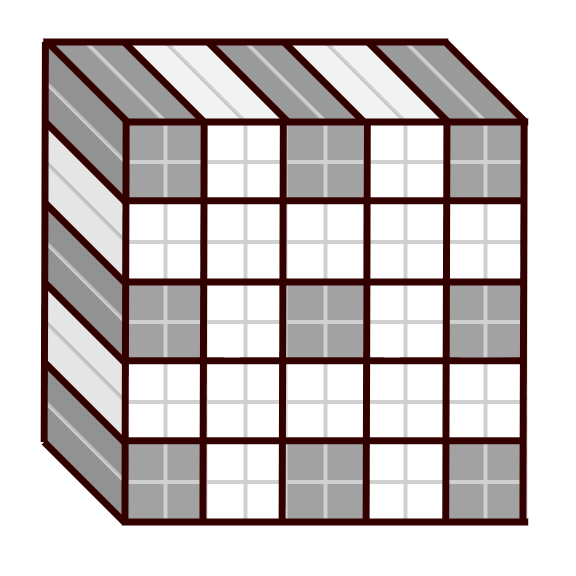}
        \caption{\centering Locality with spatial Selection}\label{fig:spatial_loc}
    \end{subfigure}%
    \caption{Patch-level locality with neighbor candidate selection.}
\end{figure}

This is implemented by introducing a $\operatorname{PixelUnshuffle}$ (\cite{shi2016real}) preprocessing layer, which increases the number of channels by a factor of $p^2$ while reducing spatial resolution by $p$. This transformation enables richer, region-aware feature interactions prior to neighbor selection. Formally, for an input of shape $[H, W, C]$, its patchified version has shape $[h/p, w/p, cp^2]$. Figures~\ref{fig:random_loc} and~\ref{fig:spatial_loc} illustrate the patchification process in the context of random and spatial feature selection, respectively.

Formally, for an input $\mathbf{X} \in \mathbb{R}^{H \times W \times C}$, the $\operatorname{PixelUnshuffle}$ operation with downscale factor $p$ produces a patchified tensor $\mathbf{X}_p \in \mathbb{R}^{(H/p) \times (W/p) \times (Cp^2)}$ by folding each $p \times p$ spatial block into the channel dimension: 

\begin{equation}
    \mathbf{X}_p\!\left[\,i,\, j,\, c \cdot p^2 + r \cdot p + s\,\right]
    = \mathbf{X}\!\left[\,i \cdot p + r,\; j \cdot p + s,\; c\,\right],
\end{equation}

\noindent where $i \in \{0, \dots, H/p - 1\}$, $j \in \{0, \dots, W/p - 1\}$, $c \in \{0, \dots, C-1\}$, and $r, s \in \{0, \dots, p-1\}$ index the row and column offsets within each $p \times p$ patch. Setting $p = 1$ recovers the identity, leaving the input unchanged and reducing to the standard per-feature case.

After the \convnn\ operation is applied to the patchified tensor $\mathbf{X}_p \in \mathbb{R}^{(H/p) \times (W/p) \times (C'p^2)}$, the original spatial resolution is restored by applying the inverse $\operatorname{PixelShuffle}$ operation (\cite{shi2016real}) with upscale factor $p$. This folds $p^2$ channel entries back into the $p \times p$ spatial block, yielding an output tensor of shape $\mathbb{R}^{H \times W \times C'}$:

\begin{equation}
    \mathbf{Y}\!\left[\,i \cdot p + r,\; j \cdot p + s,\; c\,\right]
    = \mathbf{X}_p'\!\left[\,i,\, j,\, c \cdot p^2 + r \cdot p + s\,\right],
\end{equation}

\noindent where $\mathbf{X}_p'$ denotes the output of the \convnn\ operation on the patchified input. This is the exact inverse of the $\operatorname{PixelUnshuffle}$ mapping defined above, recovering the original spatial dimensions $H \times W$ while reducing the channel dimension from $C'p^2$ back to $C'$. 

\section{Connection between \convnn\ and Convolutional Operations}
\label{sec:convnn_conv_connection}
We now show that standard and depthwise convolutions are special cases of \convnn. The key insight is that by restricting the query and key projections to raw normalized spatial coordinates and using Euclidean distance as the similarity function, the $k$-NN neighbor selection in \convnn\ recovers exactly the local spatial neighborhoods defined by a convolutional kernel window. The value projection is set to the identity, so no feature transformation occurs before aggregation. We present the construction for both 1D and 2D inputs. 

\paragraph{1D Convolution.}
Given an input $X \in \mathbb{R}^{N \times C}$, we first append the raw normalized 1D coordinate to the input along the channel dimension, yielding $X_{\text{pos}} \in \mathbb{R}^{N \times (C+1)}$, where the appended coordinate for the position $i$ is $p_i = i / (N -1) \in [0, 1]$. The query, key, and value projections are then defined as: 
\begin{equation}
    Q = X_{\text{pos}}[:, -1:], \quad
    K = X_{\text{pos}}[:, -1:], \quad
    V = X,
    \qquad Q, K \in \mathbb{R}^{N \times 1}, \quad V \in \mathbb{R}^{N \times C}.
\end{equation}
The similarity matrix is computed as the negative exponential squared Euclidean distance between positional coordinates: 
\begin{equation}
    S_{ij} = \exp(-\|p_i - p_j\|_2^2), \qquad S \in \mathbb{R}^{N \times N},
\end{equation}
\noindent so that positions close in space receive higher similarity scores. Applying $\argkmax_k$ to $S$ yields the $K$ spatially nearest neighbor indices for each query position: 
\begin{equation}
    I_i = \argkmax_k(S)[i, :] \in \{0, \dots, N-1\}^{K}.
\end{equation}
Setting $\rho (\mathbf{z}) = \mathbf{1}_K$ (uniform weights, no softmax), the neighborhood matrix for position $i$ is: 
\begin{equation}
    X_{\text{nn},i} = V[I_i, :] \in \mathbb{R}^{K \times C}.
\end{equation}

Concatenating over all positions gives $X_{\text{nn}} \in \mathbb{R}^{(KN) \times C}$. Applying a standard \textsc{Conv1D} with $C'$ output channels, kernel size $K$, and stride $K$ yields: 

\begin{equation}
    y_{i,c'} = \operatorname{StandardConv1D}(V[I_i,:]) = \sum_{k=1}^{K} \sum_{c=1}^{C}
    V[I_{i,k},\, c] \cdot w_{k,c,c'}.
\end{equation}
As shown in Section~\ref{subsubsec:pe_mimic_conv}, when $K = R$ the Euclidean-distance neighbor selection recovers a centered local window around $i$, making the expression above equivalent to standard 1D convolution from Section~\ref{subsec:std_conv}: 
\begin{equation}
    \text{Conv}(W, X)_{(i,c')}
    = \sum_{p=-\lfloor R/2 \rfloor}^{\lfloor R/2 \rfloor}\sum_{c=1}^{C}
      W_{p,c,c'} \cdot X_{i+p,\, c}.
\end{equation}
Analogously, applying a depthwise \textsc{Conv1D} with kernel size $K$ and stride $K$ recovers depthwise 1D convolution from Section~\ref{subsec:dw_conv}: 
\begin{equation}
    y_{i,c} = \sum_{k=1}^{K}
    V[I_{i,k},\, c] \cdot w_{k,c}
    \;\equiv\;
    \text{DepthwiseConv}(W, X)_{(i,c)}
    = \sum_{p=-\lfloor R/2 \rfloor}^{\lfloor R/2 \rfloor}
      W_{p,c} \cdot X_{i+p,\, c}.
\end{equation}

\paragraph{2D Convolution.}
Given an input $\mathbf{X} \in \mathbb{R}^{H \times W \times C}$, we append the raw normalized 2D spatial coordinates $(p_i^x, p_i^y)$ to the input along the channel dimension, yielding $\mathbf{X}_{\text{pos}} \in \mathbb{R}^{H \times W \times (C+2)}$, where for spatial position $(i, j)$: 
\begin{equation}
    p_{(i,j)}^x = \frac{i}{H-1} \in [0,1], \qquad
    p_{(i,j)}^y = \frac{j}{W-1} \in [0,1].
\end{equation}
The spatial dimensions are then flattened to obtain $X_{\text{pos, flat}} \in \mathbb{R}^{HW \times (C+2)}$. Letting $N = HW$, the query, key, and value projections are defined as: 

\begin{equation}
    Q = X_{\text{pos,flat}}[:, -2:], \quad
    K = X_{\text{pos,flat}}[:, -2:], \quad
    V = X_{\text{flat}},
    \qquad Q, K \in \mathbb{R}^{N \times 2}, \quad V \in \mathbb{R}^{N \times C}.
\end{equation}
The similarity matrix is the negative exponential squared Euclidean distance between 2D coordinate pairs: 
\begin{equation}
    S_{ij} = \exp(-\|p_i - p_j\|_2^2), \qquad S \in \mathbb{R}^{N \times N},
    \qquad p_i = (p_i^x, p_i^y) \in \mathbb{R}^2.
\end{equation}

The remainder of the procedure follows the 1D case exactly, with $\argkmax_k(S)$ selecting the $K = R^2$ spatially nearest neighbors for each query position. The resulting \textsc{Conv1D} aggregation is equivalent to standard 2D convolution from Section~\ref{subsec:std_conv}: 

\begin{equation}
    y_{(i,j),c'} = \sum_{k=1}^{R^2} \sum_{c=1}^{C}
    V[I_{(i,j),k},\, c] \cdot w_{k,c,c'}
    \;\equiv\;
    \text{Conv}(W, \mathbf{X})_{(i,j,c')}
    = \sum_{p=-\lfloor R/2\rfloor}^{\lfloor R/2\rfloor}
      \sum_{q=-\lfloor R/2\rfloor}^{\lfloor R/2\rfloor}
      \sum_{c=1}^{C}
      W_{p,q,c,c'} \cdot \mathbf{X}_{i+p,\, j+q,\, c},
\end{equation}
\noindent and depthwise \textsc{Conv1D} aggregation recovers depthwise 2D convolution from Section~\ref{subsec:dw_conv}: 
\begin{equation}
    y_{(i,j),c} = \sum_{k=1}^{R^2}
    V[I_{(i,j),k},\, c] \cdot w_{k,c}
    \;\equiv\;
    \text{DepthwiseConv}(W, \mathbf{X})_{(i,j,c)}
    = \sum_{p=-\lfloor R/2\rfloor}^{\lfloor R/2\rfloor}
      \sum_{q=-\lfloor R/2\rfloor}^{\lfloor R/2\rfloor}
      W_{p,q,c} \cdot \mathbf{X}_{i+p,\, j+q,\, c}.
\end{equation}

\subsection{Positional Encoding Mimics Spatial Summation}
\label{subsubsec:pe_mimic_conv}
We now show formally that Euclidean-distance neighbor selection over normalized spatial coordinates recovers the centered local window of a convolutional kernel, and characterize its behavior at sequence boundaries.

\paragraph{1D case.}
Let positions be indexed $i = 0, \dots, N-1$ with normalized coordinates $p_i = i/(N-1) \in [0, 1]$. The negative exponential squared Euclidean similarity score between position $i$ and $j$ is: 
\begin{equation}
    S_{ij} = \exp(-\|p_i - p_j\|_2^2) = \exp(-\left(\frac{i-j}{N-1}\right)^2).
\end{equation}

Since $S_{ij}$ is strictly decreasing in $|i - j|$, the $K = R$ indices with the highest similarity to position $i$ are exactly the $R$ positions closest in index distance. For interior positions where $\lfloor R/2 \rfloor \leq i \leq N - 1 - \lfloor R/2 \rfloor$, these form a symmetric centered window: 
\begin{equation}
    I_i = \left\{i - \left\lfloor\frac{R}{2}\right\rfloor,\;
                 \dots,\; i-1,\; i,\; i+1,\;
                 \dots,\; i + \left\lfloor\frac{R}{2}\right\rfloor\right\},
\end{equation}

\noindent placing $i$ at the center, which matches the standard centered convolution convention. At the sequence boundaries, however, the window cannot extend symmetrically in both directions. Specifically: 

\begin{itemize}
    \item \textbf{Left boundary} ($i < \lfloor R/2 \rfloor$): there are
    fewer than $\lfloor R/2 \rfloor$ positions to the left of $i$, so
    $\argkmax_k$ selects all available left neighbors and fills the
    remaining slots from the right, producing a \emph{right-skewed}
    window. In the extreme case $i = 0$, the window is fully
    right-aligned: $I_0 = \{0, 1, \dots, R-1\}$.

    \item \textbf{Right boundary} ($i > N - 1 - \lfloor R/2 \rfloor$):
    symmetrically, the window becomes \emph{left-skewed}. In the extreme
    case $i = N-1$, the window is fully left-aligned:
    $I_{N-1} = \{N-R, \dots, N-2, N-1\}$.
\end{itemize}

This boundary behavior corresponds exactly to the behavior of a zero-padding-free convolution, where kernel positions that would fall outside the input are instead filled by the nearest valid neighbor. Substituting the interior window into the \convnn\ aggregation with uniform weights recovers the standard centered 1D convolution: 
\begin{equation}
    y_{i,c'} = \sum_{k=1}^{R} \sum_{c=1}^{C}
    V[I_{i,k},\, c] \cdot w_{k,c,c'}
    = \sum_{p=-\lfloor R/2\rfloor}^{\lfloor R/2\rfloor} \sum_{c=1}^{C}
    X_{i+p,\, c} \cdot W_{p,c,c'}.
\end{equation}

For odd kernel sizes $R$, a natural integer center exists and the interior window is exactly symmetric with $\lfloor R/2 \rfloor$ neighbors on each side. For even $R$, no integer center exists, so the $K = R$ nearest neighbors include an asymmetric split (e.g., $R/2 - 1$ to the left and $R/2$ to the right, depending on tie-breaking in $\argkmax_k$). The \convnn\ equivalence to standard centered convolution therefore holds exactly for odd $R$, and holds up to a half-pixel asymmetry at the center for even $R$. 
 
\paragraph{2D case.}
Let positions be indexed $(i, j)$ with normalized coordinates $p_{(i,j)} = (i/(H-1),\; j/(W-1)) \in [0,1]^2$. The negative exponential squared Euclidean similarity is:

\begin{equation}
    S_{(i,j),(i',j')}
    = \exp( -\left(\frac{i-i'}{H-1}\right)^2 - \left(\frac{j-j'}{W-1}\right)^2).
\end{equation}

For a kernel of size $R \times R$ ($K = R^2$), the $K$ highest-scoring neighbors of $(i, j)$ form an $R \times R$ spatial window centered at $(i, j)$ for interior positions: 

\begin{equation}
    I_{(i,j)} = \left\{(i+p,\; j+q) \;\middle|\;
    p, q \in \left\{-\left\lfloor\tfrac{R}{2}\right\rfloor, \dots,
                     \left\lfloor\tfrac{R}{2}\right\rfloor\right\}\right\}.
\end{equation}

At spatial boundaries (e.g., when $i=0$, $i=H-1$, $j=0$, or $j=W-1$) the 2D window shifts analogously to the 1D boundary case: the window remains anchored to the nearest valid positions along each axis independently, producing a right-skewed, left-skewed, top-skewed, or bottom-skewed window as appropriate. Substituting the interior window into the \convnn\ aggregation exactly recovers the standard centered 2D convolution from Section~\ref{subsec:std_conv}. 

\paragraph{Exact Recovery via Zero-Padding.}

The boundary skewing described above can be eliminated entirely by applying zero-padding to the input before the \convnn\ operation, exactly as done in standard convolution. For a 1D input $X \in \mathbb{R}^{N \times C}$, padding by $\lfloor R/2 \rfloor$ on each side yields a padded input $X_{\text{pad}} \in \mathbb{R}^{(N + 2\lfloor R/2 \rfloor) \times C}$. For a 2D input $\mathbf{X} \in \mathbb{R}^{H \times W \times C}$, padding by $\lfloor R/2 \rfloor$ zeros along each spatial boundary yields $\mathbf{X}_{\text{pad}} \in \mathbb{R}^{(H + 2 \lfloor R/2 \rfloor) \times (W + 2\lfloor R/2 \rfloor) \times C}$. The normalized spatial coordinates are then recomputed over the padded dimensions. 

Under this scheme, every query position has exactly $R$ (or $R^2$) valid neighbors within the padded input, so $\argkmax_k$ always recovers a fully symmetric centered window with no boundary skewing. The \convnn\ operation then produces an output of the same spatial dimensions as the padded input; removing the $\lfloor R/2 \rfloor$-wide border from the output recovers a result of the original spatial dimension $N$ (or $H \times W$), which is exactly equivalent to standard convolution with same-padding. For odd kernel sizes $R$ with a natural center, this construction recovers the convolutional operations from Sections~\ref{subsec:std_conv} and~\ref{subsec:dw_conv} exactly, with no approximation at any position. 

\paragraph{Code Implementation.}
To follow PyTorch's channel-first convention for convolution operations, the implementation uses channel-first layout $[B, C, N]$ for 1D inputs and $[B, C, H, W]$ for 2D inputs throughout. All internal permutations are handled within the module so that the input and output shapes match PyTorch's standard \textsc{Conv1d} and \textsc{Conv2d} interfaces directly.

The 1D implementation is given in Code~\ref{code:convnn_convolution_1d}. The input follows the channels-first convention $[B, C, N]$ and the output is $[B, C', N]$. A normalized coordinate channel $p_i = i/(N -1) \in [0, 1]$ serves as both query and key, while the value is the identity projection of the input features. Euclidean-distance neighbor selection with $\rho = \mathbf{1}_K$ (uniform weights) recovers the centered convolutional window established in Section~\ref{subsubsec:pe_mimic_conv}. Setting \texttt{padding}$=\lfloor R/2 \rfloor$ eliminates boundary skewing and exactly recovers standard or depthwise \textsc{Conv1d}. 

The 2D implementation is given in Code~\ref{code:convnn_convolution_2d} and follows the same procedure with two modifications. The input $[B, C, H, W]$ receives two appended coordinate channels encoding the normalized 2D positions $p^x_{(i,j)} = i/(H-1)$ and $p^y_{(i,j)} =j/(W-1)$, and the spatial dimensions are flattened to $[B, C+2, HW]$ before similarity computation. Neighbor selection proceeds identically to the 1D case with $K = R^2$, and padding removal requires an intermediate reshape to $[B, C, H_p, W_p, K]$ to trim the border along both spatial axes before aggregation. The output is unflattened from $[B, C', HW]$ back to $[B, C', H, W]$.

Since \texttt{torch.topk} returns the $K$ nearest neighbor indices ordered by descending similarity score rather than by spatial position, an additional \texttt{torch.sort} is applied to $I_i$ along the neighbor dimension. This reorders the gathered neighbors in ascending index order, restoring the spatial ordering of the convolutional kernel window established in Section~\ref{subsubsec:pe_mimic_conv} and ensuring that the \textsc{Conv1D} weights $w_{k, c, c'}$ correspond consistently to the same relative spatial offset $p = k - \lfloor R/2 \rfloor$ across all query positions.

\vspace{10pt}
\begin{lstlisting}[
    nolol, 
    basicstyle=\footnotesize\ttfamily,
    caption={\convnn\ instantiated as a 1D convolutional operation.}, 
    label={code:convnn_convolution_1d}, 
    captionpos=b
]
class ConvNNConv1D(nn.Module):
    def __init__(self, in_channels, out_channels, 
                       K, padding=0, convolution_type='standard'
                ):
        """
        Args:
            in_channels      : input channel dimension C
            out_channels     : output channel dimension C'
            K                : number of nearest neighbors (R)
            padding          : zero-padding per side (lfloor R/2 rfloor)
            convolution_type : 'standard' or 'depthwise'
        """
        super().__init__()
        assert convolution_type in ['standard', 'depthwise']
        self.K       = K
        self.padding = padding
        self.conv    = nn.Conv1d(
            in_channels=in_channels, out_channels=out_channels,
            kernel_size=K, stride=K,
            groups=1 if convolution_type == 'standard' else in_channels
        )
    
    def positional_encoding(self, X):
        # X: [B, C, N] -> coords: [B, C+1, N]
        # Normalized coordinates p_i = i / (N - 1) in [0, 1]
        B, C, N = X.shape 
        coords = torch.linspace(0, 1, steps = N, device=X.device)   # [N]
        coords_exp = coords.view(1, 1, N).expand(B, -1, -1)         # [B, 1, N]
        return torch.cat([X, coords_exp], dim=-2)                   # [B, C+1, N]

    def similarity(self, Q, K):
        # Negative exponential squared Euclidean distance: S_ij = exp(-||p_i - p_j||^2)
        Q_sq = torch.sum(Q ** 2, dim=-1, keepdim=True)       # [B, N, 1]
        K_sq = torch.sum(K ** 2, dim=-1, keepdim=True)       # [B, N, 1]
        dot  = torch.matmul(Q, K.transpose(-2, -1))          # [B, N, N]
        dist =  Q_sq + K_sq.transpose(-2, -1) - 2 * dot      # [B, N, N]
        return torch.exp(-torch.clamp(dist, min=0.0))        # higher = closer

    def forward(self, X):
        # X: [B, C, N]
        B, C, N = X.shape
        
        # Apply zero-padding along sequence dimension if specified
        if self.padding > 0:
            X = F.pad(X, (self.padding, self.padding), mode='constant', value=0) #[B, C, N+2p]

        B, C, N_p = X.shape

        # Step 1 - Similarity Computation
        X_coord = self.positional_encoding(X)                         # [B, C+1, N_p]
        Q =  K = X_coord[:, -1:, :].transpose(-2, -1)                 # [B, N_p, 1]
        V = X_coord[:, :-1, :]                                        # [B, C, N_p]
        S = self.similarity(Q, K)                                     # [B, N_p, N_p]

        # Step 2 - Neigbor Selection: I_i = argkmax_k(S)[i, :] in {0, ..., N_p-1}^K
        _, I = torch.topk(S, k=self.K, dim=-1, largest=True)          # [B, N_p, K]
        I, _ = torch.sort(I, dim=-1)                                  # [B, N_p, K]

        # rho(z) = 1_K (uniform weights, no softmax)
        # Gather V[I_i, :] in R^{K x C}
        # then expand to [B, C, N_p, K] for gather along dim=2
        V_exp = V.unsqueeze(-1).expand(B, C, N_p, self.K)             # [B, C, N_p, K]
        I_exp = I.unsqueeze(1).expand(B, C, N_p, self.K)              # [B, C, N_p, K]
        X_nn  = torch.gather(V_exp, dim=2, index=I_exp)               # [B, C, N_p, K]

        # Remove padded positions after gathering
        if self.padding > 0:
            X_nn = X_nn[:, :, self.padding:-self.padding, :]          # [B, C, N, K]

        # Step 3 - Weighted Aggregation: Reshape to [B, C, N*K] for Conv1D with kernel_size=K, stride=K
        X_nn = X_nn.permute(0, 1, 3, 2).reshape(B, C, -1)             # [B, C, N*K]
        Y    = self.conv(X_nn)                                        # [B, C', N]
        return Y
\end{lstlisting}
\vspace{10pt}

\begin{lstlisting}[
    nolol, 
    basicstyle=\footnotesize\ttfamily,
    caption={\convnn\ instantiated as a 2D convolutional operation.}, 
    label={code:convnn_convolution_2d}, 
    captionpos=b
]
class ConvNNConv2D(nn.Module):
    def __init__(self, in_channels, out_channels, 
                       K, H, W, padding=0, convolution_type='standard'
                ):
        """
        Args:
            in_channels      : input channel dimension C
            out_channels     : output channel dimension C'
            K                : number of nearest neighbors (R^2 for R x R kernel)
            H, W             : spatial dimension of the input feature map
            padding          : zero-padding per side (lfloor R/2 rfloor)
            convolution_type : 'standard' or 'depthwise'
        """
        super().__init__()
        assert convolution_type in ['standard', 'depthwise']
        self.K       = K
        self.H       = H
        self.W       = W
        self.padding = padding
        self.conv    = nn.Conv1d(
            in_channels=in_channels, out_channels=out_channels,
            kernel_size=K, stride=K,
            groups=1 if convolution_type == 'standard' else in_channels
        )
        self.flatten = nn.Flatten(start_dim=2)          # [B, C, H, W] -> [B, C, HW]
        self.unflatten = nn.Unflatten(dim=2, unflattened_size=(H, W))
    
    def positional_encoding(self, X):
        # X: [B, C, H, W] -> coords: [B, C+2, H, W]
        # Normalized coordinates p_i = i / (H - 1), p_j = j / (W - 1) in [0, 1]
        B, C, H, W = X.shape 
        y_coords = torch.linspace(0, 1, steps = H, device=X.device)    # [H]
        x_coords = torch.linspace(0, 1, steps = W, device=X.device)    # [W]
        y_grid, x_grid = torch.meshgrid(y_coords, x_coords, indexing='ij')
        grid = torch.stack((x_grid, y_grid), dim=0).unsqueeze(0).expand(B, -1, -1, -1) # [B, 2, H, W]                       
        return torch.cat([X, grid], dim=1)                             # [B, C+2, H, W]

    def similarity(self, Q, K):
        # Negative exponential squared Euclidean distance: S_ij = exp(-||p_i - p_j||^2)
        Q_sq = torch.sum(Q ** 2, dim=-1, keepdim=True)      # [B, HW, 1]
        K_sq = torch.sum(K ** 2, dim=-1, keepdim=True)      # [B, HW, 1]
        dot  = torch.matmul(Q, K.transpose(-2, -1))         # [B, HW, HW]
        dist =  Q_sq + K_sq.transpose(-2, -1) - 2 * dot     # [B, HW, HW]
        return torch.exp(-torch.clamp(dist, min=0.0))       # higher = closer

    def forward(self, X):
        # X: [B, C, H, W]
        B, C, H, W = X.shape
        
        # Apply zero-padding along sequence dimension if specified
        if self.padding > 0:
            X = F.pad(X, (self.padding,) * 4, mode='constant', value=0) #[B, C, H+2p, W+2p]

        B, C, H_p, W_p = X.shape
        HW_p = H_p * W_p
        
        # Step 1 - Similarity Computation
        X_coord = self.positional_encoding(X)                    # [B, C+2, H_p, W_p]
        X_coord = self.flatten(X_coord)                          # [B, C+2, H_p*W_p]
        Q =  K = X_coord[:, -2:, :].transpose(-2, -1)            # [B, HW_p, 2]
        V = X_coord[:, :-2, :]                                   # [B, C, HW_p]
        S = self.similarity(Q, K)                                # [B, HW_p, HW_p]

        # Step 2 - Neigbor Selection: I_i = argkmax_k(S)[i, :] in {0, ..., HW_p-1}^K
        _, I = torch.topk(S, k=self.K, dim=-1, largest=True)     # [B, HW_p, K]
        I, _ = torch.sort(I, dim=-1)                             # [B, HW_p, K]

        # rho(z) = 1_K (uniform weights, no softmax)
        # Gather V[I_i, :] then expand to [B, C, HW_p, K] for gather along dim=2
        V_exp = V.unsqueeze(-1).expand(B, C, HW_p, self.K)       # [B, C, HW_p, K]
        I_exp = I.unsqueeze(1).expand(B, C, HW_p, self.K)        # [B, C, HW_p, K]
        X_nn  = torch.gather(V_exp, dim=2, index=I_exp)          # [B, C, HW_p, K]

        # Remove padded positions after gathering
        if self.padding > 0:
            X_nn = X_nn.view(B, C, H_p, W_p, self.K)
            X_nn = X_nn[:, :, 
                        self.padding:-self.padding, 
                        self.padding:-self.padding, :]           # [B, C, H, W, K]
            X_nn = X_nn.reshape(B, C, H*W, self.K)               # [B, C, HW, K]

        # Step 3 - Weighted Aggregation: Reshape to [B, C, N*self.K] for Conv1D with kernel_size=K, stride=K
        X_nn = X_nn.permute(0, 1, 3, 2).reshape(B, C, -1)        # [B, C, HW*K]
        Y    = self.conv(X_nn)                                   # [B, C', HW]
        Y = self.unflatten(Y)                                    # [B, C', H, W]
        return Y
\end{lstlisting}
\vspace{-7pt}

\subsection{Hybrid Branching Layer}
\label{subsec:hybrid_branching_layer}

\begin{figure}[H]
    \vspace{7pt}
    \centering
    \includegraphics[width=0.9\linewidth]{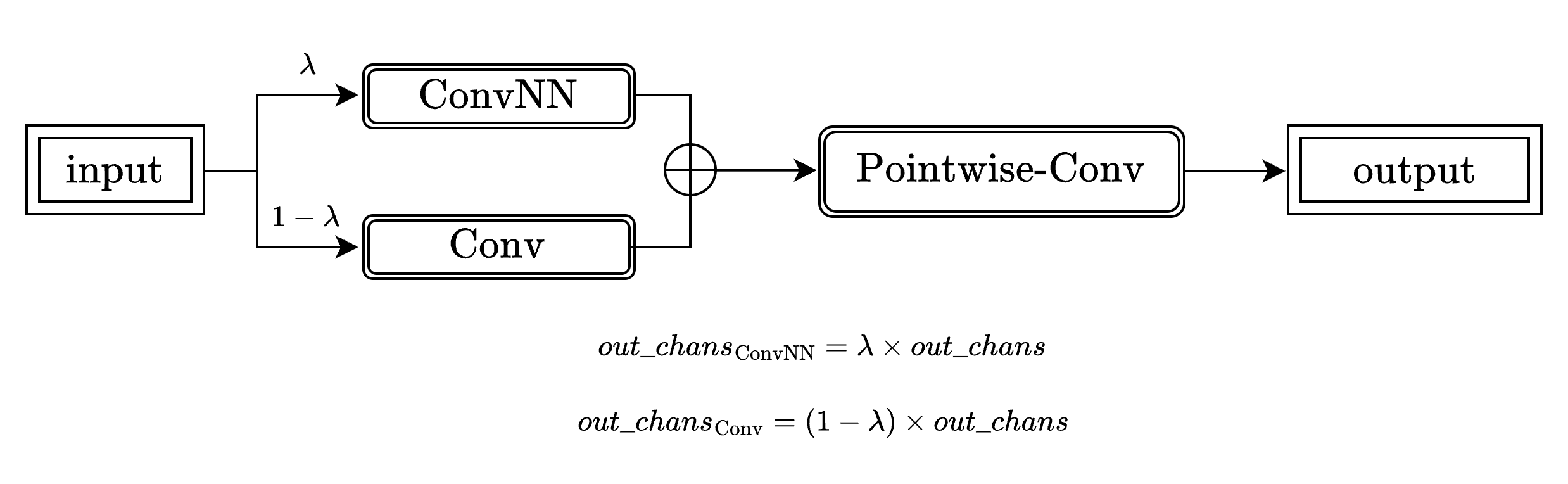}
    \caption{Hybrid Branching Layer combining \convnn\ and standard convolution with branch ratio $\lambda$.}
    \label{fig:hybrid_branching_layer}
\end{figure}

A key distinction between \convnn\ and standard convolution operations lies in their receptive behavior. Standard convolutions aggregate information from a fixed spatially local window, providing a strong local inductive bias that has proven highly effective for low-level feature extraction in computer vision. \convnn, by contrast, can select neighbors based on feature similarity rather than spatial proximity, enabling global context to influence each representation regardless of spatial distance. 

Building on prior work that integrates local and global feature modeling (\cite{szegedy2015going, peng2021conformer, chen2022regionvit}), we introduce a \textit{hybrid branching layer} that processes the input through two parallel branches — one convolutional and one \convnn\ — and fuses their outputs via a pointwise ($1 \times 1$) convolution as illustrated in Figure~\ref{fig:hybrid_branching_layer}. This design is analogous to the multi-scale parallel branch structure of the Inception module (\cite{szegedy2015going}), but replaces the multi-kernel convolution branches with a local-global pair. 

A scalar parameter $\lambda \in [0, 1]$ controls the proportion of output channels allocated to the \convnn\ branch. Specifically, the \convnn\ branch produces $\lfloor \lambda \cdot C' \rfloor$ output channels and the convolutional branch produces $C' - \lfloor \lambda \cdot C' \rfloor$ output channels, where $C'$ is the total output channel count. Both branches take the same input and use the same kernel size (or equivalently, neighbor count $K = R^2$ for 2D), keeping parameter counts comparable while allowing $\lambda$ to regulate the relative contribution of local and global processing. The two branch outputs are concatenated along the channel dimension and passed through a $1 \times 1$ pointwise convolution to mix channels and produce the final output of size $C'$. 

The \convnn\ branch uses the identity projection ($Q = K = V =  \mathbf{X}$), cosine similarity as the similarity metric, and uniform weights ($\rho = \mathbf{1}_K$) with zero-padding as in Code~\ref{code:convnn_convolution_2d}. No sorting of top-$K$ indices is required since the convolutional equivalence proof does not apply here as the neighbors are selected by feature similarity rather than spatial proximity. The layer is agnostic to input dimensionality: for 1D inputs $[B, C, N]$ it uses \texttt{ConvNNConv1D} and \texttt{nn.Conv1d}; for 2D inputs $[B, C, H, W]$ it uses \texttt{ConvNNConv2D} and \texttt{nn.Conv2d}. 

\vspace{10pt}
\begin{lstlisting}[nolol, 
    basicstyle=\footnotesize\ttfamily,
    caption={Hybrid Branching Layer combining \convnn\ and convolution. Both branches accept channels-first inputs and support standard and depthwise variations.}, 
    label={code:hybrid_branching_layer}, captionpos=b]
class HybridBranchingLayer(nn.Module):
    def __init__(self, in_channels, out_channels, K, branch_ratio=0.5):
        """
        Args:
            in_channels  : input channel dimension C
            out_channels : total output channel dimension C'
            K            : kernel size R for convolution; neighbor count for ConvNN
            branch_ratio : lambda in [0,1], fraction of C' from ConvNN branch
        """
        super().__init__()
        assert 0.0 <= branch_ratio <= 1.0
        self.out_ch_convnn = int(out_channels * branch_ratio)
        self.out_ch_conv   = out_channels - self.out_ch_convnn

        # Branch 1: ConvNN (standard, depthwise, or depthwise-separable)
        self.branch_convnn = ConvNNConv(in_channels, self.out_ch_convnn, K, ...)

        # Branch 2: Convolution (standard, depthwise, or depthwise-separable)
        self.branch_conv   = Conv(in_channels, self.out_ch_conv, K, ...)

        # Pointwise convolution to mix concatenated branch outputs
        self.pointwise     = Conv(out_channels, out_channels, kernel_size=1, stride=1, ...)

    def forward(self, x):
        # x: [B, C, N] for 1D  |  [B, C, H, W] for 2D
        x1  = self.branch_convnn(x)          # [B, out_ch_convnn, N] or [B, ., H, W]
        x2  = self.branch_conv(x)            # [B, out_ch_conv, N]   or [B, ., H, W]
        out = torch.cat([x1, x2], dim=1)     # [B, C', N]            or [B, C', H, W]
        out = self.pointwise(out)            # [B, C', N]            or [B, C', H, W]
        return out
\end{lstlisting}

\section{Connection between \convnn, Self-Attention, and KVT-Attention}
\label{sec:convnn_attn_connection}
We now show that self-attention is a special case of \convnn. Given an input $X \in \mathbb{R}^{N \times d_{\text{model}}}$, we set the query, key, and value projections to learned linear layers exactly as in Section~\ref{subsec:self_attention}:
\begin{equation}
    Q = XW^Q, \quad K = XW^K, \quad V = XW^V,
    \qquad W^Q, W^K, W^V \in \mathbb{R}^{d_{\text{model}} \times d_k}.
\end{equation}

The similarity matrix is the scaled dot product: 
\begin{equation}
    S = \frac{QK^\top}{\sqrt{d_k}} \in \mathbb{R}^{N \times N}
\end{equation}

Applying $\kmax_k$ and $\argkmax_k$ to $S$ yields the top-$K$ similarity values and neighbor indices for each query: 
\begin{equation}
    \mathbf{s}_i = \kmax_k(S)[i, :] \in \mathbb{R}^{K}, \qquad
    I_i = \argkmax_k(S)[i, :] \in \{0, \dots, N-1\}^{K}.
\end{equation}

Setting $\rho = \operatorname{softmax}$, the neighborhood matrix for each query $\mathbf{x}_i$ is: 

\begin{equation}
    X_{\text{nn},i}
    = \operatorname{diag}\!\left(\operatorname{softmax}(\mathbf{s}_i)\right)
      V[I_i, :] \in \mathbb{R}^{K \times d_k}.
\end{equation}

Concatenating over all queries gives $X_{\text{nn}} \in \mathbb{R}^{(KN) \times d_k}$. Applying a depthwise \textsc{Conv1D} with kernel size $K$ and stride $K$ yields:
\begin{equation}
    y_i
    = \operatorname{DepthwiseConv1D}(\operatorname{diag}\!\left(\operatorname{softmax}(\mathbf{s}_i)\right)
      V[I_i, :]) \in \mathbb{R}^{d_k}.
\end{equation}

This depthwise convolution expands element wise as:

\begin{equation}
    y_{i,m} = \sum_{k=1}^{K}
    \operatorname{softmax}(\mathbf{s}_i)_k
    \cdot V[I_{i,k},\, m]
    \cdot w_{k,m},
\end{equation}

\noindent where $w_{k, m}$ is the $k$-th kernel weight for output channel $m$. When $K = N$ (all key elements are neighbors) and the depthwise convolution weights are fixed to uniform equal to 1 ($w_{k, m} = 1$ for all $k, m$), this reduces exactly to standard self-attention: 
\begin{equation}
    y_{i,m} = \sum_{k=1}^{N}
    \operatorname{softmax}(\mathbf{s}_i)_k \cdot V[I_{i,k},\, m] \cdot 1,
\end{equation}

\noindent which is directly analogous to $y_i = \sum_{j=1}^{N} a_{i,j} v_j$ from Equation~\eqref{eq:self_attention_partition}, with $a_{i,j} = \operatorname{softmax}(\mathbf{s}_i)_k$ and $v_j = V[I_{i,k}, m]$, where $j = I_{i,k}$ is the index of the $k$-th nearest neighbor of $\mathbf{x}_i$.

\paragraph{KVT-Attention.}
When $K < N$ with fixed unit depthwise convolution weights ($w_{k, m}$, \convnn\ exactly recovers the $K$-NN attention formulation of KVT (\cite{DBLP:conf/eccv/WangWWLCLJ22}), demonstrating that \convnn\ subsumes both standard self-attention and its sparse variants as special cases. 

In KVT (\cite{DBLP:conf/eccv/WangWWLCLJ22}), for the $i$-th query $q_i$, the $K$ most similar keys are first identified to form the sparse neighbor sets $\mathcal{N}_i^k \subset \{k_1, \dots, k_N\}$ and the corresponding value set $\mathcal{N}_i^v$. The attention weights over these $K$ neighbors are computed as: 

\begin{equation}
    A_i = \operatorname{softmax}\!\left(
    \frac{\langle q_i,\; (k_{j_1}, \dots, k_{j_K}) \rangle}{\sqrt{d_k}}
    \right), \qquad k_{j_l} \in \mathcal{N}_i^k,
\end{equation}

\noindent where $\langle q_i, k_{j_l} \rangle = q_i^\top k_{j_l}$ denotes the dot product between the query and its $l$-th nearest key neighbor. Gathering the corresponding value neighbors using indices $\mathcal{N}_i^v$ to form $V_{\text{knn}} \in \mathbb{R}^{K \times d_k}$, the output at position $i$ is: 

\begin{equation}
    y_i = \sum_{l=1}^{K} A_{i,l} \cdot v_{j_l},
    \qquad v_{j_l} \in \mathcal{N}_i^v.
\end{equation}

This is precisely recovered by \convnn\ when $K < N$ with unit depthwise convolution weights ($w_{k,m} = 1$), since: 

\begin{equation}
    y_{i,m} = \sum_{k=1}^{K}
    \operatorname{softmax}(\mathbf{s}_i)_k
    \cdot V[I_{i,k},\, m] \cdot 1,
\end{equation}

\noindent where $I_{i, k}$ indexes the $k$-th nearest neighbor of $\mathbf{x}_i$, exactly matching the KVT formulation above. 

\begin{figure}[H]
    \centering
    \vspace{10pt}
    \includegraphics[width=0.8\linewidth]{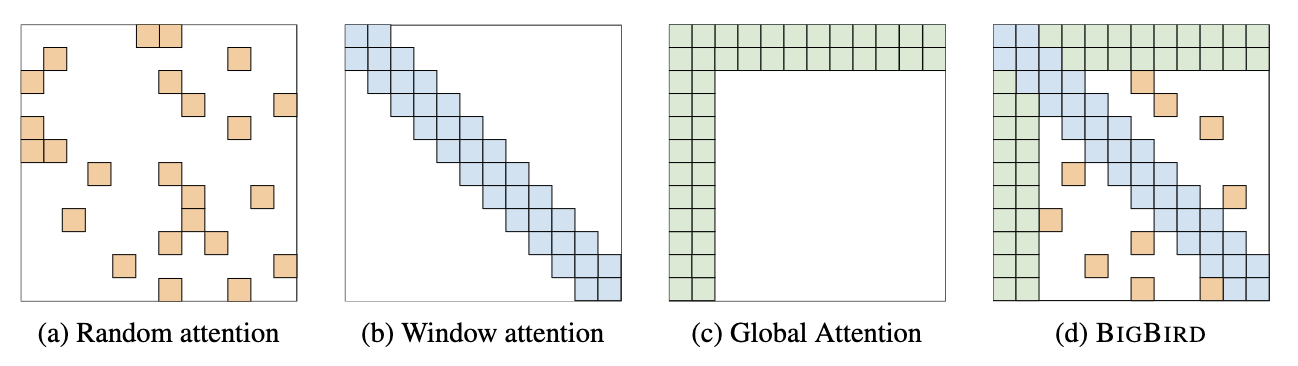}
    \caption{Attention patterns described in BigBird (\cite{zaheer2020big}),
             combining local windowed, global, and random attention.}
    \label{fig:bigbird_attention}
\end{figure}

\begin{figure}[H]
    \centering
    \vspace{10pt}
    \includegraphics[width=0.8\linewidth]{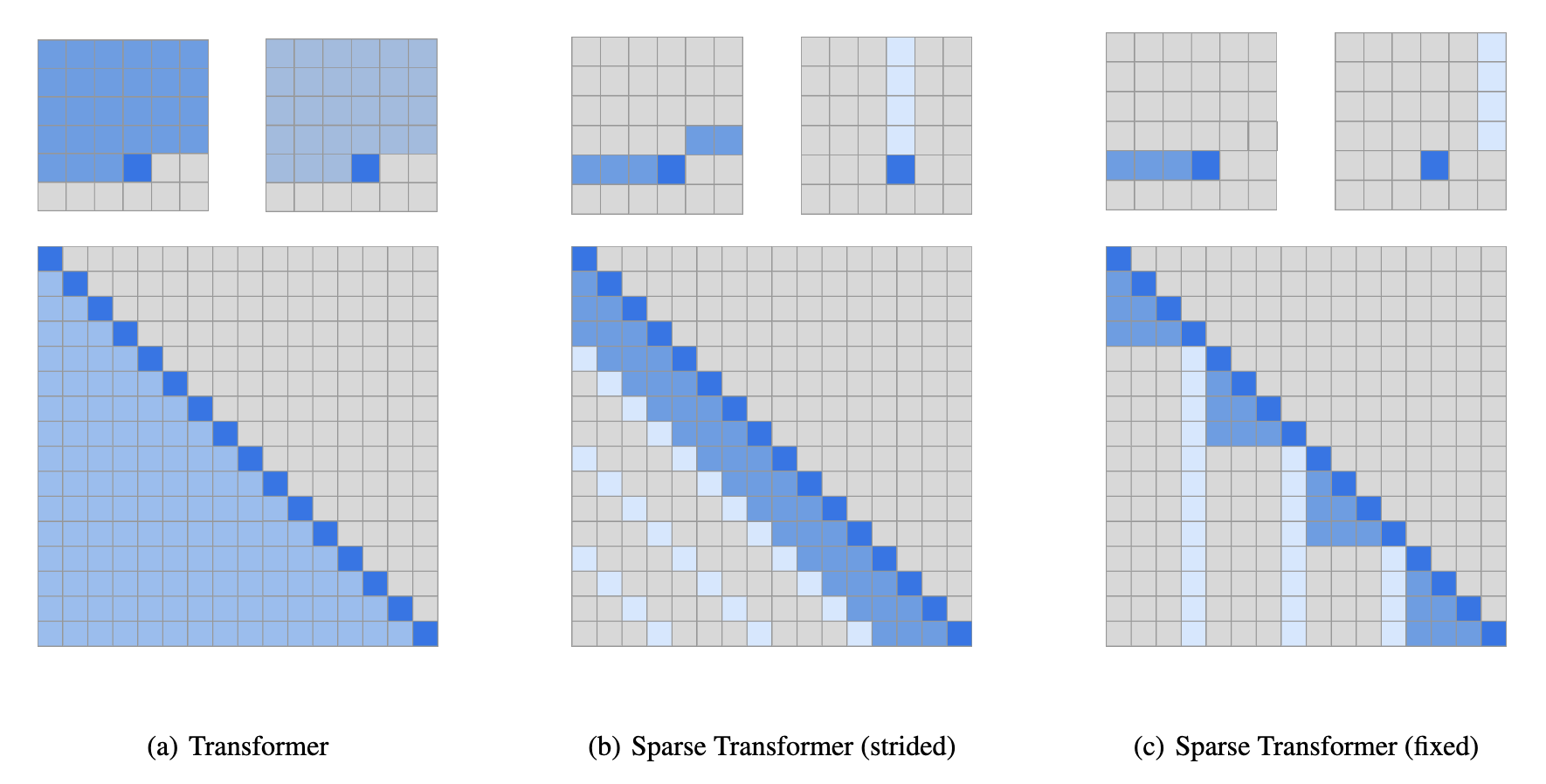}
    \caption{Factorized sparse attention patterns from the Sparse
             Transformer (\cite{child2019generating}).}
    \label{fig:sparse_transformer}
\end{figure}

\paragraph{Sparse Attention Variants.}
Beyond KVT-attention, \convnn\ can generalize a broad class of sparse and locally-constrained attention patterns by applying a mask to the similarity matrix $S = QK^\top$ before neighbor selection. Formally, given a mask $M \in \{0, -\infty\}^{N \times N}$ that encodes which positions are permitted to attend to one another, the masked similarity matrix is: 

\begin{equation}
    S' = S + M,
\end{equation}

\noindent after which $\argkmax_k(S')$ selects only from the unmasked entries. This single modification allows a wide range of sparse attention designs. As illustrated in Figure~\ref{fig:bigbird_attention}, BigBird attention (\cite{zaheer2020big}) combines local sliding-window attention, global token attention, and random attention within a single mask pattern. The Sparse Transformer (\cite{child2019generating}), shown in Figure~\ref{fig:sparse_transformer}, employs a strided and fixed factorized attention pattern to reduce the full $O(N^2)$ attention matrix to a sparse subset of interactions. Both designs, along with other sparse-local attention variants (\cite{aguilera2024local}), can be expressed within \convnn\ by specifying the corresponding mask $M$, demonstrating that \convnn\ provides a unified interface for the full spectrum from dense global attention to structured sparse attention. 

\paragraph{Causal (Masked) Self-Attention}
In particular, causal self-attention (Section~\ref{par:causal_self_attention}) is recovered as a direct special case by setting $M_{ij} = -\infty$ for all $j > i$ and $M_{ij} = 0$ otherwise, which is precisely the upper triangular mask described in Equation~\ref{eq:causal_attention_matrix}. This demonstrates that \convnn generalizes the autoregressive attention as a structured masking patten within the same unified framework. 

\subsection{Extension of \convnn\ to Multi-Head Operation in Multi-Head Attention}
\label{subsec:multihead_convnn}
The single-head \convnn\ formulation described above operates on the full $d_k$-dimensional projection space. To extend \convnn\ to the multi-head setting of Section~\ref{subsec:multihead_attention}, we adopt the standard head-splitting strategy: the projected tensors $Q, K, V \in \mathbb{R}^{N \times d_k}$ are partitioned along the channel dimension into $H$ heads, producing $Q_H, K_H, V_H \in \mathbb{R}^{H \times N \times d_k/H}$. Rather than running $H$ independent \convnn\ operations sequentially, we fold the head dimension into the batch dimension, yielding tensors of shape $\mathbb{R}^{BH \times N \times d_k/H}$, and apply the \convnn\ operation once across the combined batch. The similarity matrix, neighbor selection, and weighted aggregation all proceed identically as in the single-head case, but now independently for every head within the expanded batch. After aggregation, the $BH$ dimension is split back to recover the head dimension $\mathbb{R}^{B \times H \times N \times d_k/H}$, and the standard \texttt{combine\_heads} operation reshapes this to $\mathbb{R}^{B \times N \times d_k}$, followed by the output projection $W^O$. 

Consider a concrete example with batch size $B = 1$. After the $Q, K, V$ projections, the the query, key, and value are of shape $\mathbb{R}^{1 \times N \times d_k}$. After head splitting, it is now of shape $\mathbb{R}^{1 \times H \times N \times d_k/H}$. Folding the head into the batch dimension produces $\mathbb{R}^{H \times N \times d_k/H}$, on which \convnn\ operates identically to the single-head case. The output is then reshaped back to $\mathbb{R}^{1 \times H \times N \times d_k/H}$, the heads are combined to recover $\mathbb{R}^{1 \times N \times d_k}$, and $W^O$ produces the final output $\mathbb{R}^{N \times d_{\text{model}}}$. This procedure is mathematically equivalent to running $H$ independent \convnn\ operations in parallel and concatenating the results, and generalizes directly to arbitrary batch sizes $B > 1$. 

The helper functions used to implement the head-batch folding are as follows: 

\vspace{10pt}
\begin{lstlisting}[
    nolol, 
    basicstyle=\footnotesize\ttfamily,
    caption={Head-splitting and batch-folding utilities for extending \convnn\ to the multi-head setting.}, 
    label={code:convnn_head_utils}, 
    captionpos=b
]
# Standard split/combine heads from multi-head attention
def split_heads(x, num_heads):
    # x: [B, N, d_k] -> [B, H, N, d_k/H]
    B, N, d_k = x.size()
    return x.view(B, N, num_heads, d_k // num_heads).transpose(-2, -1)

def combine_heads(x):
    # x: [B, H, N, d_k/H] -> [B, N, d_k]
    B, H, N, d_k_h = x.size()
    return x.transpose(-2, -1).contiguous().view(B, N, H * d_k_h)

# ConvNN-specific: fold head dimension into batch for ConvNN operation
def combine_batch_head(x):
    # x: [B, H, N, d_k/H] -> [B*H, N, d_k/H] 
    B, H, N, d_k_h = x.size()
    return x.reshape(B * H, N, d_k_h)

def split_batch_head(x, num_heads):
    # x: [B*H, d_k/H, N] -> [B, H, N, d_k/H]
    BH, d_k_h, N = x.size()
    B = BH // num_heads
    return x.permute(0, 2, 1).contiguous().view(B, num_heads, N, d_k_h)

def merge_attn_matrix(x, num_heads):
    # x: [B, H, N, N] -> [B*H, N, N]
    B, H, N, _ = x.size()
    return x.reshape(B * H, N, N)
\end{lstlisting}
\vspace{-5pt}

Together with the equivalence established in Sections~\ref{sec:convnn_conv_connection} and~\ref{sec:convnn_attn_connection}, this demonstrates that \convnn\ is a strict generalization of both convolutional and attention-based operations, recovering each as a special case through appropriate choices of similarity function, positional encoding, weighting function $\rho$, and aggregation kernel. 

\subsection{Code for \convnn\ Self-Attention.}
Self-attention (\cite{vaswani2017attention}) is recovered exactly by setting $K = N$ (all tokens are neighbors), using a depthwise \textsc{Conv1D} with fixed unit weights and no bias, and applying scaled dot-product similarity. Under these conditions, the weighted aggregation step reduces to a standard softmax-weighted sum over all value vectors, which is precisely Equation~\eqref{eq:self_attention_partition}. The PyTorch implementation of \convnn\ as self-attention is given in Code~\ref{code:convnn_self_attention}. 

To extend this instantiation to multi-head self-attention, the $Q, K, V$ projections are first split into $H$ heads using \texttt{split\_heads}, producing tensors of shape $\mathbb{R}^{B \times H \times N \times d_k/H}$. The similarity matrix $S$ is computed per-head and the resulting attention matrix is merged into the batch dimension via \texttt{merge\_attn\_matrix}. The value tensor $V$ is folded into the batch dimension via \texttt{combine\_batch\_head}, and all subsequent \convnn\ operations (neighbor selection, modulation, and aggregation) proceed on the combined $\mathbb{R}^{BH \times N \times d_k/H}$ tensor. The output is then unfolded via \texttt{split\_batch\_head}, the heads are recombined with \texttt{combine\_heads}, and the final output projection $W^O$ is applied, exactly recovering multi-head self-attention as in Equation~\eqref{eq:multihead_attention}. 

\vspace{10pt}
\begin{lstlisting}[
    nolol, 
    basicstyle=\footnotesize\ttfamily,
    caption={\convnn\ instantiated as self-attention. Setting $K = N$, fixed unit depthwise weights, and scaled dot-product similarity exactly recovers standard self-attention.}, 
    label={code:convnn_self_attention}, 
    captionpos=b
]
class ConvNNSelfAttention(nn.Module):
    def __init__(self, d_model, d_k, N):
        """
        Args:
            d_model : input and output channel dimension
            d_k     : projected query/key/value dimension
            N       : sequence length (K is fixed to N)
        """
        super().__init__()
        self.K    = N           # K = N: all tokens are neighbors
        self.C_qk = d_k
        self.f_Q  = nn.Linear(d_model, d_k)
        self.f_K  = nn.Linear(d_model, d_k)
        self.f_V  = nn.Linear(d_model, d_k)
        self.rho  = nn.Softmax(dim=-1)
        self.conv = nn.Conv1d(   # depthwise, fixed unit weights, no bias
            in_channels=d_k,
            out_channels=d_k,
            kernel_size=N,
            stride=N,
            bias=False,
            groups=d_k
        )
        nn.init.constant_(self.conv.weight, 1.0)
        self.conv.weight.requires_grad = False
        
    def similarity(self, Q, K):
        # Scaled dot-product: S = Q K^T / sqrt(d_k) in R^{N x N}
        scale = torch.sqrt(torch.tensor(self.C_qk, dtype=Q.dtype, device=Q.device))
        return Q @ K.transpose(-2, -1) / scale

    def forward(self, X):
        # X: [B, N, C]
        B, N, C = X.shape

        # Step 1 - Similarity Computation
        # Q, K: [B, N, C_qk],   V: [B, N, C_v]
        Q, K, V = self.f_Q(X), self.f_K(X), self.f_V(X)
        S = self.similarity(Q, K)               # S: [B, N, N]

        # Step 2 - Neighbor Selection and Modulation
        # s_i = kmax_k(S)[i, :],  I_i = argkmax_k(S)[i, :]
        s, I = torch.topk(S, k=self.K, dim=-1)  # s, I: [B, N, K]

        # rho(s_i): weighting function - softmax or uniform
        s = self.rho(s)                         # [B, N, K]

        # Gather V[I_i, :] in R^{K x C_v}
        # V: [B, N, C_v] -> transpose to [B, C_v, N]
        # then expand to [B, C_v, N, K] for gather along dim=2
        C_v   = V.shape[-1]
        V_t   = V.permute(0, 2, 1)                           # [B, C_v, N]
        V_exp = V_t.unsqueeze(-1).expand(B, C_v, N, self.K)  # [B, C_v, N, K]
        I_exp = I.unsqueeze(1).expand(B, C_v, N, self.K)     # [B, C_v, N, K]
        V_nn  = torch.gather(V_exp, dim=2, index=I_exp)      # [B, C_v, N, K]

        # X_nn_i = diag(rho(s_i)) * V[I_i, :]
        s_exp = s.unsqueeze(1).expand(B, C_v, N, self.K)     # [B, C_v, N, K]
        X_nn  = s_exp * V_nn                                 # [B, C_v, N, K]

        # Step 3 - Weighted Aggregation
        # Reshape to [B, C_v, N*K] for Conv1D with kernel_size=K, stride=K
        X_nn = X_nn.permute(0, 1, 3, 2).reshape(B, C_v, -1)  # X_nn: [B, C_v, N*K]
        Y    = self.conv(X_nn)                               # Y: [B, C', N]
        Y    = Y.permute(0, 2, 1)                            # Y: [B, N, C']

        return Y
\end{lstlisting}

\subsection{Code for \convnn\ as KVT-Attention.}
KVT-attention (\cite{DBLP:conf/eccv/WangWWLCLJ22}) is recovered by relaxing $K < N$, restricting the softmax-weighted aggregation to only the $K$ most similar tokens rather than the full sequence. All other components, fixed unit depthwise weights, no bias, and scaled dot-product similarity, remain identical to the self-attention instantiation above. The sole architectural difference is therefore the value of $K$ passed at construction time. The PyTorch implementation of \convnn\ as KVT-attention is given in Code~\ref{code:convnn_kvt_attention}.

The same multi-head extension applies directly to the KVT instantiation. The procedure is identical to that described above for multi-head self-attention (split heads, fold into batch, run \convnn\ with $K < N$ and fixed unit weights, unfold, combine heads, and apply $W^O$) with the sole difference that neighbor selection is restricted to the $K$ most similar tokens per query rather than the full sequence. 

\vspace{10pt}
\begin{lstlisting}[
    nolol, 
    basicstyle=\footnotesize\ttfamily,
    caption={\convnn\ instantiated as KVT-attention (\cite{DBLP:conf/eccv/WangWWLCLJ22}). Identical to the self-attention instantiation except $K < N$, restricting aggregation to the $K$ most similar tokens per query.}, 
    label={code:convnn_kvt_attention}, 
    captionpos=b
]
class ConvNNKVTAttention(nn.Module):
    def __init__(self, d_model, d_k, K):
        """
        Args:
            d_model : input and output channel dimension
            d_k     : projected query/key/value dimension
            K       : number of nearest neighbors (K < N)
        """
        super().__init__()
        self.K    = K           
        self.C_qk = d_k
        self.f_Q  = nn.Linear(d_model, d_k)
        self.f_K  = nn.Linear(d_model, d_k)
        self.f_V  = nn.Linear(d_model, d_k)
        self.rho  = nn.Softmax(dim=-1)
        self.conv = nn.Conv1d(   # depthwise, fixed unit weights, no bias
            in_channels=d_k,
            out_channels=d_k,
            kernel_size=K,
            stride=K,
            bias=False,
            groups=d_k
        )
        nn.init.constant_(self.conv.weight, 1.0)
        self.conv.weight.requires_grad = False

    def similarity(self, Q, K):
        # Scaled dot-product: S = Q K^T / sqrt(d_k) in R^{N x N}
        scale = torch.sqrt(torch.tensor(self.C_qk, dtype=Q.dtype, device=Q.device))
        return Q @ K.transpose(-2, -1) / scale

    def forward(self, X):
        # X: [B, N, C]
        B, N, C = X.shape

        # Step 1 - Similarity Computation
        # Q, K: [B, N, C_qk],   V: [B, N, C_v]
        Q, K, V = self.f_Q(X), self.f_K(X), self.f_V(X)
        S = self.similarity(Q, K)               # S: [B, N, N]

        # Step 2 - Neighbor Selection and Modulation
        # s_i = kmax_k(S)[i, :],  I_i = argkmax_k(S)[i, :]
        s, I = torch.topk(S, k=self.K, dim=-1)  # s, I: [B, N, K]

        # rho(s_i): weighting function - softmax or uniform
        s = self.rho(s)                         # [B, N, K]

        # Gather V[I_i, :] in R^{K x C_v}
        # V: [B, N, C_v] -> transpose to [B, C_v, N]
        # then expand to [B, C_v, N, K] for gather along dim=2
        C_v   = V.shape[-1]
        V_t   = V.permute(0, 2, 1)                           # [B, C_v, N]
        V_exp = V_t.unsqueeze(-1).expand(B, C_v, N, self.K)  # [B, C_v, N, K]
        I_exp = I.unsqueeze(1).expand(B, C_v, N, self.K)     # [B, C_v, N, K]
        V_nn  = torch.gather(V_exp, dim=2, index=I_exp)      # [B, C_v, N, K]

        # X_nn_i = diag(rho(s_i)) * V[I_i, :]
        s_exp = s.unsqueeze(1).expand(B, C_v, N, self.K)     # [B, C_v, N, K]
        X_nn  = s_exp * V_nn                                 # [B, C_v, N, K]

        # Step 3 - Weighted Aggregation
        # Reshape to [B, C_v, N*K] for Conv1D with kernel_size=K, stride=K
        X_nn = X_nn.permute(0, 1, 3, 2).reshape(B, C_v, -1)  # X_nn: [B, C_v, N*K]
        Y    = self.conv(X_nn)                               # Y: [B, C', N]
        Y    = Y.permute(0, 2, 1)                            # Y: [B, N, C']

        return Y
\end{lstlisting}

\subsection{Code for \convnn\ as Multi-Head Self-Attention.}

\begin{lstlisting}[
    nolol, 
    basicstyle=\footnotesize\ttfamily,
    caption={\convnn\ with multi-head support. The head utility functions \texttt{split\_heads}, \texttt{combine\_heads}, \texttt{combine\_batch\_head}, \texttt{split\_batch\_head}, and \texttt{merge\_attn\_matrix} are defined in Code~\ref{code:convnn_head_utils}.}, 
    label={code:multihead_convnn_attention}, 
    captionpos=b
]
class MultiHeadConvNNAttention(nn.Module):
    def __init__(self, d_model, d_k, K, H):
        """
        Args:
            d_model : input and output channel dimension
            d_k     : projected query/key/value dimension (must be divisible by H)
            K       : number of nearest neighbors (K <= N)
            H       : number of attention heads
        """
        super().__init__()
        assert d_k % H == 0, "d_k must be divisible by H"
        self.K, self.H    = K, H
        self.d_k          = d_k
        self.d_k_h        = d_k // H          # per-head dimension
        self.f_Q          = nn.Linear(d_model, d_k)
        self.f_K          = nn.Linear(d_model, d_k)
        self.f_V          = nn.Linear(d_model, d_k)
        self.f_O          = nn.Linear(d_k, d_model)
        self.rho          = nn.Softmax(dim=-1)
        self.conv         = nn.Conv1d(         # depthwise, kernel=K, stride=K
            in_channels=self.d_k_h,
            out_channels=self.d_k_h,
            kernel_size=K,
            stride=K,
            bias=False,
            groups=self.d_k_h
        )
        
    def similarity(self, Q, K):
        # Scaled dot-product: S = Q K^T / sqrt(d_k) in R^{N x N}
        scale = torch.sqrt(torch.tensor(self.C_qk, dtype=Q.dtype, device=Q.device))
        return Q @ K.transpose(-2, -1) / scale

    def forward(self, X):
        # X: [B, N, d_model]
        B, N, _ = X.shape
        BH = B * self.H

        # Step 1 - Similarity Computation
        # Project and split heads using split_heads()
        # Q, K, V: [B, N, d_k] -> [B, H, N, d_k/H]
        Q = split_heads(self.f_Q(X), self.H)
        K = split_heads(self.f_K(X), self.H)
        V = split_heads(self.f_V(X), self.H)

        # S: [B, H, N, N] -> [B*H, N, N]
        S = self.similarity(Q, K)
        S = merge_attn_matrix(S, self.H)       # [B*H, N, N]

        # V: [B, H, N, d_k/H] -> [B*H, d_k/H, N]
        V = combine_batch_head(V)              # [B*H, d_k/H, N]

        # Step 2 - Neighbor Selection and Modulation
        # s_i = kmax_k(S)[i, :],  I_i = argkmax_k(S)[i, :]
        s, I = torch.topk(S, k=self.K, dim=-1) # s, I: [B*H, N, K]
        s = self.rho(s)                        # softmax(s_i): [B*H, N, K]

        # Gather V[I_i, :] in R^{K x d_k/H}
        V_exp = V.unsqueeze(-1).expand(BH, self.d_k_h, N, self.K)  # [B*H, d_k/H, N, K]
        I_exp = I.unsqueeze(1).expand(BH, self.d_k_h, N, self.K)   # [B*H, d_k/H, N, K]
        V_nn  = torch.gather(V_exp, dim=2, index=I_exp)            # [B*H, d_k/H, N, K]

        # X_nn_i = diag(rho(s_i)) * V[I_i, :]
        s_exp = s.unsqueeze(1).expand(BH, self.d_k_h, N, self.K)   # [B*H, d_k/H, N, K]
        X_nn  = s_exp * V_nn                                       # [B*H, d_k/H, N, K]

        # Step 3 - Weighted Aggregation
        # Reshape to [B*H, d_k/H, N*K] for Conv1D with kernel_size=K, stride=K
        X_nn = X_nn.permute(0, 1, 3, 2).reshape(BH, self.d_k_h, -1)# [B*H, d_k/H, N*K]
        Y    = self.conv(X_nn)                                     # [B*H, d_k/H, N]

        # Recover head dimension and combine using split_batch_head() and combine_heads() from Code~\ref{code:convnn_head_utils}
        Y = split_batch_head(Y, self.H)                            # [B, H, N, d_k/H]
        Y = combine_heads(Y)                                       # [B, N, d_k]

        # Final output projection W^O
        Y = self.f_O(Y)                                            # [B, N, d_model]
        return Y
\end{lstlisting}

%% file: Chapters/Chapter3.tex

\chapter{Fast Convolutional Nearest Neighbors}
\label{Chapter3} 

\section{Fast \convnn with Custom Triton Kernels}
\label{sec:fast_convnn}

\subsection{Motivation and Overview}
\label{subsec:fast_convnn_overview}

The standard \convnn formulation introduced in Chapter~\ref{Chapter2} incurs two principal computational costs at training and inference time. First, computing the full $N \times N$ similarity matrix requires $O(N^2)$ time and space in terms of the input length $N$, which becomes prohibitive for long sequences or larger feature maps. Second, the three-step pipeline (gather neighbors from $V$, apply attention weights $\rho(\mathbf{s}_i)$, and apply the \textsc{Conv1D} kernel) requires materializing several intermediate tensors of shape $[B, N, K, C]$ in GPU memory, creating significant memory pressure during the backward pass. 

To address these costs, we introduce a fused Triton kernel (\cite{10.1145/3315508.3329973}) that merges the gather, attention weighting, and depthwise or standard \textsc{Conv1D} aggregation into a single GPU kernel launch. This eliminates the large intermediate tensors, reduces memory bandwidth usage, and allows the backward pass to recompute required values on the fly rather than storing them. The implementation provides custom forward and backward passes via \texttt{torch.autograd.Function}, making it a transparent drop-in replacement for the standard \convnn\ aggregation step with no change to the surrounding model code. 

\subsection{Fused Kernel Design}
\label{subsec:triton_kernel_design}

The core insight behind the fused kernel is that the gather, weighting, and convolution steps in \convnn aggregation are all reductions over the $K$ neighbor dimension and can therefore be expressed as a single loop over $K$ that accumulates into an output vector. For each query position $i \in \{1, \dots, N\}$ in each batch/batch-head slice $b \in \{1, \dots, B\}$, the kernel computes: 

\begin{equation}
    y_{b, i, c} = \sum_{k=1}^{K}
    \rho(\mathbf{s}_{b, i})_k \cdot
    V_b[I_{b, i, k}, c] \cdot
    W_{c,k},
\end{equation}

\noindent where $I_{b, i, k}$ is the index of the $k$-th nearest neighbor position $i$, $\rho(\mathbf{s}_{b, i})_k$ is the corresponding softmax attention weight, and $W_{c, k}$ is the depthwise convolution weight for channel $c$ and kernel position $k$. For standard convolution, the per-channel weight $W \in \mathbb{R}^{C \times K}$ is replaced by a cross-channel weight matrix $\mathbf{W} \in \mathbb{R}^{C' \times C \times K}$ and the sum additionally reduces over the input channel dimension $C$: 

\begin{equation}
    y_{b,i,c'} = \sum_{k=1}^{K} \sum_{c=1}^{C}
    \rho(\mathbf{s}_{b,i})_k \cdot
    V_b[I_{b, i, k}, c] \cdot
    \mathbf{W}_{c',c,k}.
\end{equation}

The kernel is launched on a 1D grid of $B \times N$ program instances, where $B$ is either the regular batch or $B = B \cdot H$ for the merged batch-head of the multi-head functionality. Each program instance handles all $C$ channels for a single $(b, i)$ pair via a vectorized load-accumulate loop over $K$, using a channel block size $\texttt{BLOCK\_C} = 2^{\lceil \log_2 C \rceil}$ to align memory accesses to powers of two. The accumulator is maintained in \texttt{float32} regardless of input precision and cast back to the input type (fp16, bf16, or fp32) before the store, ensuring numerical stability under automatic mixed precision training. 

Codes~\ref{code:fast_dw_conv} and ~\ref{code:fast_std_conv} provide the forward kernel implementations for the depthwise and standard convolution variants, respectively. Both use the same grid and accumulation structure and only differ in the weight load pattern and the reduction axis. The depthwise kernel loads a scalar weight $W_{c, k}$ per channel per neighbor, while the standard kernel loads a row $W_{c', :, k}$ and performs a dot product across $C$. 

\subsubsection{Backward Pass}
Custom backward passes are implemented for both variants using \texttt{torch.autograd.Function}. The backward pass receives the output gradient $\nabla \mathbf{Y} \in \mathbb{R}^{B \times N \times C}$ and computes the gradients with respect to $V$, $\rho(\mathbf{s})$ (the softmax attention weights), and $W$ (the convolution weight). Since the neighbor indices $I$ are discrete and selected by $\text{argmax}$, their gradient is \texttt{None}. Gradients are accumulated back to their original $V$ positions via \texttt{scatter\_add\_}, which correctly handles the case where a single position in $V$ is selected as a neighbor by multiple queries. For the standard convolution backward, the three gradient terms are computed efficiently using \texttt{einsum}, avoiding explicit loops over $K$ and reducing the implementation to three batched matrix contractions over the $b$, $i$, $k$, $c$, and $c'$ indices. 

\subsubsection{Integration with \convnn}
Code~\ref{code:fast_std_conv} shows how the fused kernel slots into the \convnn\ forward pass described in Section~\ref{sec:core_framework}. The standard similarity computation (Step 1) and $\text{top-}K$ neighbor selection (Step 2) proceed identically to the original implementation. After applying $\rho = \operatorname{softmax}$ to the top-$K$ scores $\mathbf{s}_i$, the batch and head dimensions are merged to form the $B$ axis matching the kernel's expected input layout $[B, N, \cdot]$, and the fused function replaces the gather-weight-conv sequence of Step 3. The output $\mathbf{Y} \in \mathbb{R}^{B \times N \times C}$ is returned in the same shape as the original \convnn\ output, so no downstream code requires modification. 

\subsection{Speed and Memory Benchmarks}
\label{subsec:fast_convnn_benchmark}

\begin{table}[H]
    \centering
    \vspace{5pt}
    \caption{Speed and memory benchmark comparing standard PyTorch
         and Triton-fused \convnn\ aggregation. Configuration:
         batch size = 32, sequence length = 197, $d_{\text{hidden}} =
         768$, NH = 12, $K = 8$, depthwise convolution.}
    \begin{tabularx}{\linewidth}{Xccc}
        \toprule
        \textbf{Metric} & \textbf{Original} & \textbf{Triton} & \textbf{Improvement} \\
        \midrule
        Forward time (ms)         & 4.92   & 4.24   & $1.16\times$ faster \\
        Backward time (ms)        & 9.29   & 6.84   & $1.36\times$ faster \\
        Forward peak memory (MB)  & 795.01 & 341.95 & $2.32\times$ lower  \\
        Backward peak memory (MB) & 904.88 & 657.54 & $1.38\times$ lower  \\
        \bottomrule
    \end{tabularx}
    \label{tab:fastconvnn_speed}
    \vspace{10pt}
\end{table}

Table~\ref{tab:fastconvnn_speed} reports forward and backward pass timings and peak memory usage for a representative configuration (batch size = 32, sequence length = 197, $d_{\text{hidden}} = 768$, $\text{NH} =  12$, $K = 9$) on a single H100 GPU. The Triton implementation achieves 1.16$\times$ faster forward pass, 1.36$\times$ faster backward pass, 2.32$\times$ lower forward peak memory, and 1.38$\times$ lower backward peak memory compared to the standard PyTorch implementation. The memory reduction is the more impactful gain in practice, as the standard implementation must store the full $[B, N, K, C]$ gather buffer for the backward pass, whereas the fused kernel recomputes this on the fly from the stored indices, values, and weights. 

\subsection{Numerical Correctness}
To verify that Triton implementation is numerically equivalent to the standard implementation, we compare the forward and backward outputs at batch size = 4, sequence length = 197, $d_{\text{hidden}} = 768$, NH = 12, and $K = 8$. The maximum absolute forward difference is $0.0 \times 10^{-7}$, confirming bit-exact agreement on the forward pass. The backward pass involves floating point reductions in a different order and produces small numerical differences: the maximum input gradient difference is $1.5 \times 10^{-7}$, the maximum convolution weight gradient difference is $2.86 \times 10^{-6}$, and the maximum query weight gradient difference is $2.62 \times 10^{-6}$. These differences are well within the tolerance of standard mixed-precision training and have no measurable effect on model performance, as confirmed by the matching ImageNet-1K results reported in Section~\ref{subsec:imagenet_vit}.

\subsection{Code for Fast \convnn}
\begin{lstlisting}[
    nolol, 
    basicstyle=\footnotesize\ttfamily,
    caption={Fused Triton kernel for \convnn\ aggregation with depthwise convolution. The forward kernel fuses neighbor gathering, attention weighting, and depthwise \textsc{Conv1D} into a single GPU kernel loop over $K$, accumulating in \texttt{float32} before casting to the input type.}, 
    label={code:fast_dw_conv}, 
    captionpos=b
]

import triton
import triton.language as tl
from torch.amp import custom_fwd, custom_bwd

@triton.jit
def fused_prime_depthwise_conv_fwd_kernel(
    # Pointers to matrices
    v_ptr, indices_ptr, values_ptr, weight_ptr, out_ptr,
    # Strides to handle memory layout
    stride_vb, stride_vt, stride_vd,
    stride_ib, stride_it, stride_ik,
    stride_wb, stride_wk,
    stride_ob, stride_ot, stride_od,
    # Matrix dimensions
    B_NH, T, D: tl.constexpr, K: tl.constexpr,
    # Meta-parameters
    BLOCK_D: tl.constexpr
):
    """
    Fuses the gathering of V, multiplication by Top-K attention weights, 
    and the depthwise convolution step.
    """
    pid_b_t = tl.program_id(0) # 1D grid covering Batch*Heads and Seq_len
    pid_b = pid_b_t // T
    pid_t = pid_b_t % T

    # Set up channel offsets
    d_offsets = tl.arange(0, BLOCK_D)
    mask_d = d_offsets < D

    # Initialize accumulator for the convolution sum
    acc = tl.zeros([BLOCK_D], dtype=tl.float32)

    # Loop over the Top-K elements
    for k in range(K):
        # 1. Load the index and attention value for the k-th top element
        idx_offset = pid_b * stride_ib + pid_t * stride_it + k * stride_ik
        v_idx = tl.load(indices_ptr + idx_offset)
        attn_val = tl.load(values_ptr + idx_offset)

        # 2. Load the V vector for all D channels at the gathered index
        v_offsets = pid_b * stride_vb + v_idx * stride_vt + d_offsets * stride_vd
        v_vec = tl.load(v_ptr + v_offsets, mask=mask_d, other=0.0)

        # 3. Load the Depthwise Convolution weight for this K step
        w_offsets = d_offsets * stride_wb + k * stride_wk
        w_vec = tl.load(weight_ptr + w_offsets, mask=mask_d, other=0.0)

        # 4. Multiply and accumulate (Gather * Attn_Value * Conv_Weight)
        acc += v_vec * attn_val * w_vec
        
    # Cast the fp32 accumulator back to the pointer's original dtype (fp16, bf16, or fp32)
    acc_casted = acc.to(v_ptr.dtype.element_ty) 
    
    # Store the final convolved output
    out_offsets = pid_b * stride_ob + pid_t * stride_ot + d_offsets * stride_od
    tl.store(out_ptr + out_offsets, acc_casted, mask=mask_d)

class FusedPrimeDepthwiseConvFunction(torch.autograd.Function):
    @staticmethod
    @custom_fwd(device_type='cuda') # Tell AMP to let this pass through natively
    def forward(ctx, v, topk_indices, topk_values, conv_weight):
        # Save tensors needed for the backward pass
        ctx.save_for_backward(v, topk_indices, topk_values, conv_weight)
        
        B_NH, T, D = v.shape
        _, _, K = topk_indices.shape
        
        # Ensure contiguous memory for predictable strides
        v = v.contiguous()
        topk_indices = topk_indices.contiguous()
        topk_values = topk_values.contiguous()
        weight = conv_weight.squeeze(1).contiguous() # Squeeze from (D, 1, K) to (D, K)
        
        out = torch.empty_like(v)
        
        # Grid computes one block per query token per batch/head
        grid = lambda meta: (B_NH * T, )
        BLOCK_D = triton.next_power_of_2(D)
        
        fused_prime_depthwise_conv_fwd_kernel[grid](
            v, topk_indices, topk_values, weight, out,
            v.stride(0), v.stride(1), v.stride(2),
            topk_indices.stride(0), topk_indices.stride(1), topk_indices.stride(2),
            weight.stride(0), weight.stride(1),
            out.stride(0), out.stride(1), out.stride(2),
            B_NH, T, D, K,
            BLOCK_D=BLOCK_D
        )
        return out

    @staticmethod
    @custom_bwd(device_type='cuda') # Tells AMP how to handle the backward pass
    def backward(ctx, grad_out):
        v, topk_indices, topk_values, conv_weight = ctx.saved_tensors
        B_NH, T, D = v.shape
        _, _, K = topk_indices.shape
        weight = conv_weight.squeeze(1) # (D, K)
        
        grad_out = grad_out.contiguous()

        # Flatten indices to (B_NH, T*K, 1) and expand to D channels
        idx_flat = topk_indices.view(B_NH, T * K, 1).expand(-1, -1, D)
        
        # Gather V directly to shape (B_NH, T*K, D), then reshape
        v_gathered = torch.gather(v, 1, idx_flat).view(B_NH, T, K, D)

        # Pre-compute shapes for broadcasting and CAST to grad_out's dtype (fp16/bf16)
        target_dtype = grad_out.dtype
        grad_out_exp = grad_out.unsqueeze(2)                
        val_exp = topk_values.unsqueeze(-1).to(target_dtype)                 
        weight_t_exp = weight.t().unsqueeze(0).unsqueeze(0).to(target_dtype) 
        v_gathered = v_gathered.to(target_dtype)

        # Gradient w.r.t topk_values (dVal) - Cast back to original dtype (fp32)
        grad_val = (grad_out_exp * v_gathered * weight_t_exp).sum(dim=-1).to(topk_values.dtype) 

        # Gradient w.r.t conv_weight (dW) - Cast back to original weight dtype (fp32)
        grad_weight_raw = (grad_out_exp * v_gathered * val_exp).sum(dim=(0, 1)) 
        grad_weight = grad_weight_raw.t().unsqueeze(1).to(weight.dtype) 

        # Gradient w.r.t V (dV)
        dv_gathered = grad_out_exp * val_exp * weight_t_exp 
        grad_v = torch.zeros_like(v)
        
        # Now both are guaranteed to be target_dtype (e.g., float16)
        grad_v.scatter_add_(1, idx_flat, dv_gathered.view(B_NH, T * K, D))

        # topk_indices is discrete, so its gradient is None.
        return grad_v, None, grad_val, grad_weight

\end{lstlisting}
\vspace{10pt}

\begin{lstlisting}[
    nolol, 
    basicstyle=\footnotesize\ttfamily,
    caption={Fused Triton Kernel for \convnn\ aggregation with standard convolution. Extends the depthwise variant by replacing the per-channel scalar weight $W_{d, k}$ with a cross-channel weight matrix $W_{d_{\text{out}},d_{\text{in}}, k}$, performing a matrix-vector product across input channels at each neighbor step.}, 
    label={code:fast_std_conv}, 
    captionpos=b
]

import triton
import triton.language as tl
from torch.amp import custom_fwd, custom_bwd

@triton.jit
def fused_prime_standard_conv_fwd_kernel(
    # Pointers to matrices
    v_ptr, indices_ptr, values_ptr, weight_ptr, out_ptr,
    # Strides to handle memory layout
    stride_vb, stride_vt, stride_vd,
    stride_ib, stride_it, stride_ik,
    stride_w_dout, stride_w_din, stride_w_k, # 3 Strides for (D_out, D_in, K)
    stride_ob, stride_ot, stride_od,
    # Matrix dimensions
    B_NH, T, D: tl.constexpr, K: tl.constexpr,
    # Meta-parameters
    BLOCK_D: tl.constexpr
):
    """
    Fuses the gathering of V, multiplication by Top-K attention weights, 
    and a STANDARD 1D convolution step across all channels.
    """
    pid_b_t = tl.program_id(0) # 1D grid covering Batch*Heads and Seq_len
    pid_b = pid_b_t // T
    pid_t = pid_b_t % T

    # Set up channel offsets for both output (d_out) and input (d_in) dimensions
    d_out = tl.arange(0, BLOCK_D)
    d_in = tl.arange(0, BLOCK_D)
    mask_out = d_out < D
    mask_in = d_in < D

    # Initialize accumulator for the output channels
    acc = tl.zeros([BLOCK_D], dtype=tl.float32)

    # Loop over the Top-K elements
    for k in range(K):
        # 1. Load the index and attention value for the k-th top element
        idx_offset = pid_b * stride_ib + pid_t * stride_it + k * stride_ik
        v_idx = tl.load(indices_ptr + idx_offset)
        attn_val = tl.load(values_ptr + idx_offset)

        # 2. Load the Input V vector for all D_in channels at the gathered index
        v_offsets = pid_b * stride_vb + v_idx * stride_vt + d_in * stride_vd
        v_vec = tl.load(v_ptr + v_offsets, mask=mask_in, other=0.0) # Shape: (D_in,)

        # 3. Load the Standard Convolution weight slice for this K step
        w_offsets = d_out[:, None] * stride_w_dout + d_in[None, :] * stride_w_din + k * stride_w_k
        w_slice = tl.load(weight_ptr + w_offsets, mask=(mask_out[:, None] & mask_in[None, :]), other=0.0) # Shape: (D_out, D_in)

        # 4. Multiply and accumulate (Matrix-Vector Product: W @ V * Attn)
        # Broadcasting v_vec across d_out handles the cross-channel mixing
        prod = w_slice * v_vec[None, :] * attn_val
        acc += tl.sum(prod, axis=1) # Sum over D_in (axis 1)
        
    # Cast the fp32 accumulator back to the pointer's original dtype (fp16, bf16, or fp32)
    acc_casted = acc.to(v_ptr.dtype.element_ty) 
    
    # Store the final convolved output
    out_offsets = pid_b * stride_ob + pid_t * stride_ot + d_out * stride_od
    tl.store(out_ptr + out_offsets, acc_casted, mask=mask_out)

class FusedPrimeStandardConvFunction(torch.autograd.Function):
    @staticmethod
    @custom_fwd(device_type='cuda')
    def forward(ctx, v, topk_indices, topk_values, conv_weight):
        ctx.save_for_backward(v, topk_indices, topk_values, conv_weight)
        
        B_NH, T, D = v.shape
        _, _, K = topk_indices.shape
        
        # Ensure contiguous memory for predictable strides
        v = v.contiguous()
        topk_indices = topk_indices.contiguous()
        topk_values = topk_values.contiguous()
        weight = conv_weight.contiguous() # (D_out, D_in, K)
        
        out = torch.empty_like(v)
        
        # Grid computes one block per query token per batch/head
        grid = lambda meta: (B_NH * T, )
        BLOCK_D = triton.next_power_of_2(D)
        
        fused_prime_standard_conv_fwd_kernel[grid](
            v, topk_indices, topk_values, weight, out,
            v.stride(0), v.stride(1), v.stride(2),
            topk_indices.stride(0), topk_indices.stride(1), topk_indices.stride(2),
            weight.stride(0), weight.stride(1), weight.stride(2), # 3 Strides mapped
            out.stride(0), out.stride(1), out.stride(2),
            B_NH, T, D, K,
            BLOCK_D=BLOCK_D
        )
        return out

    @staticmethod
    @custom_bwd(device_type='cuda') 
    def backward(ctx, grad_out):
        v, topk_indices, topk_values, conv_weight = ctx.saved_tensors
        B_NH, T, D = v.shape
        _, _, K = topk_indices.shape
        weight = conv_weight # (D_out, D_in, K)
        
        grad_out = grad_out.contiguous()

        # Flatten indices to (B_NH, T*K, 1) and expand to D channels
        idx_flat = topk_indices.view(B_NH, T * K, 1).expand(-1, -1, D)
        
        # Gather V directly to shape (B_NH, T, K, D_in)
        v_gathered = torch.gather(v, 1, idx_flat).view(B_NH, T, K, D)

        # Cast to target dtype for AMP stability
        target_dtype = grad_out.dtype
        v_gathered = v_gathered.to(target_dtype)
        topk_values = topk_values.to(target_dtype)
        weight = weight.to(target_dtype)

        # Pre-compute weighted V (Prime) -> (B_NH, T, K, D_in)
        v_prime = v_gathered * topk_values.unsqueeze(-1) 

        # EINSUM BACKWARD PASS (Much cleaner for standard convolution)
        # b: Batch, t: Sequence, k: Kernel, i: In_Channels, o: Out_Channels

        # 1. Gradient w.r.t conv_weight (dW) -> (D_out, D_in, K)
        grad_weight = torch.einsum('bto, btki -> oik', grad_out, v_prime).to(conv_weight.dtype)

        # 2. Gradient w.r.t v_prime (dv_prime) -> (B_NH, T, K, D_in)
        dv_prime = torch.einsum('bto, oik -> btki', grad_out, weight)

        # 3. Gradient w.r.t topk_values (dVal) -> (B_NH, T, K)
        grad_val = (dv_prime * v_gathered).sum(dim=-1).to(topk_values.dtype)

        # 4. Gradient w.r.t V (dV) before scattering
        dv_gathered_final = dv_prime * topk_values.unsqueeze(-1)
        
        # Ensure grad_v strictly matches target_dtype
        grad_v = torch.zeros_like(v, dtype=target_dtype)
        
        # Scatter add gradients back to original V locations
        grad_v.scatter_add_(1, idx_flat, dv_gathered_final.view(B_NH, T * K, D))

        # topk_indices is discrete, so its gradient is None.
        return grad_v, None, grad_val, grad_weight
\end{lstlisting}





%% file: Chapters/Chapter4.tex

\chapter{Proof of Concept Small Scale Experiments, Results, and Discussions} 

\label{Chapter4} 

\section{Initial Proof of Concept Validation on Small Scale Datasets}
\label{sec:small_scale_poc}
\subsection{Motivation for CIFAR Datasets in Architecture Validation}
Before large-scale evaluation on ImageNet-1K, we validate our framework on CIFAR-10 and CIFAR-100 to establish proof-of-concept and enable rapid iteration. These datasets strike a critical balance of being computationally tractable (enabling quick experimental cycles) while remaining sufficiently challenging to discern meaningful operational differences. CIFAR datasets have proven invaluable for early stage evaluation of novel attention mechanisms (\cite{touvron2021training, DBLP:conf/iccv/LiuL00W0LG21}), allowing researchers to validate hypotheses before committing to expensive large-scale training runs. 

\subsection{CIFAR-10 and CIFAR-100 Dataset Overview}
CIFAR-10 and CIFAR-100 were introduced by \citet{krizhevsky2009learning} as the Canadian Institute for Advanced Research datasets. Both consists of $32 \times 32$ RGB color images with 60,000 images split into 50,000 training and 10,000 test images. CIFAR-10 organizes these images into 10 semantically distinct categories (airplane, automobile, bird, cat, deer, dog, frog, horse, ship, truck), yielding 5,000 training and 1,000 test images per class. CIFAR-100 partitions the same 60,000 images into 100 fine-grained categories arranged into 20 superclasses of 5 classes each, reducing the per-class counts to 500 training and 100 test images. Despite their modest scale by modern standards, the CIFAR datasets remain valid benchmarks as the $32 \times 32$ resolution constraint demands that models learn globally coherent patterns, and the limited dataset size discourages memorization in favor of genuine feature learning. 

\begin{figure}[H]
\vspace{10pt}
    \begin{subfigure}[t]{0.45\linewidth}
        \centering
        \includegraphics[width=\linewidth]{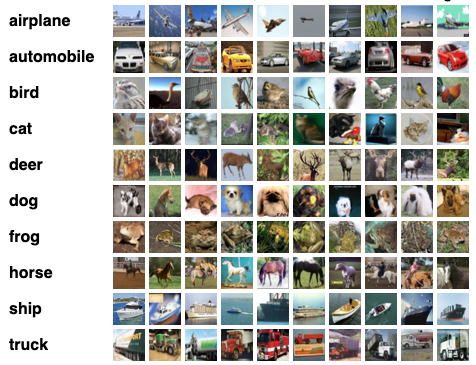}
        \caption{\centering CIFAR-10}
        \label{fig:cifar10}
    \end{subfigure}%
    \hfill
    \begin{subfigure}[t]{0.55\linewidth}
        \centering
        \includegraphics[width=\linewidth]{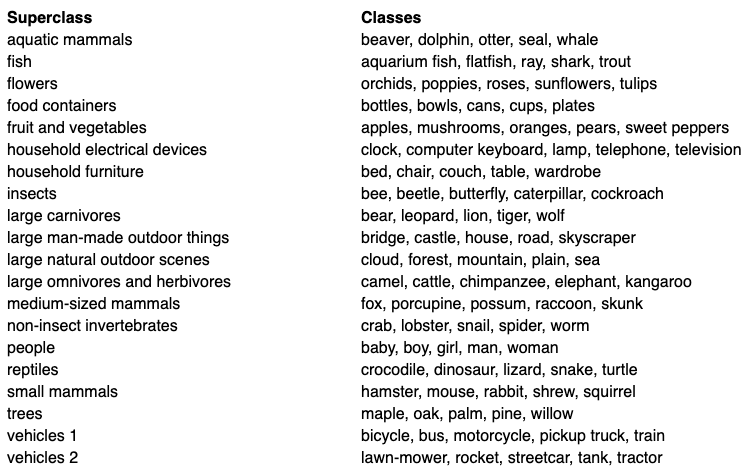}
        \caption{\centering CIFAR-100}
        \label{fig:cifar100}
    \end{subfigure}%
    \caption{CIFAR Dataset Classes and Sample Data}
    \label{fig:cifar_classes}
\end{figure}

\subsection{Experimental Design: Supporting Multiple Architectures}
A key advantage of CIFAR evaluation is the ability to fairly compare disparate architectures (CNNs vs. Vision Transformers) while isolating the contribution of the operational mechanisms. Our experimental protocol addresses the inherent architectural differences. 

\subsubsection{Preprocessing for Convolutional Networks}
For VGG-11 experiments, we maintain the native $32 \times 32$ image resolution. This choice preserves the spatial locality and local feature hierarchy that convolutional architectures exploit. At this resolution, \convnn operations perform neighbor selection in relatively dense feature maps, allowing us to isolate the benefits of adaptive spatial-feature fusion without confounding resolution effects. The small receptive field ensures that \convnn neighborhoods capture meaningfully related spatial regions. 

\subsubsection{Preprocessing for Vision Transformers}
For ViT-Tiny experiments, we resize all images to $224 \times 224$ resolution to align with the patchification mechanism of the Vision Transformer architecture. Vision Transformers divide images into non-overlapping $16 \times 16$ patches and flatten them into sequences; smaller resolution inputs yield fewer tokens (e.g., $32 \times 32$ produces only 4 patches in each dimension = 16 total patches), which may not provide sufficient sequence length for attention mechanisms to learn meaningful relationships. Resizing to $224 \times 224$ generates $(224/16)^2 = 196$ patches, enabling the transformer's positional encoding and multi-head attention to operate in a space where sequence modeling aspects become meaningful. 

\subsection{Normalization}
We apply per-channel standardization using dataset-specific statistics computed from the training set only to prevent data leakage into the test set. For CIFAR-10, we use $\mu = (0.4914, 0.4822, 0.4465)$ and $\sigma = (0.2470, 0.2435, 0.2616)$, and for CIFAR-100, $\mu = (0.5071, 0.4867, 0.4408)$ and $\sigma = (0.2675, 0.2565, 0.2761)$, where $\mu$ and $\sigma$ denote the per-channel mean and standard deviation computed over the RGB channels of the training set. Normalization is applied after all geometric transformations and immediately before the model's forward pass.

\subsection{Data Augmentation Policy}
We deliberately employ \textit{no data augmentation} beyond normalization (e.g., no random cropping, horizontal flipping, color jittering, mixup, or cutout). This controlled experimental design serves several purposes. 

\begin{enumerate}
    \item \textbf{Architectural Standardization:} Isolates the contribution of \convnn mechanisms within the two architectures without confounding regularization from data augmentation
    \item \textbf{Controlled Comparison:} Enables fair evaluation of \convnn relative to standard attention and convolutional baselines
    \item \textbf{Hypothesis Validation:} Tests whether our method provides intrinsic regularization benefits or merely complements standard augmentation
\end{enumerate}

We acknowledge that production-ready vision models benefit substantially from aggressive augmentation strategies (\cite{cubuk2020randaugment}). However, augmentation-free training is more principled for these small-scale proof-of-concept experiments. For the large-scale ImageNet-1K experiments, we reintroduce augmentation and follow the standard policy described in (\cite{DBLP:conf/iccv/LiuL00W0LG21}).

\subsection{Training Configuration}
\subsubsection{Optimization and Loss Function}
We employ a standard CrossEntropyLoss for all CIFAR experiments. Models are optimized with AdamW (\cite{loshchilov2017decoupled}) optimizer using learning rate $\text{lr} = 10^{-4}$, weight decay $\lambda = 0.01$, and default momentum parameters $\beta_1 = 0.9, \beta_2 = 0.999$. The learning rate of $10^{-4}$ is selected to enable stable convergence on modest-scale datasets, preventing optimization instabilities that can arise with larger learning rates on CIFAR's limited training data. 

\subsubsection{Learning Rate Schedule}
Unlike large-scale ImageNet-1K experiments, CIFAR experiments employ \textit{constant} learning rates without decay scheduling. This design choice reflects the smaller dataset size and limited training duration of 150 epochs. Learning rate schedules provide diminishing returns on short training runs, and constant rate provides a clearer baseline for comparing convolution and attention mechanisms without introducing scheduler-related variance. 

\subsubsection{Training Hyperparameters}
All CIFAR experiments are trained with a batch size of 256 for 150 epochs. Gradient norms are clipped to 1.0 to stabilize training. A dropout rate of 0.1 is applied to both standard and attention layers. All runs use a base random seed of 42, varied across runs to report performance. 

\subsubsection{Numerical Precision}
All experiments employ full Float32 precision. No mixed precision training or TensorFloat-32 optimizations are applied, ensuring numerically stable comparisons across all methods on small scale datasets where such optimization provide minimal computational benefit. 

\subsubsection{Hardware Configuration and Implementation}
Experiments are implemented in PyTorch 2.0 (\cite{pytorch2025}) and executed on single NVIDIA A100 PCIe (80 GB memory) GPUs. The large GPU memory enables batch size 256 with large intermediate feature maps, and the A100's 40 TFLOP FP32 compute ensures typical training completes within 4--6 hours per model. All experiments used fixed random seeds to ensure reproducibility, with seed values specified per experimental run.

\section{Results for CIFAR-10 and CIFAR-100 Datasets}
We evaluate \convnn\ against convolutional and attention-based baselines on the CIFAR-10 and CIFAR-100 classification as a proof-of-concept validation. These datasets provide a controlled setting for isolating the contributions of the \convnn\ formulation across two complementary architectural contexts: Convolutional Neural Networks and Vision Transformers. All experiments follow the training protocol described above in Section~\ref{sec:small_scale_poc}, with no data augmentation beyond normalization to ensure that observed differences reflect architectural properties rather than regularization effects. 

\subsection{\convnn\ as a Drop-in Replacement for Convolution}
\label{subsec:cifar_vgg}

\subsubsection{Model Setup}
For convolutional experiments, we use a VGG-11 backbone (\cite{DBLP:journals/corr/SimonyanZ14a}). To facilitate direct comparison with standard convolutions and isolate the effect of the neighbor selection mechanism, we set the $Q$, $K$, and $V$ projections to the identity function and fix $\rho(\mathbf{z}) = \mathbf{1}_K$ (uniform weights, no softmax):

\begin{equation}
    Q = K = V = X_{\text{flat}}, \qquad X_{\text{flat}} \in
    \mathbb{R}^{N \times C}, \quad N = H \cdot W.
\end{equation}

The similarity matrix is computed using cosine similarity over the flattened spatial features: 

\begin{equation}
    S_{ij} = \frac{q_i^\top k_j}{\|q_i\|\,\|k_j\|}
    = \frac{q_i^\top k_j}{\sqrt{q_i^\top q_i}\,\sqrt{k_j^\top k_j}},
    \qquad S \in \mathbb{R}^{N \times N}.
\end{equation}

Neighbor selection and aggregation then proceed as in Section~\ref{sec:convnn_conv_connection}, without positional encoding and without sorting of top-$K$ indices, since neighbor selection here is driven by feature similarity rather than spatial proximity. This configuration is identical to the \convnn\ branch used in the hybrid branching layer described in Section~\ref{subsec:hybrid_branching_layer}, where \convnn\ corresponds to $\lambda = 1$ and \textsc{Conv2D} corresponds to $\lambda = 0$; in both cases the pointwise $1 \times 1$ convolution at the output of the branching layer is retained. All experiments use $k = 9$ for \convnn\ and $3 \times 3$ kernels for \textsc{Conv2D} layers with sparse candidate search disabled, except where stated otherwise. 

\subsubsection{Hybrid Branching Results}

\begin{table}[H]
    \centering
    \footnotesize
    \setlength{\tabcolsep}{4pt}
    \vspace{5pt}    
    \begin{tabular}{clcccccc}
    \toprule
    & & & \multicolumn{2}{c}{CIFAR-10} & \multicolumn{2}{c}{CIFAR-100} & \\
    \cmidrule(lr){4-5} \cmidrule(lr){6-7}
    \textbf{Pos. Enc.} & \multicolumn{2}{c}{Layer Type} & Test Loss & Acc (\%) & Test Loss & Acc (\%) & GFLOPS \\
    \midrule
    
    \multirow{9}{*}{\textbf{Without}} 
    & \multicolumn{2}{c}{\textsc{Conv2D}} & 1.529 & 80.65 & 3.939 & 44.83 & \textbf{0.293} \\
    \cmidrule{2-8}
    & \multirow{7}{*}{$\lambda$} 
      & 0.125 & \textbf{1.393} & \textbf{81.10} & \textbf{3.905} & \textbf{45.08} & \multirow{7}{*}{0.325} \\
    & & 0.250 & \textbf{1.467} & \textbf{81.21} & \textbf{3.518} & \textbf{47.87} & \\
    & & 0.375 & \textbf{1.422} & \textbf{80.95} & \textbf{3.578} & \textbf{48.34} & \\
    & & 0.500 & \textbf{1.424} & \textbf{81.26} & \textbf{3.474} & \textbf{48.37} & \\
    & & 0.625 & \textbf{1.403} & \textbf{80.79} & \textbf{3.281} & \textbf{49.74} & \\
    & & 0.750 & \textbf{1.479} & 79.58 & \textbf{3.516} & \textbf{46.93} & \\
    & & 0.875 & \textbf{1.332} & 78.18 & \textbf{3.906} & \textbf{45.72} & \\
    \cmidrule{2-8}
    & \multicolumn{2}{c}{\convnn} & 2.694 & 54.00 & 5.011 & 29.50 & 0.325 \\
    
    \midrule
    
    \multirow{9}{*}{\textbf{With}} 
    & \multicolumn{2}{c}{\textsc{Conv2D}} & 1.529 & 80.65 & 3.939 & 44.83 & \textbf{0.293} \\
    \cmidrule{2-8}
    & \multirow{7}{*}{$\lambda$} 
      & 0.125 & \textbf{1.420} & \textbf{80.97} & \textbf{3.665} & \textbf{47.73} & \multirow{7}{*}{0.331} \\
    & & 0.250 & \textbf{1.447} & \textbf{80.91} & \textbf{3.702} & \textbf{47.65} & \\
    & & 0.375 & \textbf{1.386} & \textbf{82.08} & \textbf{3.696} & \textbf{48.39} & \\
    & & 0.500 & \textbf{1.213} & \textbf{81.77} & \textbf{3.588} & \textbf{48.80} & \\
    & & 0.625 & \textbf{1.341} & \textbf{80.81} & \textbf{3.323} & \textbf{49.23} & \\
    & & 0.750 & \textbf{1.283} & 80.26 & \textbf{3.533} & \textbf{48.78} & \\
    & & 0.875 & \textbf{1.225} & 78.70 & \textbf{3.539} & \textbf{46.26} & \\
    \cmidrule{2-8}
    & \multicolumn{2}{c}{\convnn} & 2.388 & 57.15 & 4.586 & 32.33 & 0.331 \\
    \midrule[1pt] 
    \multicolumn{2}{c}{\# Params.} &\multicolumn{2}{c}{130.015M} &\multicolumn{2}{c}{130.384M} \\

    \bottomrule 
    \end{tabular}
    \caption{Test loss and accuracy for VGG-11 on CIFAR-10 and CIFAR-100 with varying $\lambda$ in the hybrid branching module. Bold values indicate values that outperform those of the \textsc{Conv2D} baseline.}
    \label{tab:vary_lambda}
\end{table}

Table~\ref{tab:vary_lambda} reports test loss and accuracy on CIFAR-10 and CIFAR-100 when varying the branch ratio $\lambda$ in the hybrid branching module, both with and without positional encoding. When positional encoding is enabled, normalized coordinates in $[-1, 1]$ are appended along the channel dimension before computing the $Q$ and $K$ projections, yielding $X_{\text{pe}} \in \mathbb{R}^{H \times W \times (C+2)}$. This adds no learnable parameters as the \textsc{Conv1D} aggregation weights depend only on the channel dimension of $V \in \mathbb{R}^{N \times C}$, and the only additional cost is the marginal computation of the similarity matrix over the two extra coordinate channels. Despite the pure \convnn\ configuration ($\lambda = 1$) underperforming relative to the pure convolutional baseline ($\lambda = 0$), the hybrid branching model consistently surpasses both across most of the range $0 < \lambda < 1$. This effect is particularly pronounced on CIFAR-100, where the branching model improves accuracy by nearly 5\% over the \textsc{Conv2D} baseline. Adding positional encoding to the $Q$ and $K$ projections yields further, albeit modest, gains across both datasets, with the improvement being most noticeable in the pure \convnn\ ($\lambda = 1$) configuration.

\subsubsection{Training Dynamics}

\begin{figure}[H]
    \vspace{10pt}
    \centering
    \includegraphics[width=1.0\linewidth]{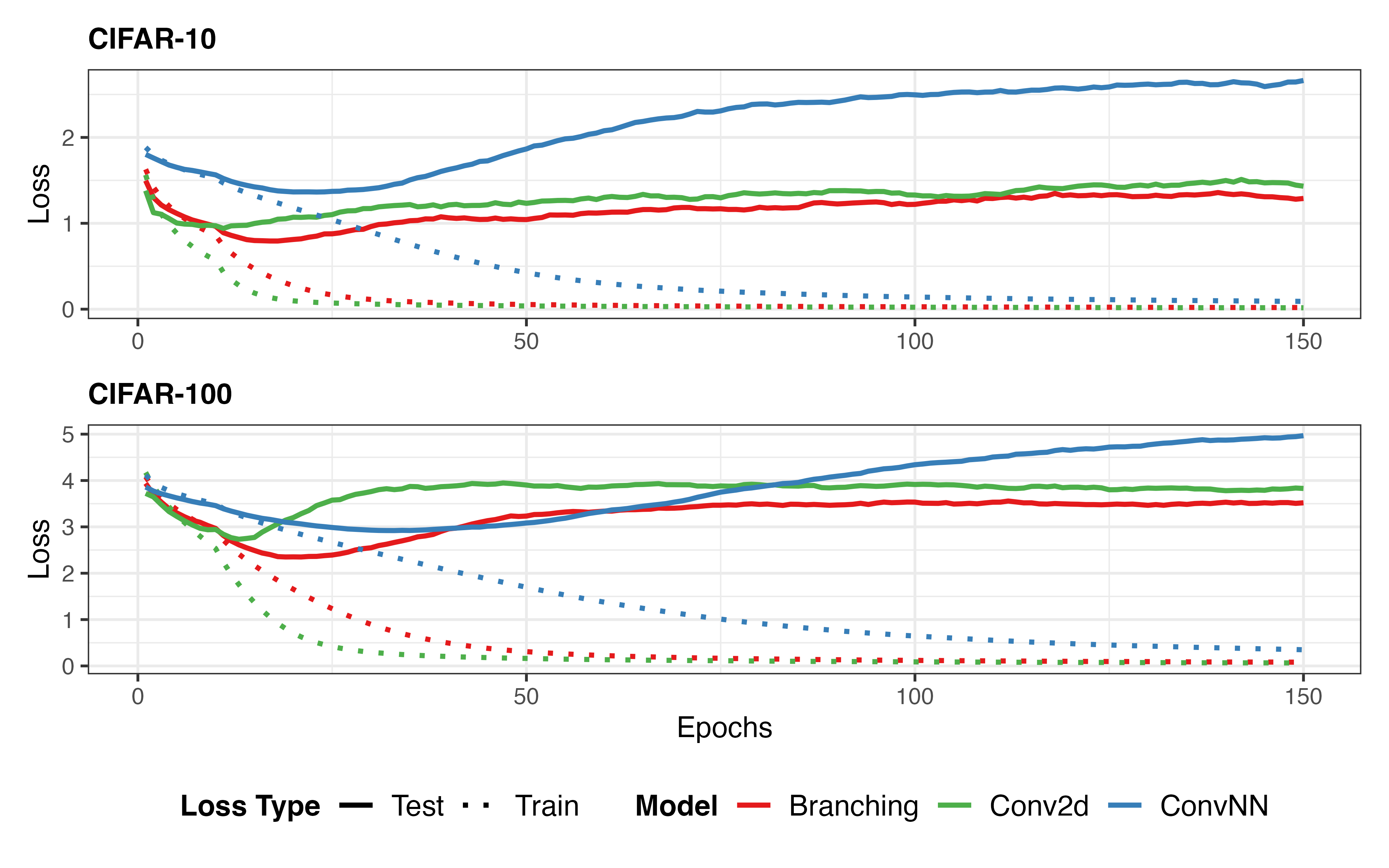}
    \vspace{-10pt}
    \caption{Training and test loss for VGG-11 on CIFAR-10 and CIFAR-100. Branching uses $\lambda=0.5$. All curves are smoothed with a moving average of window size 10.}
    \label{fig:vgg_branching}
\end{figure}

Figure~\ref{fig:vgg_branching} presents the training and test loss trajectories for VGG-11 on CIFAR-10 and CIFAR-100 using \textsc{Conv2D}, \convnn, and the hybrid branching layer with $\lambda = 0.5$, where similarities are computed from raw input features and no positional encodings. The curve reveal an important qualitative difference showing \textsc{Conv2D} rapidly converging early followed by overfitting, while the hybrid branching configuration demonstrates greater resistance to overfitting, achieves lower overall loss around epoch 40, and converges to the best final performance. The fact that the branching model outperforms both of its constituent branches individually underscores the complementarity of local and global feature selection of convolutional branch capturing fine-grained spatial patterns while \convnn\ branch introducing global context. Their combination provides a more robust inductive bias than either alone.

\subsubsection{Learning Rate Sensitivity Analysis}
\label{subsubsec:vgg_lr_analysis}

\begin{figure}[H]
    \centering
    \begin{subfigure}[t]{.5\linewidth}
        \centering
        \includegraphics[width=\linewidth]{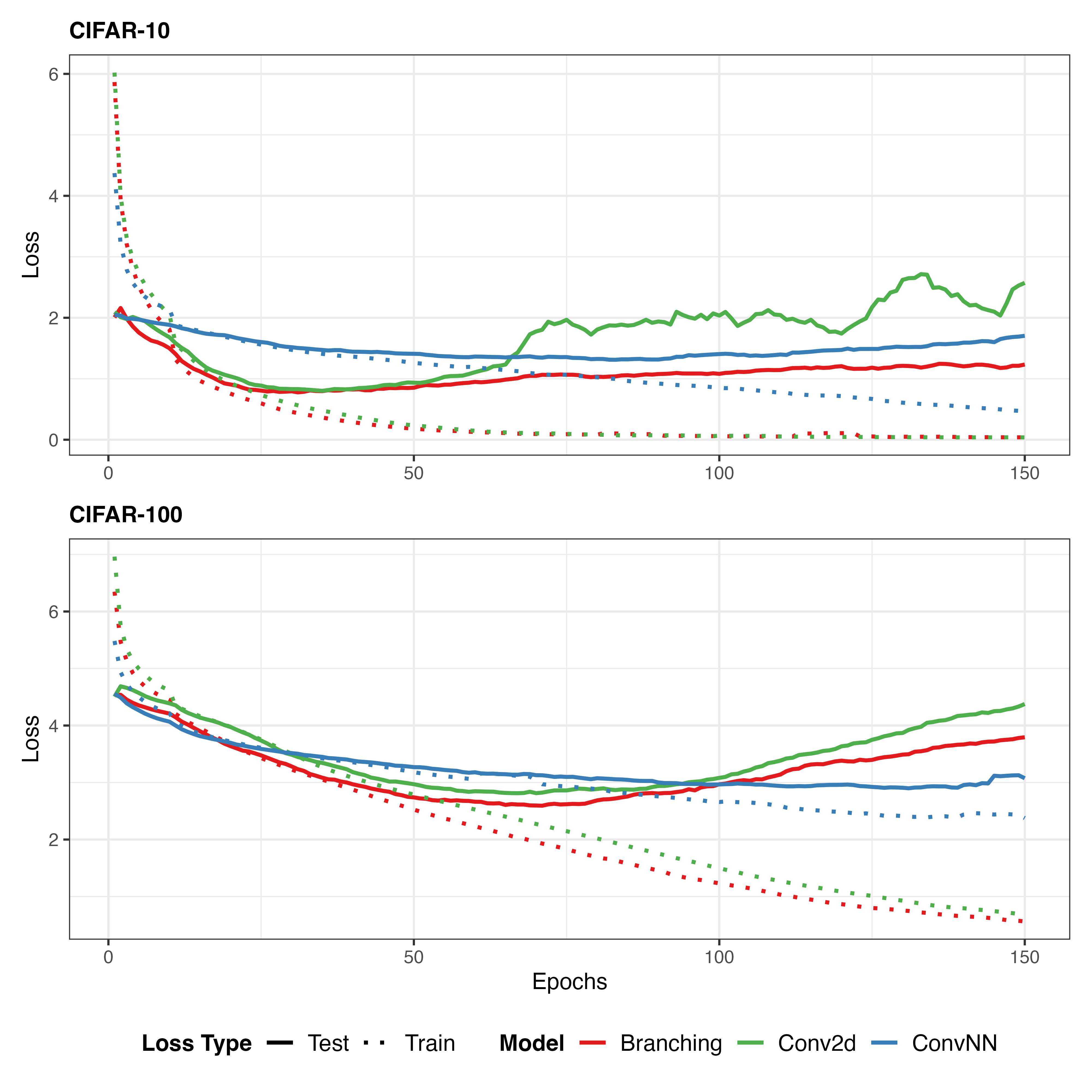}
        \caption{$lr = 10^{-3}$}
        \label{fig:vgg_loss_1e3}
    \end{subfigure}%
    \begin{subfigure}[t]{.5\linewidth}
        \centering
        \includegraphics[width=\linewidth]{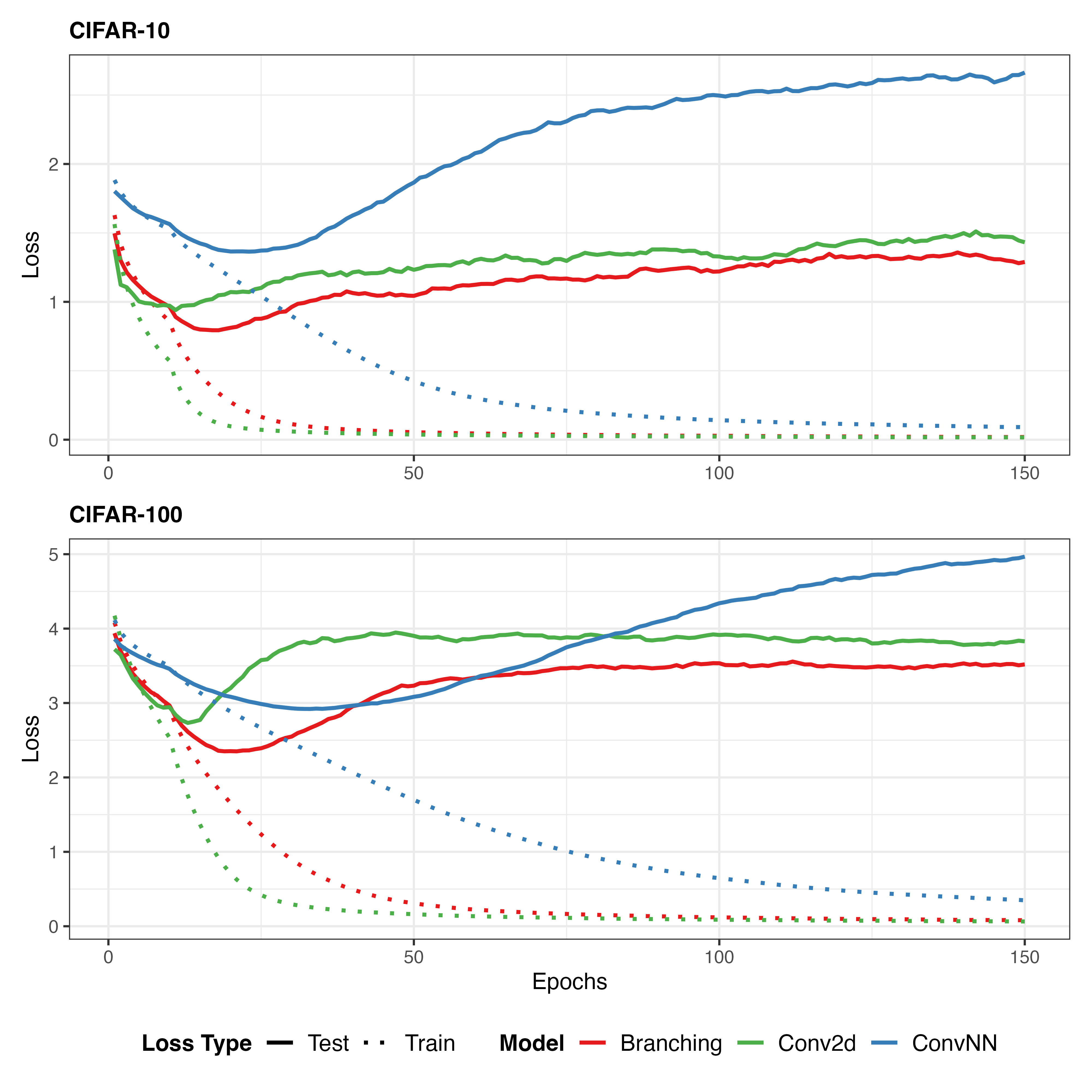}
        \caption{$lr = 10^{-4}$}
        \label{fig:vgg_loss_1e4}
    \end{subfigure}
    
    \begin{subfigure}[t]{.5\linewidth}
        \centering
        \includegraphics[width=\linewidth]{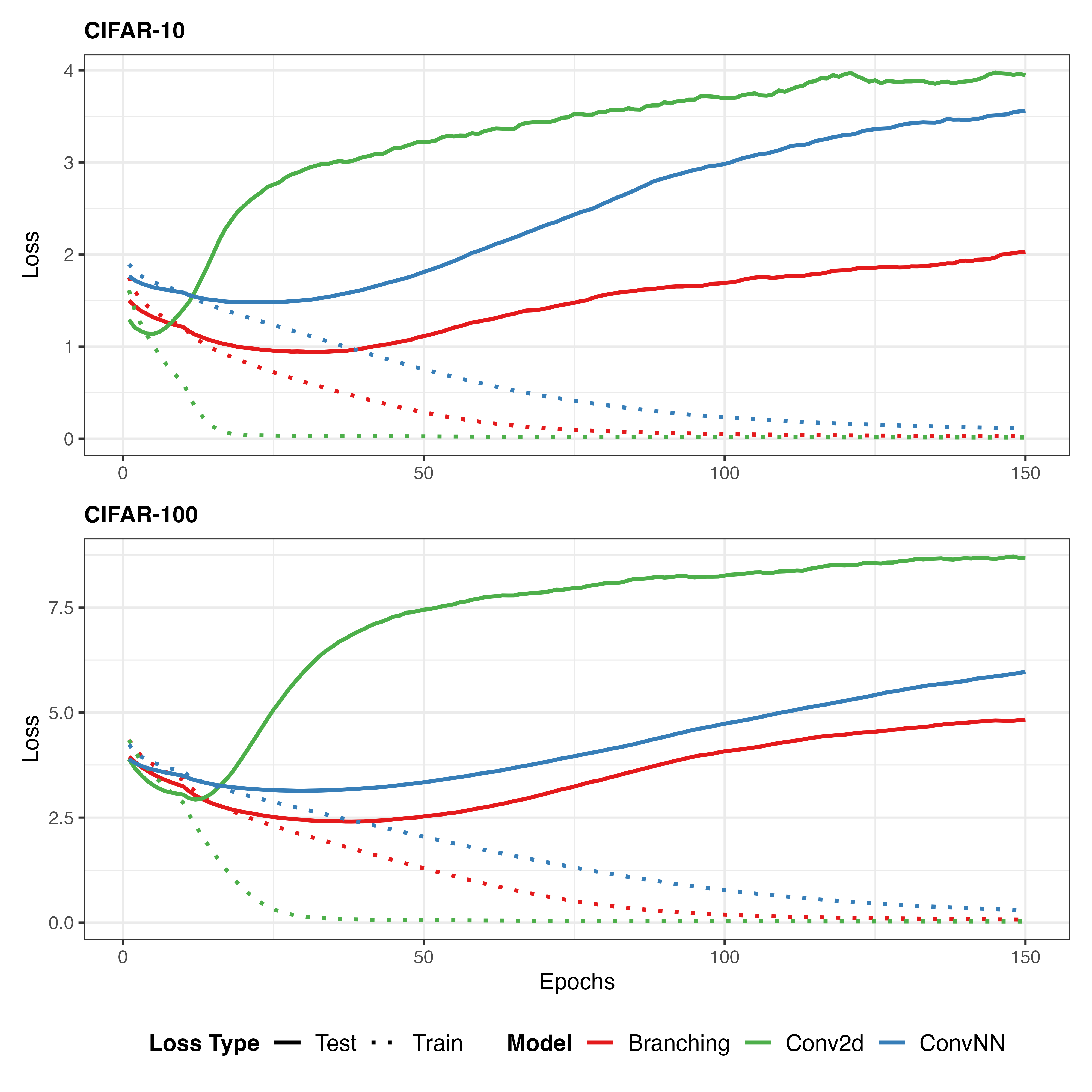}
        \caption{$lr = 10^{-5}$}
        \label{fig:vgg_loss_1e5}
    \end{subfigure}%
    \vspace{-10pt}
    \caption{Learning rate sensitivity analysis for Branching and \convnn on VGG-11. For visualization purposes, all loss curves are smoothed using a moving average with a window size of 10.}
    \label{fig:vgg_lr_ablation}
\end{figure}

We investigate the robustness of the above analysis further across different learning rates of $10^{-3}, 10^{-4}, 10^{-5}$. Figure~\ref{fig:vgg_lr_ablation} highlights that hybrid branching approach remains stable and effective across all learning rate settings on both CIFAR-10 and CIFAR-100. Notably, pure \convnn outperforms the \textsc{Conv2D} baseline at both extreme learning rates ($10^{-3}$ and $10^{-5}$), suggesting it benefits from either aggressive or conservative optimization rate. However, the hybrid branching consistently achieves the best performance across all learning rates, confirming that its regularization benefits are robust to optimization of training hyperparameters.

\subsubsection{Effect of Neighbor Count $k$}

\begin{figure}[H]
    \centering
    \vspace{10pt}
    \includegraphics[width=1.0\linewidth]{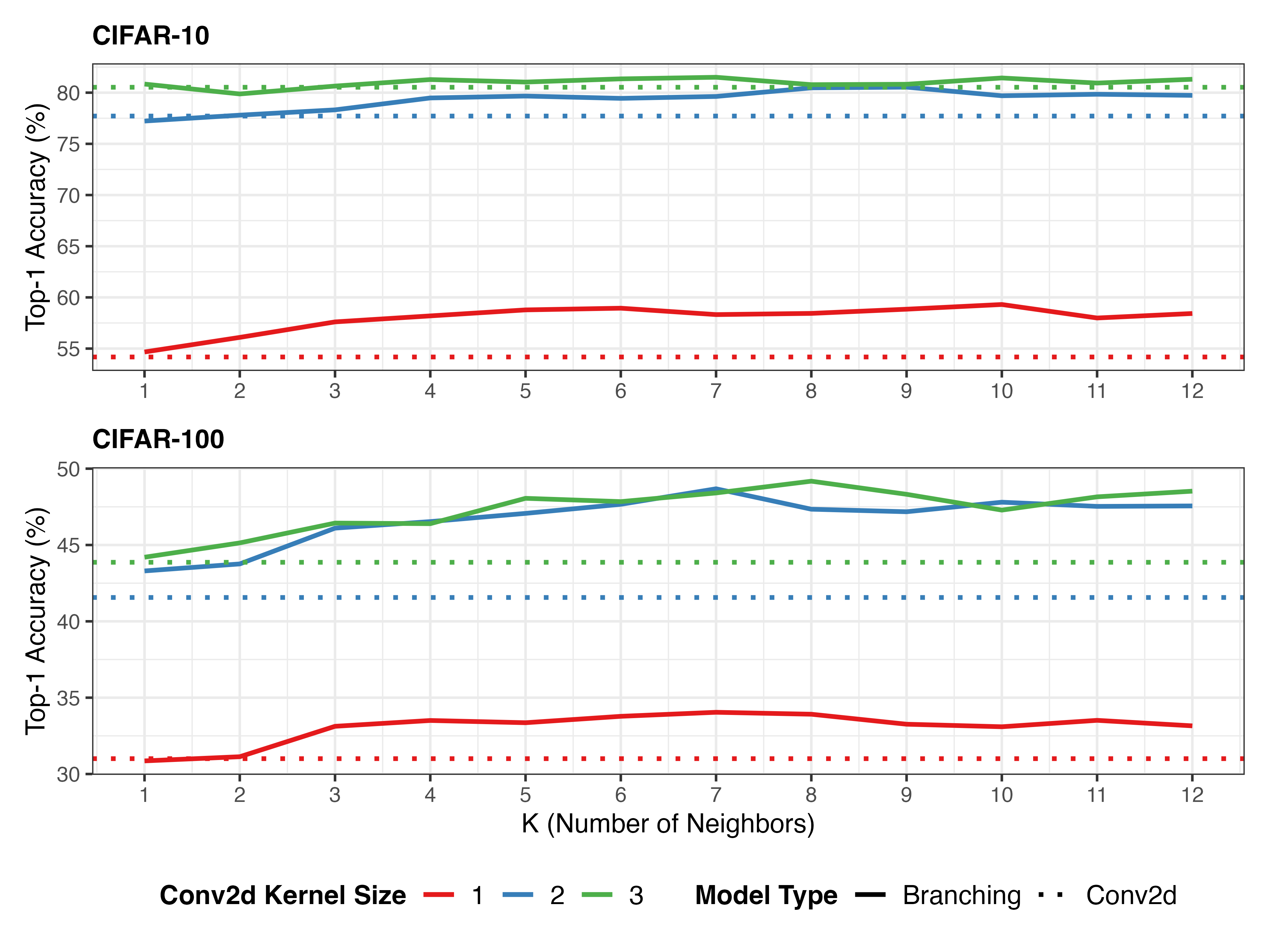}
    \vspace{-10pt}
    \caption{Impact of neighbor count $k$ on VGG-11 performance for the hybrid branching module against the \textsc{Conv2D} baseline.}
    \label{fig:vgg_k_ablation}
\end{figure}

Figure~\ref{fig:vgg_k_ablation} examines the sensitivity of the hybrid branching module to the number of neighbors $k$ on the \convnn\ branch and the kernel size on the \textsc{Conv2D} branch. Three observations emerge. First, the addition of \convnn\ branch consistently improves over the sole \textsc{Conv2D} baseline across all values of $k$. Second, increasing $k$ beyond a moderate value does not proportionally improve accuracy, indicating that a small neighborhood is sufficient to capture the most informative global context and that larger $k$ introduces diminishing returns at higher computational cost. Third, the benefit of \convnn is more pronounced on CIFAR-100 than on CIFAR-10, suggesting that global feature selection is more valuable when number of classes is larger and per-class training examples are fewer.

\subsubsection{Effect of Sparse Candidate Search}

\begin{figure}[H]
    \centering
    \begin{subfigure}[t]{\linewidth}
        \centering
        \includegraphics[width=1.0\linewidth]{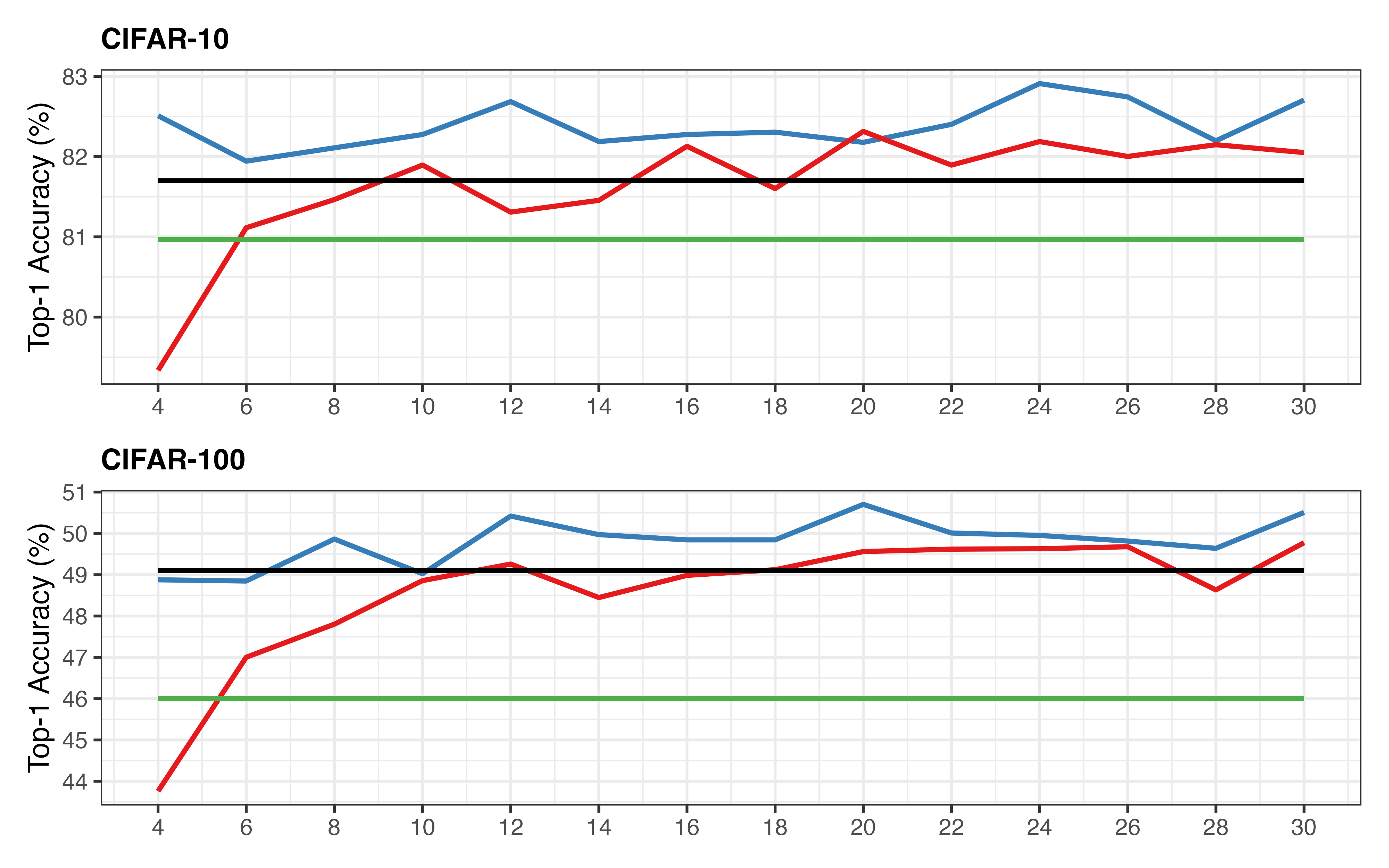}
    \end{subfigure}%
    \hfill
    \begin{subfigure}[t]{\linewidth}
        \centering
        \includegraphics[width=1.0\linewidth]{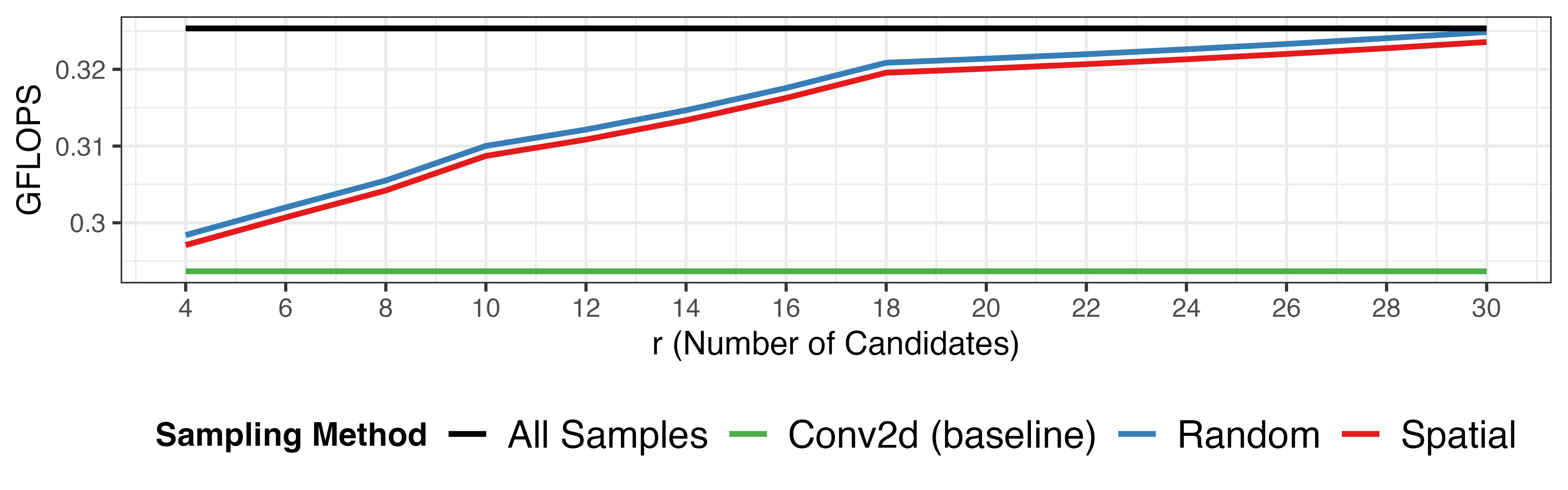}
    \end{subfigure}%
    \vspace{-10pt}
    \caption{Impact of candidate pool size $r$ on VGG-11 performance and GFLOPS. The Random and Spatial candidate selection curves on GFLOPS are slightly offset vertically for clarity.}
    \label{fig:vgg_random_spatial}
\end{figure}

Figure~\ref{fig:vgg_random_spatial} compares the random and spatial candidate selection strategies against the full-feature \convnn\ baseline and \textsc{Conv2D}. As expected, sparse variants significantly reduce computational cost, with GFLOPS approaching of that \textsc{Conv2D} for small values of $r$. Notably, increasing the number of candidate neighbors generally improves accuracy, and in some configurations sparse \convnn\ in the hybrid branching even surpasses the full-feature variant. This behavior is intuitive in the sense that restricting neighbor selection to a spatially diverse or randomly sampled subset encourages \convnn\ to capture more varied and distant relationships, mitigating locality bias and complementing the \textsc{Conv2D} branch. The random selection strategy consistently outperforms spatial selection across all values of $r$, and combined with its lower computational cost, presents an appealing operating point. 

\subsubsection{Effect of Patch Size}

\begin{table}[H]
\caption{Patch-Level Comparisons with $p \in [1, 2]$ for VGG-11 on CIFAR-10 and CIFAR-100 datasets with no sparse neighbor search.}
\small
\centering
    \begin{tabular}{clcccc}
    
    \multicolumn{6}{l}{\textbf{CIFAR-10}} \\
    \toprule
    Layer Type & $p$ & Test Loss & Acc (\%) & GFLOPS & \# Params \\
    \midrule[1pt]
    \textsc{Conv2D} & 1 & 1.529 & 80.65 & \textbf{0.293} & 130.015M\\
    \midrule[.5pt]
    Branching ($\lambda = 0.5$) & 1 & 1.424 & \textbf{81.26} & 0.325 & 130.015M\\
    Branching ($\lambda = 0.5$) & 2 & 1.403 & 74.81 & 0.307 & 199.152M\\
    \convnn & 1 & 2.694 & 54.00 & 0.325 & 130.015M \\
    \convnn & 2 & 1.631 & 60.32 & 0.307 & 268.289M \\
    
    \\[-5pt]
    
    \multicolumn{6}{l}{\textbf{CIFAR-100}} \\
    \toprule
    Layer Type & $p$ & Test Loss & Acc (\%) & GFLOPS & \# Params \\
    \midrule[1pt]
    \textsc{Conv2D} & 1 & 3.939 & 44.83 & \textbf{0.293} & \textbf{130.384M}\\
    \midrule[.5pt]
    Branching ($\lambda = 0.5$) & 1 & 3.474 & \textbf{48.37} & 0.325 & 130.384M\\
    Branching ($\lambda = 0.5$) & 2 & 4.242 & 37.01 & 0.308 & 199.512M\\
    \convnn & 1 & 5.011 & 29.50 & 0.325 & 130.384M \\
    \convnn & 2 & 4.487 & 28.17 & 0.308 & 268.658M \\   
    
    \end{tabular}
    \label{tab:patch_comparison}
    \vspace{10pt}
\end{table}

Table~\ref{tab:patch_comparison} examines the effect of increasing the patch size $p$ from 1 to 2, which reduces GFLOPS but significantly degrades accuracy across all configurations. The parameter count increases substantially with $p$ due to the quadratic scaling of channel dimension with patch size. For $p = 2$, \convnn\ parameters nearly double from 130M to 268M. This parameter inflation likely requires significantly more training to converge and explains the poor performance. The hybrid branching model exhibits greater robustness to patch size variation than pure \convnn, but both configurations confirm that fine-grained spatial resolution ($p=1$) is critical for effective feature-similarity neighbor selection on small images such as those in CIFAR-10 and CIFAR-100

\subsection{\convnn\ as a Drop-in Replacement for Self-Attention}
\label{subsec:cifar_vit}
For attention-based experiments, we use a ViT-Tiny backbone (\cite{DBLP:conf/iclr/DosovitskiyB0WZ21}). To ensure a fair comparison with standard attention, we employ the full set of learned linear projections, omit positional encoding, and set $\rho = \operatorname{softmax}$. Following the connection to KVT-attention established in Section~\ref{sec:convnn_attn_connection}, \convnn\ uses a depthwise \textsc{Conv1D} with learnable weights initialized to 1 and no bias. All methods, including standard self-attention and \convnn-attention, use a single attention head ($\text{NH} = 1$) throughout the CIFAR experiments, isolating the effect of the neighbor selection mechanism from the representational capacity introduced by multi-head splitting. The multi-head setting is evaluated separately in the ImageNet-1K experiments of Section~\ref{subsec:imagenet_vit}. Unless otherwise stated, $k = 9$ and $r = 32$. We compare against standard self-attention (\cite{vaswani2017attention}), KVT-attention at $k = 9$ and $k = 100$ (\cite{DBLP:conf/eccv/WangWWLCLJ22}), Local attention (\cite{aguilera2024local}), and Sparse attention (\cite{child2019generating}), all trained from scratch within the same ViT-Tiny architecture. 

\subsubsection{Quantitative Comparison}

\begin{table}[H]
\centering
\footnotesize
\vspace{4pt}
\caption{Test loss and accuracy for ViT-Tiny on CIFAR-10 and
         CIFAR-100. Bold values indicate the best result per column;
         underlined values indicate the second best. Methods:
         $^\dagger$\citet{vaswani2017attention},
         $^\ddagger$\citet{aguilera2024local},
         $^\S$\citet{DBLP:conf/eccv/WangWWLCLJ22},
         $^\P$\citet{child2019generating}.}
\setlength{\tabcolsep}{4pt}
\resizebox{\linewidth}{!}{%
    \begin{tabular}{lcccccccc}
    \toprule
    Layer Type
        & \multicolumn{4}{c}{CIFAR-10}
        & \multicolumn{4}{c}{CIFAR-100} \\
    \cmidrule(lr){2-5} \cmidrule(lr){6-9}
        & Test Loss & Acc (\%) & GFLOPS & \# Params.
        & Test Loss & Acc (\%) & GFLOPS & \# Params. \\
    \midrule
    Self-Attention$^\dagger$
        & \underline{1.558} & \underline{77.41} & 2.507 & \underline{5.479M}
        & 3.817 & 50.07 & 2.507 & \underline{5.496M} \\
    Local-Attention$^\ddagger$
        & 1.626 & 76.51 & \textbf{1.991} & \textbf{4.299M}
        & \underline{3.688} & 49.73 & \textbf{1.991} & \textbf{4.317M} \\
    KVT-Attention ($k=9$)$^\S$
        & 1.775 & 75.77 & 2.508 & \underline{5.479M}
        & 3.883 & 48.21 & 2.508 & \underline{5.496M} \\
    KVT-Attention ($k=100$)$^\S$
        & 1.593 & 76.41 & 2.508 & \underline{5.479M}
        & 3.738 & \underline{50.34} & 2.508 & \underline{5.496M} \\
    Sparse-Attention$^\P$
        & 2.026 & 72.50 & 2.509 & \underline{5.479M}
        & 4.565 & 44.62 & 2.509 & \underline{5.496M} \\
    \midrule
    \convnn\ (All Features)
        & 1.978 & 73.67 & 2.333 & 5.500M
        & 3.983 & 48.67 & 2.333 & 5.517M \\
    \convnn\ (Random Selection)
        & \textbf{0.779} & \textbf{79.82} & \underline{2.037} & 5.500M
        & \textbf{2.341} & \textbf{53.60} & \underline{2.037} & 5.517M \\
    \convnn\ (Spatial Selection)
        & 2.737 & 63.01 & \underline{2.037} & 5.500M
        & 5.020 & 38.22 & \underline{2.037} & 5.517M \\
    \bottomrule
    \end{tabular}}
\label{tab:vit_exp}
\vspace{10pt}
\end{table}

Table~\ref{tab:vit_exp} summarizes the results across all attention variants. \convnn\ with random candidate selection achieves the best test loss and accuracy on both CIFAR-10 and CIFAR-100, outperforming standard self-attention, KVT-attention, Local attention, and Sparse attention. This improvement comes with only a modest increase in parameter count (5.500M vs. 5.479M) and competitive GFLOPS. In contrast, \convnn\ using all features as candidates or spatial selection underperforms the attention baseline, suggesting that the choice of neighbor selection strategy is critical to the framework's effectiveness on small datasets such as CIFAR-10 and CIFAR-100. The spatial selection strategy, which selects candidates at regular spatial intervals, effectively reintroduces a locality bias that limits the global context that \convnn\ can exploit. Random selection, by contrast, encourages a diverse and spatially distributed neighborhood that complements the global nature of the similarity-based selection.

\subsubsection{Training Dynamics}

\begin{figure}[H]
    \vspace{5pt}
    \centering
    \begin{subfigure}[t]{\linewidth}
        \centering
        \includegraphics[width=0.8\linewidth]{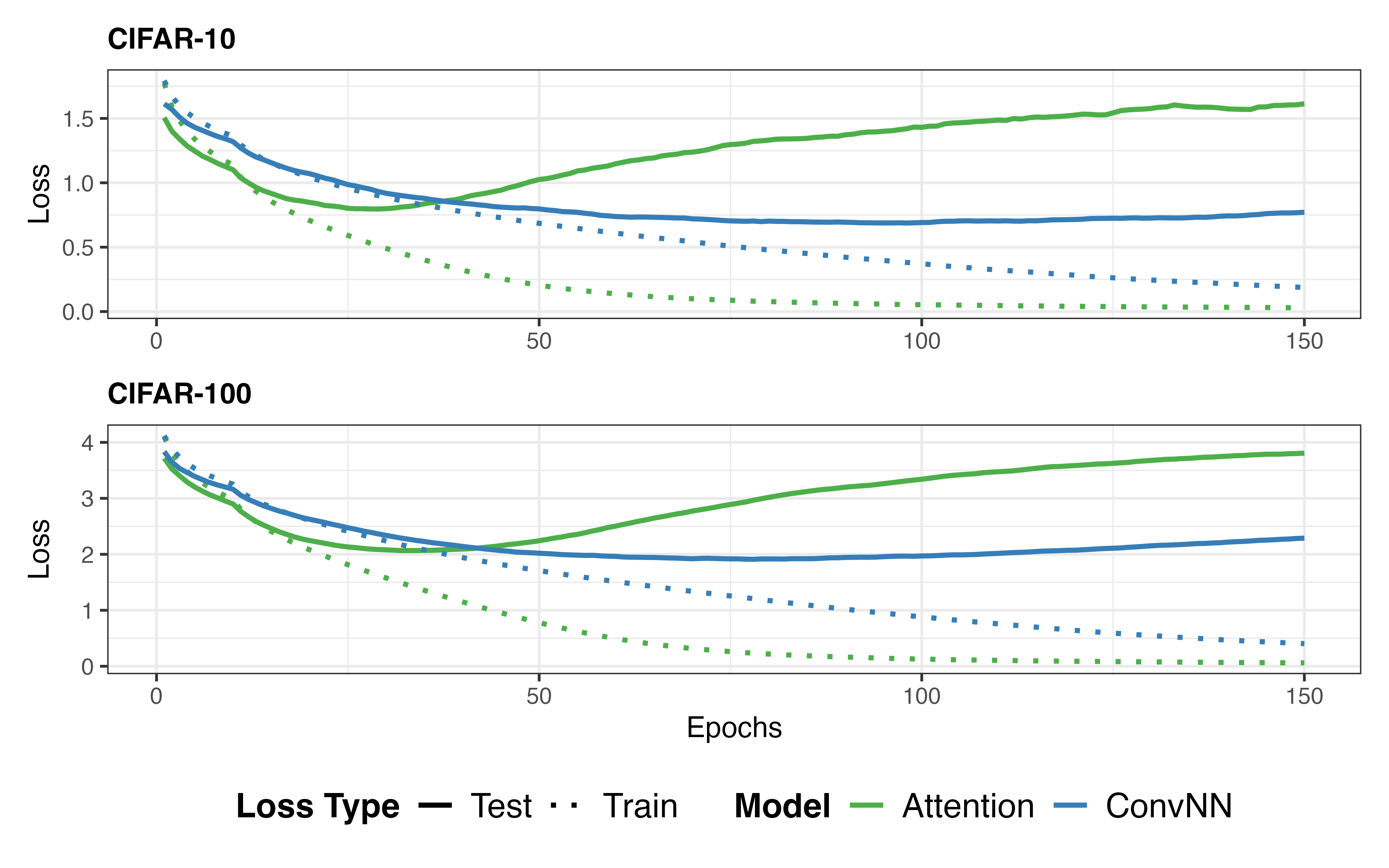}
    \end{subfigure}%
    \vspace{-10pt}
    \caption{Training and test loss for ViT-Tiny on CIFAR-10 and CIFAR-100. \convnn\ uses random candidate selection with $r=32$. All curves are smoothed with a moving average of window size 10. Learning rate is set to $10^{-4}$.}
    \label{fig:vit_loss}
\end{figure}

Figure~\ref{fig:vit_loss} shows the training and test loss trajectories for variants on ViT-Tiny. \convnn\ with random selection exhibits notably different convergence behavior from self-attention. It avoids the early overfitting seen in self-attention and achieves lower test loss throughout training. This suggests that randomized neighbor selection provides an implicit regularization effect, preventing the model from over-relying on any fixed set of image patches and encourages more generalizable representations.

\subsubsection{Effect of Neighbor Count $k$}
\begin{figure}[H]
    \centering
    \begin{subfigure}[t]{\linewidth}
        \centering
        \includegraphics[width=1.0\linewidth]{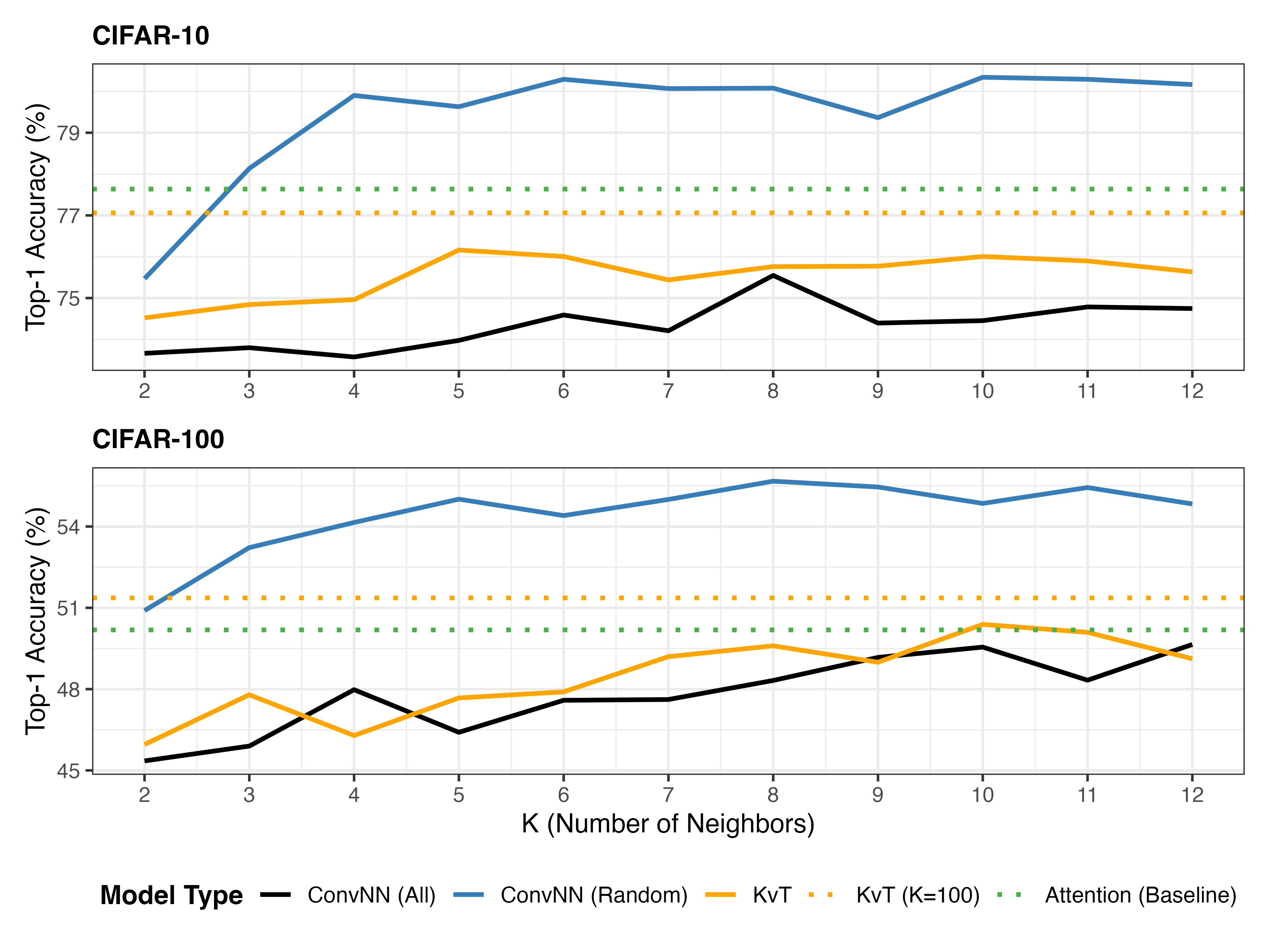}
    \end{subfigure}%
    \vspace{-10pt}
    \caption{Impact of neighbor count $k$ on ViT-Tiny performance. KVT-attention is evaluated across varying $k$ and performs best at $k=100$.}
    \label{fig:vit_vary_k}
\end{figure}

Figure~\ref{fig:vit_vary_k} shows the impact of $k$ on \convnn\ compared to standard attention and KVT-attention. When all features are used as candidates, \convnn\ achieves performance comparable to KVT-attention. Under random neighbor selection, however, \convnn\ substantially outperforms both KVT-attention and self-attention across all values of $k$. As observed in the convolutional experiments, accuracy gains diminish beyond a moderate $k$, confirming that a small but well-chosen neighborhood is sufficient to capture the most relevant global context.

\subsubsection{Effect of Sparse Neighbor Search}

\begin{figure}[H]
    \centering
    \begin{subfigure}[t]{\linewidth}
        \centering
        \includegraphics[width=1.0\linewidth]{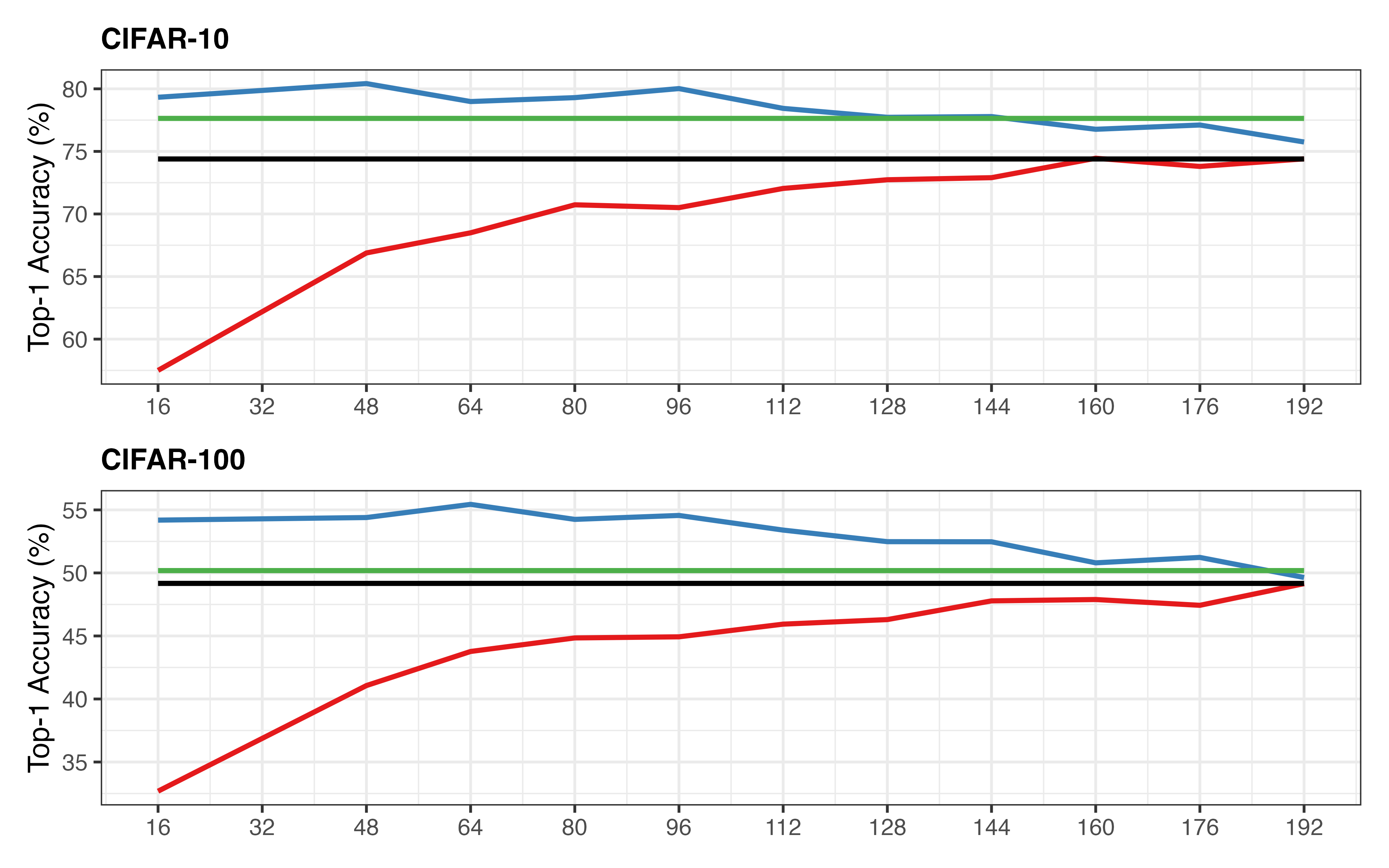}
    \end{subfigure}%
    \hfill
    \begin{subfigure}[t]{\linewidth}
        \centering
        \includegraphics[width=1.0\linewidth]{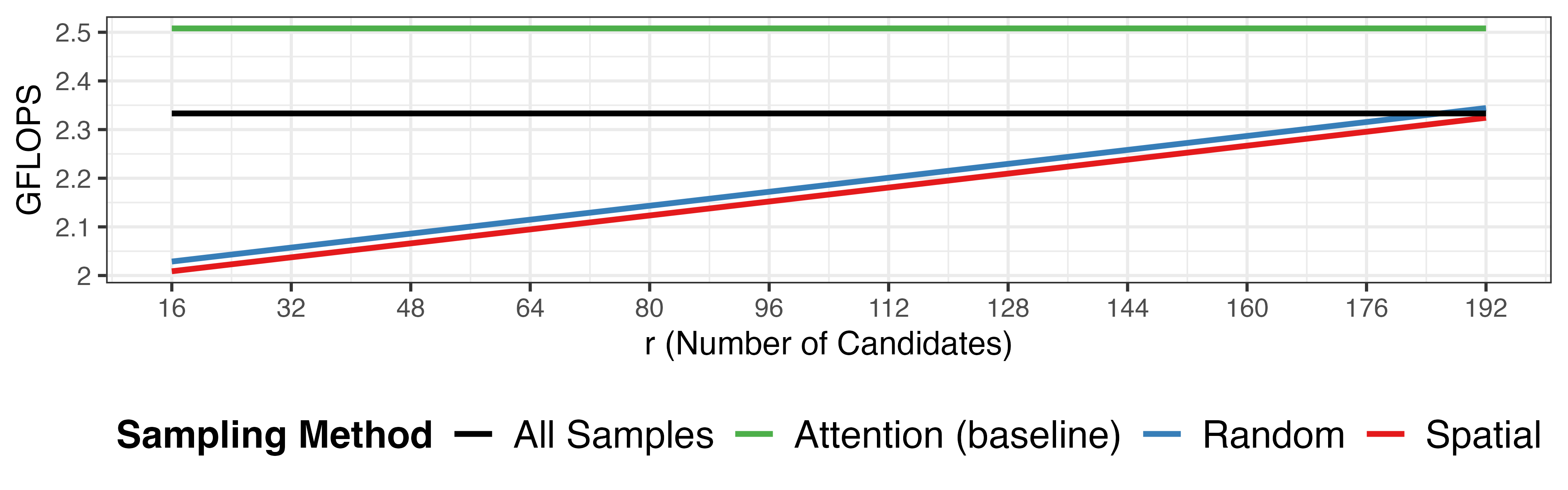}
    \end{subfigure}%
    \vspace{-10pt}
    \caption{Impact of $r$ to \convnn's performance on ViT-Tiny training. The Random and Spatial candidate selection curves are slightly offset vertically for clarity, had the same runtime throughout the experiments.}\label{fig:vit_random_spatial}
\end{figure}

Figure~\ref{fig:vit_random_spatial} compares random and spatial candidate selection against standard attention and full-feature \convnn\ across varying candidate pool sizes $r$. \convnn\ demonstrates lower GFLOPS than self-attention across all configurations. At $r=16$, both sparse strategies achieve GFLOPS approximately 20\% less than the baseline. In terms of accuracy, spatial selection consistently underperforms the baseline, while random selection outperforms it in all configurations, including those with the smallest candidate pools. This contrast reinforces the central finding: spatial selection reintroduces locality bias that limits the benefit of feature-similarity based neighbor selection, whereas random selection actively encourages \convnn\ to discover long-range relationships and thereby exploit its full representational capacity.

\subsubsection{Convolution Aggregation Variation in \convnn-Attention for ViT}
\begin{table}[H]
\small
\caption{Aggregation kernel comparison for \convnn-attention in ViT-Tiny on CIFAR-10 and CIFAR-100. D, DS, and S denote depthwise, depthwise separable, and standard convolution, respectively. Depthwise weights are initialized to 1. All \convnn\ variants use full candidate sampling and $k = 9$.}
\centering
    \begin{tabular}{clcccc}
    
    \multicolumn{5}{l}{\textbf{CIFAR-10}} \\
    \toprule
    Layer Type &  Test Loss & Acc (\%) & GFLOPS & \# Params \\
    \midrule[1pt]
    Attention & 1.558 & 77.41 & 2.507 & 5.479M\\
    \convnn (D)  & 1.978 & 73.67 & 2.333 & 5.500M\\
    \convnn (DS) & 1.939 & 71.93 & 2.333 & 5.942M \\
    \convnn (S)  & 1.940 & 72.90 & 2.333 & 9.460M \\
    
    \\[-5pt]
    
    \multicolumn{5}{l}{\textbf{CIFAR-100}} \\
    \toprule
    Layer Type &  Test Loss & Acc (\%) & GFLOPS & \# Params \\
    \midrule[1pt]
    Attention & 3.817 & 50.07 & 2.507 & 5.496M\\
    \convnn (D)  & 3.983 & 48.67 & 2.333 & 5.517M\\
    \convnn (DS) & 4.140 & 43.75 & 2.333 & 5.959M\\
    \convnn (S)  & 3.900 & 47.17 & 2.333 & 9.477M \\
    
    \end{tabular}
    \label{tab:vit_conv_variants}    
\vspace{5pt}
\end{table}

We investigate the effect of the convolutional aggregation kernel in the weighted aggregation step of \convnn-attention by comparing three variants: depthwise (D), depthwise separable (DS) (\cite{chollet2017xceptiondeeplearningdepthwise}), and standard (S) convolution. Depthwise convolution maintains channel separation with weights initialized to 1, consistent with the KVT equivalence established in Section~\ref{sec:convnn_attn_connection}. Depthwise separable convolution factorizes the aggregation into a depthwise spatial step followed by a pointwise channel-mixing step, reducing parameters relative to standard convolution. Standard convolution applies a fully learned dense kernel across all input channels. All variants use full candidate sampling without sparse selection and set $k = 9$. 

Table~\ref{tab:vit_conv_variants} reveal three key findings. First, depthwise convolution achieves the highest accuracy on both datasets while introducing the fewest additional parameters relative to the attention baseline (5.500M vs. 5.479M on CIFAR-10). Second, standard convolution nearly doubles the parameter count to approximately 9.5M without a significant accuracy gain, indicating that dense cross-channel aggregation is an inefficient use of parameters in this setting. Third, depthwise separable convolution underperforms depthwise despite its intermediate parameter count, suggesting that the pointwise channel-mixing step does not provide any benefit for neighbor aggregation and may interfere with the channel-specific similarity structure established during neighbor selection. 

Notably, the test loss values across all three variants are closely clustered (1.939 to 1.978 on CIFAR-10), indicating that the choice of aggregation kernel has a smaller effect on final performance than the neighbor selection mechanism itself. This reinforces \convnn's representational advantage over standard attention which stems primarily from its flexible neighbor selection rather than from the specific aggregation strategy employed.

\section{Conclusion of Proof of Concept Validation}
The CIFAR-10 and CIFAR-100 experiments establish several consistent findings that motivate the large-scale ImageNet-1K evaluation in Section~\ref{Chapter5}. 

In the convolutional setting, the hybrid branching layer demonstrates that combining local spatial aggregation from \textsc{Conv2D} with global feature-similarity selection from \convnn\ consistently outperforms either branch in isolation across virtually all values of $\lambda$, without introducing additional learnable parameters besides the pointwise convolution layer to merge branch outputs. The improvement is most pronounced on CIFAR-100, where the higher number of fine-grained classes places greater demand on global discriminative context. This benefit is robust to learning rate choice and is further amplified, albeit modestly, by the addition of positional encoding to the similarity computation. Sparse candidate selection via random sampling reduces GFLOPS to near \textsc{Conv2D} levels whil maintaining competitive accuracy, and in some configurations even surpasses full-feature \convnn, suggesting that controlled diversity in neighbor selection provides an additional regularization benefit. 

In the attention experiments, \convnn\ with random candidate selection achieves the best test loss and accuracy on both datasets, outperforming standard self-attention, KVT-attention, Local attention, and Sparse attention at comparable or lower GFLOPS. The training dynamics reveal that random neighbor selection confers an implicit regularization effect, preventing the early overfitting observed in standard self-attention. Spatial candidate selection, by contrast, consistently underperforms by reintroducing a locality bias that suppresses the long-range relationships that \convnn\ is designed to exploit. The ablation over aggregation kernel variants confirms that depthwise convolution is the most parameter-efficient aggregation choice, and that the neighbor selection mechanism is the primary drive of \convnn's performance advantage, with the aggregation strategy playing a secondary role. 

Together, these results validate the core claims of the \convnn\ framework that convolution and self-attention can be unified under a single neighbor selection and aggregation formulation, that the intermediate spectrum between local and global aggregation is consistently more effective than either extreme. This further demonstrates the framework's capabilities to scale gracefully to different architectural contexts, similarity metrics, and candidate selection strategies. These conclusions motivate the transition to full-scale ImageNet-1K evaluation, where multi-head attention, larger model capacities, and standard data augmentation are introduced. 

%% file: Chapters/chapter5.tex

\chapter{Large Scale Experiments, Results, and Discussions}
\label{Chapter5} 
\section{Large Scale ImageNet-1K Experimentation}
\label{sec:large_scale}

\subsection{ImageNet and the ImageNet Large Scale Visual Recognition Challenge}
The ImageNet project represents a monumental effort in dataset curation for computer vision. Originally introduced by \citet{deng2009imagenet}, ImageNet is a large-scale visual database organized according to the WordNet hierarchy, containing over 14 million hand-annotated images across more than 20,000 synsets (synonym sets). This comprehensive resource has fundamentally shaped modern deep learning research and practice. 

The success of ImageNet as a benchmark catalyzed the ImageNet Large Scale Visual Recognition Challenge (ILSVRC) (\cite{ILSVRC15}), an annual competition running from 2012 to 2017. The ILSVRC introduced standardized evaluation protocols for object detection, localization, classification and scene understanding tasks. The competition's influence cannot be overstated as it directly motivated breakthrough architectures such as AlexNet (\cite{DBLP:journals/cacm/KrizhevskySH17}), which initiated the deep learning revolution in computer vision. 

\subsection{ImageNet-1K: The Classification Benchmark}
ImageNet-1K, formally known as the ILSVRC 2012 image classification dataset (\cite{ILSVRC15}), has become the standard benchmark for evaluating visual recognition models. The dataset comprises 1,281,167 training images, 50,000 validation images (50 per class), and 100,000 test images (100 per class), spanning 1,000 semantically diverse and non-exclusive object categories drawn from WordNet. The large-scale nature of ImageNet-1K presents significant challenges that drive algorithmic innovation that is able to understand diverse object appearance, complex backgrounds, viewpoint variations, and within-class visual diversity. These challenging aspects collectively make ImageNet-1K a rigorous evaluation dataset for assessing model generalization for visual tasks. 

\subsection{Significance and Impact on Vision Transformer Development}
Since AlexNet's 2012 win on ILSVRC (\cite{DBLP:journals/cacm/KrizhevskySH17}), ImageNet-1K has remained the primary benchmark for evaluating advances in computer vision. Recent architectures, particularly Vision Transformers (\cite{DBLP:conf/iclr/DosovitskiyB0WZ21}), Data-Efficient Image Transformers (DeiT) (\cite{touvron2021training}), and Swin Transformers (\cite{DBLP:conf/iccv/LiuL00W0LG21}) are routinely evaluated on ImageNet-1K to establish SOTA performance. 

The standardization provided by ImageNet-1K enables fair comparison across methods and institutional boundaries, a critical requirement for reproducible research in computer vision. Our evaluation on ImageNet-1K allows direct comparison with prior work on attention mechanisms and convolutional approaches to visual recognition. 

\subsection{Dataset and Experimental Setup}
The ImageNet-1K dataset, derived from the ILSVRC 2012 challenge, holds 1.28M training images, 50K validation images, and 100K test images spanning 1,000 object categories. We follow closely the standard protocol established by the Swin Transformer (\cite{DBLP:conf/iccv/LiuL00W0LG21}), which has become the standard for vision transformer evaluation on large-scale datsets. 

\subsection{Data Augmentation Strategy}
\subsubsection{Training Augmentation}
Training images undergo the following sequential transformations: random resized cropping to $224 \times 224$ with bicubic interpolation, random horizontal flipping (probability 0.5), color jittering ($\lambda = 0.4$), RandAugment (\cite{cubuk2020randaugment}) with magnitude $m = 9$, number of layers $n = 2$, and standard deviation $\sigma = 0.5$ (\texttt{rand-m9-mstd0.5-inc1}), and random erasing (probability 0.25, pixel-level model). 

\subsubsection{MixUp and CutMix}
Following established practice (\cite{zhang2017mixup, yun2019cutmix}), we apply MixUp and CutMix augmentation to each training batch with probability $p = 1.0$, selecting stochastically between the two strategies with equal probability 0.5. MixUp uses mixing coefficient $\alpha = 0.8$ and CutMix uses $\alpha = 1.0$. Label smoothing of $\epsilon = 0.1$ is applied throughout training. 

\subsubsection{Validation Augmentation}
Validation and test images undergo lighter preprocessing of resizing shorter edges to 256 pixels (preserving aspect ratio) with bicubic interpolation, center crop to $224 \times 224$, and per-channel normalization using $\mu = (0.485, 0.456, 0.406)$, $\sigma = (0.229, 0.224, 0.225)$, where $\mu$ and $\sigma$ are the per-channel mean and standard deviation computed over the RGB channels of the ImageNet-1K training set.

\subsection{Training Configuration}
\subsubsection{Loss Functions and Optimization}
We employ SoftTargetCrossEntropy loss (\cite{hugger2024towards}) during training to accommodate soft labels from MixUp/CutMix (\cite{yun2019cutmix, zhang2017mixup}) operations, while using standard CrossEntropy loss for validation. All models use AdamW optimizer (\cite{loshchilov2017decoupled}) with learning rate $\text{lr} = 10^{-3}$, weight decay $\lambda = 0.05$, and momentum parameters $\beta_1 = 0.9, \beta_2 = 0.999$. Gradient norm clipping is set to 5.0.

\subsubsection{Learning Rate Schedule}
We adopt a cosine annealing schedule with linear warmup (\cite{loshchilov2016sgdr}) with initial warmup learning rate of $10^{-7}$, 20 epoch warmup phase, followed by cosine decay to $\text{lr}_{\min} = 5 \times 10^{-6}$ over 300 training epochs. 

\subsubsection{Mixed Precision and Numerical Optimization}
We leverage automatic mixed precision (AMP) training, utilizing bfloat16 precision on H100 GPUs where available, otherwise falling back to float16. Additionally, we enable TensorFloat-32 (TF32) matrix multiplications for computational efficiency on Ampere and newer GPU architectures. 

\subsection{Distributed Training}
\paragraph{Hardware Configuration}
Experiments are implemented in PyTorch 2.0 (\cite{pytorch2025}) and were conducted on the Indiana Jetstream2 platform (\cite{10.1145/3437359.3465565}) with four NVIDIA H100 GPUs. We employ PyTorch's Distributed Data Parallel (DDP) backend with NCCL communication, distributing a global batch size of 1,280 samples (320 per GPU). DistributedSampler ensures non-overlapping data partitions across all processes, preserving training reproducibility and preventing data leakage. 

\paragraph{Reproducibility}
All experiments are conducted with fixed random seeds applied to PyTorch, NumPy, and the Python random module to ensure deterministic behavior across runs.

\section{ImageNet-1K Results}
We evaluate \convnn\ on ImageNet-1K classification (\cite{ILSVRC15}) to assess its scalability beyond the proof-of-concept CIFAR experiments. Two architectural settings are used for the experimentation: ResNet-50 with hybrid branching as the convolutional replacement, and ViT-Base with \convnn as the attention replacement. All experiments follow the training protocol and data augmentation policy described above in Section~\ref{sec:large_scale}.

\subsection{\convnn\ Branching in ResNet-50}
\label{subsec:imagenet_resnet}

\subsubsection{Model Setup}
For convolutional experiments, we use a ResNet-50 backbone (\cite{he2016deep}) with hybrid branching layer ($\lambda = 0.5$) replacing all standard $3 \times 3$ convolutions. The \convnn\ branch uses identity projections and cosine similarity with uniform weights ($\rho(\mathbf{z}) = \mathbf{1}_K$), consistent with the convolutional setting described in Section~\ref{subsec:hybrid_branching_layer} and ~\ref{subsec:cifar_vgg}. Unlike the CIFAR experiments, the standalone \convnn\ and \textsc{Conv2D} baselines reported here are direct replacements for the convolutional layers without any branching module, and therefore do not include the pointwise $1 \times 1$ convolution. We vary the neighbor count $k \in \{4, 6, 9, 16\}$ and evaluate three candidate selection strategies: all-feature selection, random sparse selection with $r \in \{32, 48, 64\}$, and spatial selection with $r = 25$ samples from a $5 \times 5$ spatial grid ($r' = 5$). Sparse candidate search is disabled for the all-feature configurations. 

\subsubsection{Effect of Neighbor Count $k$ and Sparse Candidate Search}

\begin{figure}[H]
\centering
\begin{tikzpicture}
\begin{axis}[
    width=1.0\linewidth, height=9cm,
    xlabel={Neighbor count $k$},
    ylabel={Top-1 Accuracy (\%)},
    xtick={4,6,9,16},
    xticklabels={4,6,9,16},
    ymin=78.0, ymax=80.2,
    ymajorgrids=true, grid style={dotted, gray!40},
    legend style={at={(0.02,0.98)}, anchor=north west, font=\small,
                  draw=gray!40, fill=white, row sep=1pt},
    tick label style={font=\small},
    label style={font=\footnotesize},
]
\addplot[color=blue!70, mark=*, thick, mark size=2pt]
    coordinates {(4,79.06) (6,79.19) (9,79.42) (16,79.76)};
\addlegendentry{Branch, All}

\addplot[color=teal!80, mark=triangle*, thick, dashed, mark size=2.5pt]
    coordinates {(4,78.73) (6,78.84) (9,78.98) (16,79.12)};
\addlegendentry{Branch, Rand ($r=32$)}

\addplot[color=gray!90, thick, dotted, mark=none]
    coordinates {(4,78.86) (6,78.86) (9,78.86) (16,78.86)};
\addlegendentry{\textsc{Conv2D} baseline $3 \times 3$}
\end{axis}
\end{tikzpicture}
\caption{Top-1 accuracy on ImageNet-1K for ResNet-50 with hybrid
         branching ($\lambda = 0.5$) across neighbor counts $k$,
         comparing all-feature and random sparse ($r=32$) candidate
         selection against the \textsc{Conv2D} baseline.}
\label{fig:imagenet_resnet_k}
\vspace{5pt}
\end{figure}
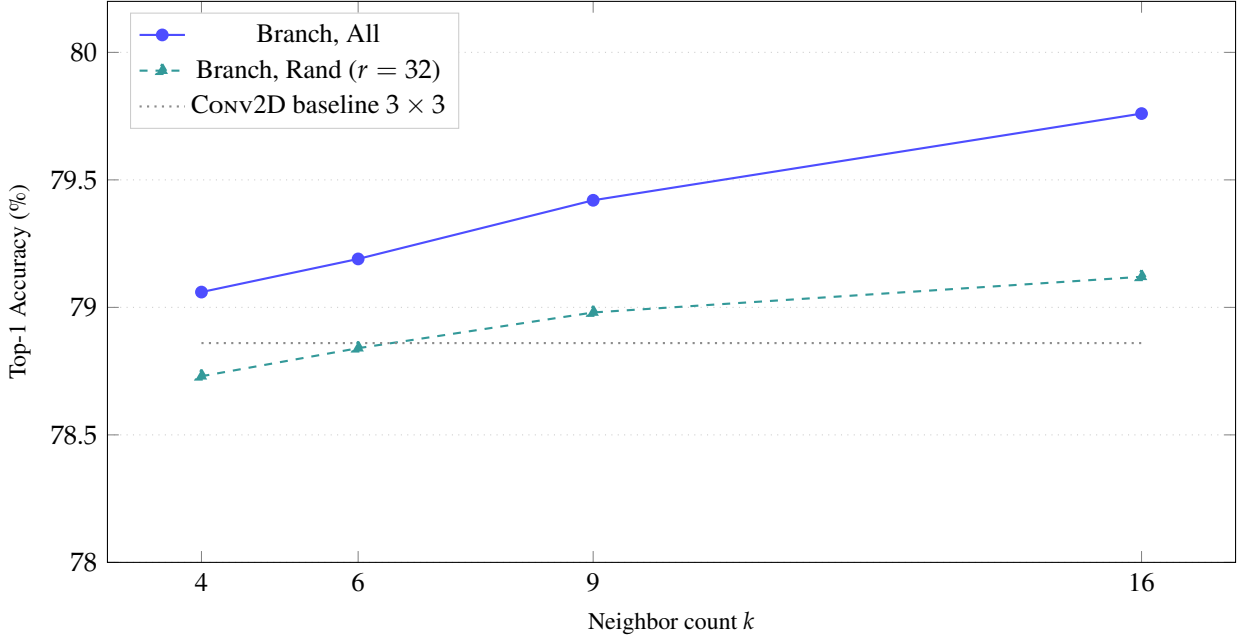

Table~\ref{tab:imagenet_resnet_results} and Figure~\ref{fig:imagenet_resnet_k} summarize the ResNet-50 results. The standard \textsc{Conv2D} baseline achieves 78.86\% top-1 accuracy. All hybrid branching configurations with all-feature selection surpass this baseline, with accuracy increasing monotonically with $k$: from 79.06\% at $k=4$ to 79.76\% at $k = 16$. This 0.90\% improvement over the baseline is achieved with only a modest increase in parameters, as the \convnn\ branch operate in tandem with the convolutional branch rather than augmenting it, with the only overhead being the pointwise $1 \times 1$ convolution that mixes the two branch outputs. The improvement reinforces the finding from the CIFAR experiments that combining local spatial aggregation with global feature-similarity selection provides a consistently better inductive bias than either alone. The pure \convnn\ replacement achieves 76.22\%, underperforming both the hybrid branching configurations and the \textsc{Conv2D} baseline, mirroring the pattern observed in the CIFAR experiments and again confirming that neither pure local nor pure global aggregation alone is optimal. 

Sparse random selection with $r = 32$ slightly underperforms all-feature selection across all values of $k$, but remains competitive with the \textsc{Conv2D} baseline while reducing GFLOPS from 10.655 to 5.596, a reduction of approximately 47\%. This makes random sparse \convnn\ a strong operating point when computational cost is constrained. The spatial selection configuration ($k=9$, $r=25$) achieves 78.82\%, on par with the baseline, suggesting that spatially regular subsampling reintroduces a locality bias that limits the benefit of feature-similarity neighbor selection, consistent with the CIFAR findings. 

\begin{figure}[H]
\centering
\begin{tikzpicture}
\begin{axis}[
    width=1.0\linewidth, height=10cm,
    xlabel={Sparse Candidate Count $r$},
    ylabel={Top-1 Accuracy (\%)},
    xtick={32, 48, 64},
    xticklabels={32\\{\scriptsize (5.596 GFLOPS)}, 48\\{\scriptsize (5.641 GFLOPS)}, 64\\{\scriptsize (5.686 GFLOPS)}},
    xticklabel style={font=\small, align=center},
    ymin=78.6, ymax=79.8,
    ymajorgrids=true, grid style={dotted, gray!40},
    legend style={at={(0.02,0.98)}, anchor=north west, font=\small,
                  draw=gray!40, fill=white, row sep=1pt},
    tick label style={font=\small},
    label style={font=\small},
]
\addplot[color=red!70, mark=*, thick, mark size=2pt]
    coordinates {(32,78.73) (48,79.03) (64,79.06)};
\addlegendentry{Branch, $k=4$ (24.191M)}

\addplot[color=orange!70, mark=*, thick, mark size=2pt]
    coordinates {(32,78.84) (48,78.99) (64,79.16)};
\addlegendentry{Branch, $k=6$ (25.104M)}

\addplot[color=green!90, mark=*, thick, mark size=2pt]
    coordinates {(32,78.98) (48,78.99) (64,79.24)};
\addlegendentry{Branch, $k=9$ (26.474M)}

\addplot[color=blue!70, mark=*, thick, mark size=2pt]
    coordinates {(32,79.12) (48,79.28) (64,79.37)};
\addlegendentry{Branch, $k=16$ (29.671M)}

\addplot[color=gray!90, thick, dotted, mark=none]
    coordinates {(32,78.86) (48,78.86) (64,78.86)};
\addlegendentry{\textsc{Conv2D} baseline $3 \times 3$ (25.583M)}
\end{axis}
\end{tikzpicture}
\caption{Top-1 accuracy on ImageNet-1K for ResNet-50 with hybrid
         branching ($\lambda = 0.5$) across sparse candidate pool
         $r$, for each neighbor count $k$, against the \textsc{Conv2D} baseline (25.583M parameters, 8.178 GFLOPS)}
\label{fig:imagenet_resnet_n}
\vspace{5pt}
\end{figure}
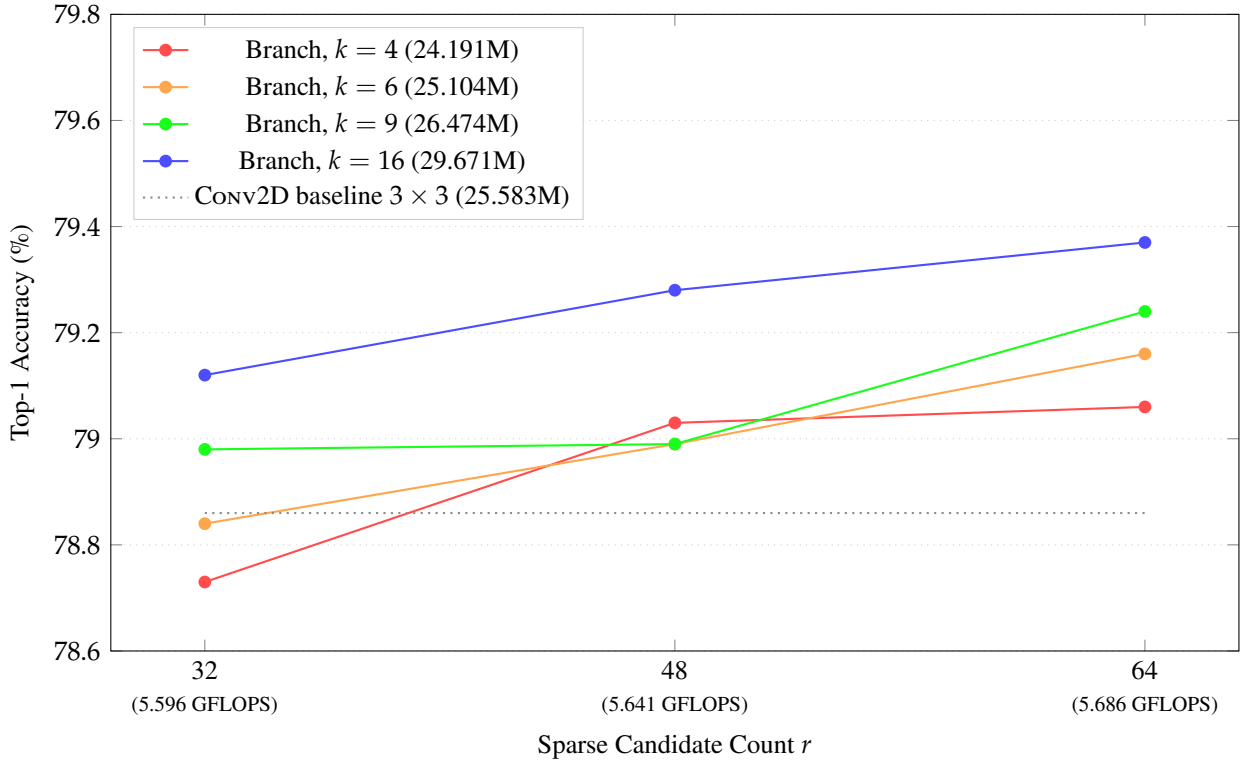

Figure~\ref{fig:imagenet_resnet_n} shows top-1 accuracy as a function of the sparse candidate pool size $r$ across all four values of $k$. The overall trend is clear: accuracy increases monotonically with $r$ for all values of $k$, which is intuitive since a large candidate pool provides a more representative sample of the global feature space from which to select the $k$ most informative neighbors. A notable anomaly occurs at $r = 48$, where $k = 4$ (24.191M) achieves higher accuracy than both $k = 6$ (25.104M) and $k = 9$ (26.474M). This result is likely attributed to the stochasticity inherent in random candidate selection rather than a systematic advantage of smaller $k$, as the global trend across $r$ and $k$ is one of jointly increasing performance. 

A particularly compelling operating point is the branching configuration with $k = 4$ and $r = 48$, which achieves 79.03\% top-1 accuracy at 5.641 GFLOPS and 24.191M parameters. Compared to the \textsc{Conv2D} baseline at 78.86\%, 8.178 GFLOPS, and 25.583M parameters, this configuration surpasses the baseline in accuracy while using approximately 1.4M fewer parameters and 31\% fewer GFLOPS. This demonstrates that random sparse \convnn\ branching layer's ability to achieve better accuracy-efficiency tradeoff than the standard convolutional baseline, making it an attractive operating point for compute-constrained deployments. 

\subsubsection{Full Results}

\begin{table}[H]
\vspace{5pt}
\centering
\caption{ResNet-50 on ImageNet-1K top-1 classification.
         Rand = random candidate selection, Spat = spatial candidate
         selection. \textbf{Bold} indicates the best result per column. \colorbox{gray!20}{Gray} cells indicate configurations that outperform the \colorbox{gray!40}{baseline} ResNet-50 \textsc{Conv2D} on that metric.  
         }
\normalsize
\setlength{\tabcolsep}{4pt}
\begin{tabular}{lcccc}
\toprule
Layer Type & Test Loss & Acc (\%) & GFLOPS & \# Params. \\
\midrule
\rowcolor{gray!40} ResNet-50, \textsc{Conv2D}                             & 1.170 & 78.86 & 8.178  & 25.583M \\
ResNet-50, \convnn\ (All, $k=9$)                       & 1.187 & 76.22 & 10.321 & 25.557M \\
\midrule
ResNet-50, Branching (All, $k=4$)                      & \cellcolor{gray!20} 1.119 & \cellcolor{gray!20} 79.06 & 10.655 & 24.191M \\
ResNet-50, Branching (All, $k=6$)                      & \cellcolor{gray!20} 1.066 & \cellcolor{gray!20} 79.19 & 10.655 & 25.104M \\
ResNet-50, Branching (All, $k=9$)                      & \cellcolor{gray!20} 1.045 & \cellcolor{gray!20} 79.42 & 10.655 & 26.474M \\
ResNet-50, Branching (All, $k=16$)                     & \cellcolor{gray!20} \textbf{0.999} & \cellcolor{gray!20} \textbf{79.76} & 10.655 & 29.671M \\
\midrule
ResNet-50 Branching (Rand, $k=4$, $r=32$)           & \cellcolor{gray!20} 1.093 & 78.73 & 5.596  & 24.191M \\
ResNet-50 Branching (Rand, $k=4$, $r=48$)           & \cellcolor{gray!20} 1.150 & \cellcolor{gray!20} 79.03 & 5.641  & 24.191M \\
ResNet-50 Branching (Rand, $k=4$, $r=64$)           & \cellcolor{gray!20} 1.147 & \cellcolor{gray!20} 79.06 & 5.686  & 24.191M \\

ResNet-50 Branching (Rand, $k=6$, $r=32$)           & \cellcolor{gray!20} 1.143 & 78.84 & 5.596  & 25.104M \\
ResNet-50 Branching (Rand, $k=6$, $r=48$)           & \cellcolor{gray!20} 1.136 & \cellcolor{gray!20} 78.99 & 5.641  & 25.104M \\
ResNet-50 Branching (Rand, $k=6$, $r=64$)           & \cellcolor{gray!20} 1.080 & \cellcolor{gray!20} 79.16 & 5.686  & 25.104M \\

ResNet-50 Branching (Rand, $k=9$, $r=32$)           & \cellcolor{gray!20} 1.142 & \cellcolor{gray!20} 78.98 & 5.596  & 26.474M \\
ResNet-50 Branching (Rand, $k=9$, $r=48$)           & 1.180 & \cellcolor{gray!20} 78.99 & 5.641  & 26.474M \\
ResNet-50 Branching (Rand, $k=9$, $r=64$)           & \cellcolor{gray!20} 1.123 & \cellcolor{gray!20} 79.24 & 5.686  & 26.474M \\

ResNet-50 Branching (Rand, $k=16$, $r=32$)          & \cellcolor{gray!20} 1.138 & \cellcolor{gray!20} 79.12 & 5.596  & 29.671M \\
ResNet-50 Branching (Rand, $k=16$, $r=48$)          & \cellcolor{gray!20} 1.088 & \cellcolor{gray!20} 79.28 & 5.641  & 29.671M \\
ResNet-50 Branching (Rand, $k=16$, $r=64$)          & \cellcolor{gray!20} 1.065 & \cellcolor{gray!20} 79.37 & 5.686  & 29.671M \\
\midrule[0.5pt]
ResNet-50 Branching (Spat, $k=9$, $r=25, r' = 5$)   & \cellcolor{gray!20} 1.082 & 78.82 & 5.576  & 26.474M \\
\bottomrule
\end{tabular}
\label{tab:imagenet_resnet_results}
\vspace{10pt}
\end{table}

\subsection{\convnn in ViT-Base}
\label{subsec:imagenet_vit}

\subsubsection{Experimental Setup}
We replace standard multi-head self-attention in a ViT-Base backbone with \convnn, using fully learned $Q$, $K$, $V$ projections, scaled dot-product similarity, and $\rho = \operatorname{softmax}$. We evaluate two head configurations: single-head ($\text{NH} = 1$) and multi-head ($\text{NH} = 12$, matching the ViT-Base default). We compare all-feature selection, random selection ($r = 32$), and spatial selection ($r = 128$), and include the Fast-\convnn Triton-accelerated kernel variant for large $k$ from Section~\ref{sec:fast_convnn}. The key distinction between \convnn and KVT-attention (\cite{DBLP:conf/eccv/WangWWLCLJ22}) lies in the aggregation kernel. Both are initialized with depthwise \textsc{Conv1D} weights to uniform values of 1, but \convnn allows these weights to be learned during training, while KVT fixes them at 1. This makes the aggregation kernel of \convnn a learned parameter, enabling the model to adaptively weight the contribution of each neighbor position. The standard multi-head self-attention baseline at 80.94\% (NH=12) serves as the primary reference throughout.

\subsubsection{Effect of Multi-Head Attention}

\begin{figure}[H]
\vspace{5pt}
\centering
\begin{tikzpicture}
\begin{axis}[
    width=0.9\linewidth, height=7cm,
    ybar=0pt, 
    bar width=25pt, 
    enlarge x limits=0.25,
    symbolic x coords={ConvNN, KVT, MHA baseline},
    xtick=data,
    xticklabel style={draw=none}, 
    ylabel={Top-1 Accuracy (\%)},
    ymin=74, ymax=83,
    ymajorgrids=true, grid style={dotted, gray!70},
    legend style={at={(0.98,0.02)}, anchor=south east, font=\footnotesize,
                  draw=gray!40, fill=white},
    tick label style={font=\small},
    label style={font=\small},
    nodes near coords,
    nodes near coords style={font=\footnotesize, /pgf/number format/fixed,
                              /pgf/number format/precision=2},
    every node near coord/.append style={yshift=2pt},
]

\addplot[fill=blue!50, draw=blue!80]
    coordinates {
        (ConvNN, 78.20)
        (KVT, 76.95)
        (MHA baseline, 79.35)
    };

\addplot[fill=orange!50, draw=orange!80] 
    coordinates {
        (ConvNN, 81.19)
        (KVT, 80.50)
        (MHA baseline, 80.94)
    };

\legend{NH=1, NH=12}

\end{axis}
\end{tikzpicture}
\caption{Effect of head count on ViT-Base top-1 accuracy for
         \convnn and KVT-attention at $k = 9$,
         compared with the MHA baseline.}
\label{fig:imagenet_nh}
\end{figure}
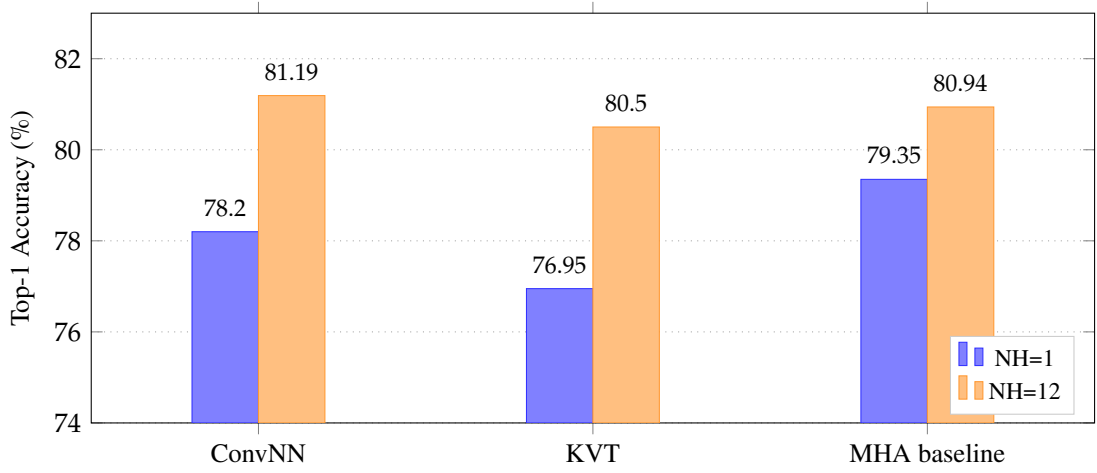

Figure~\ref{fig:imagenet_nh} illustrates the large gap between $\text{NH} = 1$ and $\text{NH} = 12$ for both \convnn and KVT. At $k = 9$, single-head \convnn achieves 78.20\% while the multi-head variant reaches 81.19\%, a gap of nearly 3\%. An analogous gap is observed for KVT (76.95\% vs. 80.50\%). This result confirms that the head-splitting strategy described in Section~\ref{subsec:multihead_convnn} is essential for competitive performance by partitioning the feature space across $H$ independent subspaces allowing each head to specialize in distinct aspect of the input, providing a richer representational capacity than a single global similarity computation. 

Notably, the parameter count of \convnn \emph{decreases} when moving from $\text{NH} = 1$ to $\text{NH} = 12$ (86.462M vs. 86.386M). This occurs because the depthwise \textsc{Conv1D} aggregation kernel operates on the per-head channel dimension $d_k / H = 64$ rather than the full $d_k = 768$ when $\text{NH} = 1$, reducing the number of learnable aggregation weights by a factor of $H$. 

\subsubsection{Effect of neighbor count $k$}

\begin{figure}[H]
\centering
\begin{tikzpicture}
\begin{axis}[
    width=0.85\linewidth, height=6cm,
    xlabel={Neighbor count $k$},
    ylabel={Top-1 Accuracy (\%)},
    xtick={9,16,25,36,100},
    xticklabels={9,16,25,36,100},
    ymin=80, ymax=82.5,
    ymajorgrids=true, grid style={dotted, gray!40},
    legend style={at={(0.98,0.02)}, anchor=south east, font=\footnotesize,
                  draw=gray!40, fill=white, row sep=1pt},
    tick label style={font=\small},
    label style={font=\small},
]
\addplot[color=blue!70, mark=*, thick, mark size=2pt]
    coordinates {(9,81.19) (16,81.60) (25,81.64) (36,81.61)};
\addlegendentry{\convnn (NH=12, All)}

\addplot[color=teal!80, mark=triangle*, thick, dashed, mark size=2.5pt]
    coordinates {(9,80.50) (16,81.04) (25,81.28) (36,81.03) (100,81.04)};
\addlegendentry{KVT (NH=12)}

\addplot[color=gray!90, thick, dotted, mark=none]
    coordinates {(9,80.94) (16,80.94) (25,80.94) (36,80.94) (100,80.94)};
\addlegendentry{MHA baseline}
\end{axis}
\end{tikzpicture}
\caption{Top-1 accuracy on ImageNet-1K for ViT-Base comparing
         \convnn and KVT-attention (both
         NH=12) across neighbor counts $k$, against the MHA baseline. 
         Results at $k = 25$ and $k = 36$ use the Fast-\convnn\ implementation with a custom Triton kernel from Section~\ref{sec:fast_convnn}.}
\label{fig:imagenet_vit_k}
\end{figure}
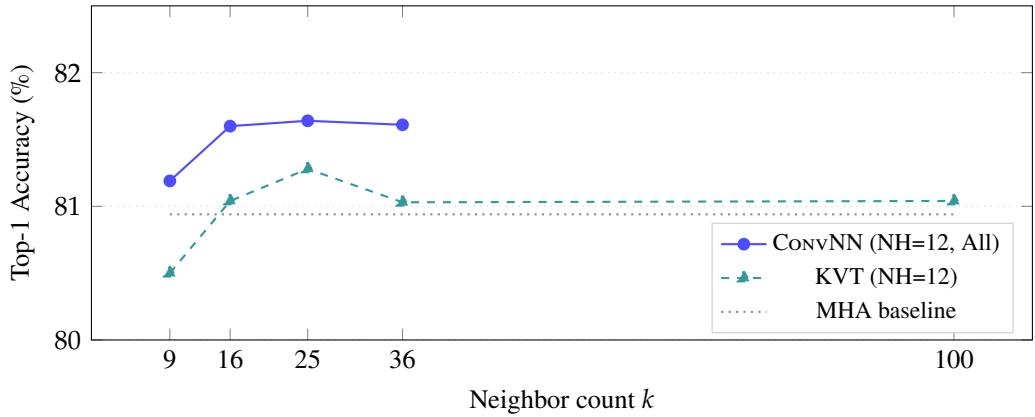

Figure~\ref{fig:imagenet_vit_k} compares \convnn and KVT-attention (both $\text{NH} = 12$) with the multi-head self-attention baseline across varying $k$. \convnn consistently outperforms KVT at matched $k$ values. At $k = 9$, \convnn achieves 81.19\% vs. KVT's 80.50\%, and at $k = 16$, 81.60\% vs. 81.04\%. The Fast-\convnn Triton variant further pushes accuracy to 81.64\% at $k = 25$, surpassing the multi-head self-attention baseline by 0.70\%. KVT accuracy peaks around $k = 25$ (81.28\%) and diminishes for larger $k$, while \convnn continues to improve up to $k = 25$ before stabilizing, suggesting that the learned aggregation kernel provides a more effective mechanism for exploiting larger neighborhoods than the uniform-weight sum used in KVT. 

\subsubsection{Effect of Sparse Candidate Search}
\begin{figure}[H]
\centering
\begin{tikzpicture}

\begin{axis}[
    name=ax1,
    width=0.48\linewidth, height=7.0cm,
    ybar, bar width=14pt,
    symbolic x coords={
        {All features},
        {Random ($r$=32)},
        {Spatial ($r$=128)},
        {MHA baseline}},
    xtick=data,
    xtick style={draw=none},
    xticklabel style={font=\small, rotate=25, anchor=east},
    ylabel={Top-1 Accuracy (\%)},
    ymin=73, ymax=84,
    ymajorgrids=true, grid style={dotted, gray!40},
    legend style={at={(0.02,0.98)}, anchor=north west, font=\small,
                  draw=gray!40, fill=white, row sep=1pt},
    tick label style={font=\small},
    label style={font=\small},
    nodes near coords,
    nodes near coords style={font=\footnotesize,
                              /pgf/number format/fixed,
                              /pgf/number format/precision=2},
    every node near coord/.append style={yshift=2pt},
]
\addplot[fill=blue!50, draw=blue!80]
    coordinates {
        ({All features},       81.19)
        ({Random ($r$=32)},    76.04)
        ({Spatial ($r$=128)},  79.31)
        ({MHA baseline},       80.94)
    };
\addlegendentry{Acc (\%)}
\end{axis}

\begin{axis}[
    name=ax2,
    at={(ax1.outer east)}, 
    anchor=outer west,
    width=0.48\linewidth, height=7.0cm,
    ybar, bar width=14pt,
    symbolic x coords={
        {All features},
        {Random ($r$=32)},
        {Spatial ($r$=128)},
        {MHA baseline}},
    xtick=data,
    xtick style={draw=none},
    xticklabel style={font=\small, rotate=25, anchor=east},
    ylabel={GFLOPS},
    ymin=30, ymax=37,
    ymajorgrids=true, grid style={dotted, gray!40},
    ylabel style={color=orange!80!black},
    yticklabel style={color=orange!80!black, font=\small},
    legend style={at={(0.98,0.98)}, anchor=north east, font=\small,
                  draw=gray!40, fill=white, row sep=1pt},
    nodes near coords,
    nodes near coords style={font=\footnotesize, color=orange!80!black,
                              /pgf/number format/fixed,
                              /pgf/number format/precision=2},
    every node near coord/.append style={yshift=2pt},
]
\addplot[fill=orange!40, draw=orange!80!black]
    coordinates {
        ({All features},       34.43)
        ({Random ($r$=32)},    33.83)
        ({Spatial ($r$=128)},  34.18)
        ({MHA baseline},       35.13)
    };
\addlegendentry{GFLOPS}
\end{axis}

\end{tikzpicture}
\caption{Top-1 accuracy and GFLOPS for ViT-Base \convnn (NH=12, $k=9$) under different candidate selection strategies, compared with the MHA baseline.}
\label{fig:imagenet_vit_sparse}
\vspace{5pt}
\end{figure}
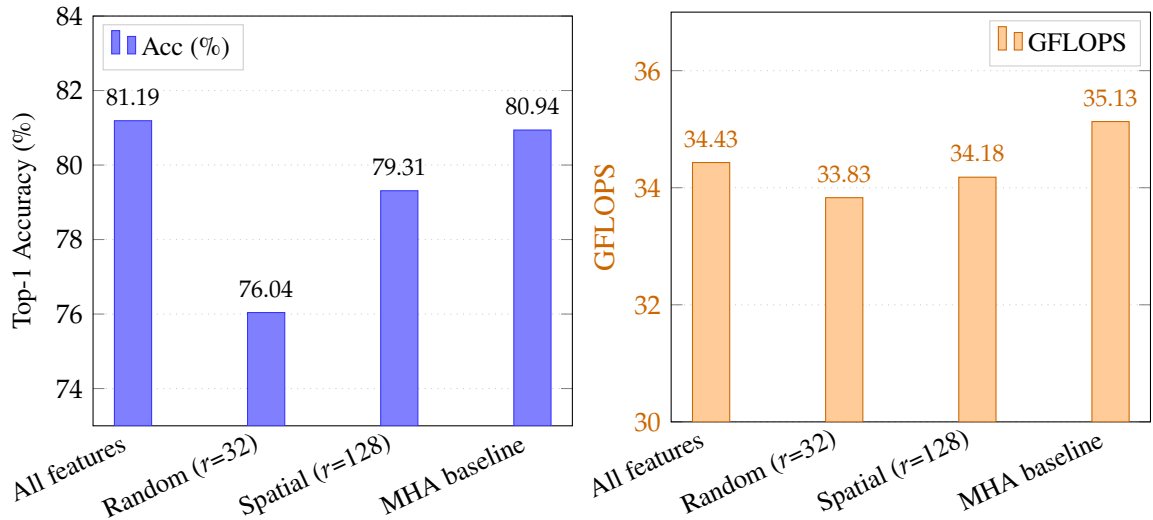

Figure~\ref{fig:imagenet_vit_sparse} compares candidate selection strategies at $\text{NH} = 12$, $k = 9$. All-feature \convnn achieves the best accuracy at 81.19\% with 34.43 GFLOPS, slightly below the multi-head self-attention baseline's 35.13 GFLOPS. Spatial selection ($r = 128$) reduces GFLOPS to 34.18 while achieving 79.31\%, below the MHA baseline. Random selection ($r = 32$) achieves the lowest GFLOPS (33.83) but also the lowest accuracy (76.04\%), a larger drop than observed in the CIFAR experiments. This suggests that at the ImageNet scale, the quality of the neighbor pool matters more than its diversity. All-feature selection, which has access to the full global context, consistently provides the best accuracy, while aggressive random subsampling loses too many informative neighbors to maintain competitive performance on a 1000-class task. 

\subsubsection{Training Dynamics and Late-Stage Convergence}

\begin{table}[H]
\centering
\small
\vspace{4pt}
\setlength{\tabcolsep}{4pt}
\caption{Top-1 accuracy on ImageNet-1K at 50-epoch intervals for ViT-Base with NH=12. \convnn use all-feature candidate selection.}
\label{tab:imagenet_vit_convergence}
\begin{tabular}{lcccccc}
\toprule
Layer Type & Ep. 50 & Ep. 100 & Ep. 150 & Ep. 200 & Ep. 250 & Ep. 300 \\
\midrule
MHA (NH=12)                              & 71.70 & 76.72 & 78.69 &  79.96 & 80.95 &  80.94 \\
\midrule[0.2pt]
\convnn (NH=12, $k=9$)        & 67.57 & 73.61 &  76.38 & 78.63 & 80.74 & 81.19    \\   
Fast-\convnn (NH=12, $k=25$)  & 69.83 & 74.91 & 77.79 &  80.13 &81.26 & 81.64 \\
\midrule[0.2pt]
KVT (NH=12, $k=9$)                      & 65.25 & 71.31 & 75.04 & 78.13 & 79.84 & 80.50 \\     
KVT (NH=12, $k=25$)                     & 69.79 & 74.75 & 77.71 & 79.80 & 80.81 & 81.28 \\
\bottomrule
\end{tabular}
\vspace{5pt}
\end{table}

Table~\ref{tab:imagenet_vit_convergence} reports top-1 accuracy at 50-epoch intervals across 300 epochs of training. A clear pattern emerges for \convnn. It lags behind the MHA baseline in the early and middle stages of training, but closes the gap and surpasses it in the final epochs. At epoch 50, \convnn ($k=9$) trails MHA by 4.13\% (67.57\% vs. 71.70\%), and this gap remains substantial through epoch 200 (78.63\% vs. 79.96\%). Only in the final 100 epochs does \convnn overtake MHA, finishing at 81.19\% vs. 80.94\%. The Fast-\convnn\ Triton variant ($k=25$) follows the same pattern and achieves the higest final accuracy of 81.64\%. 

This behavior is explained by the role of the depthwise \textsc{Conv1D} aggregation kernel. The kernel weights are initialized to 1, matching the uniform-weight KVT formulation, but must be learned from scratch during training. In the early stages, when the learning rate is high and gradient updates are large, the kernel weights undergo exploratory updates that may not yet provide a stable aggregation signal, placing \convnn at a disadvantage relative to MHA, whose softmax weighted sum requires no additional learned weights. As the cosine annealing schedule reduces the learning rate toward its minimum in the final epochs, the fine-grained weight updates characteristic of low learning rates allow the aggregation kernel to converge to a precise per-channel weighting that standard uniform aggregation cannot express. This late-stage advantage is consistent with the learning rate sensitivity analysis on CIFAR in Figure~\ref{fig:vgg_lr_ablation}, where \convnn\ outperforms the \textsc{Conv2D} baseline at low learning rates, suggesting that the learned aggregation weights benefit specifically from the precision of small gradient updates. 

KVT exhibits a similar early deficit but does not recover as fully as \convnn, peaking at 80.50\% ($k=9$) and 81.28\% ($k=25$). The gap between KVT and \convnn at the final epoch confirms that the learnable aggregation kernel is the primary driver of \convnn's advantage. Given the same neighbor selection strategy as KVT, \convnn's ability to learn non-uniform aggregation weights across $K$ positions provides a consistent accuracy gain. 

\subsubsection{Full Results}
\begin{table}[H]
\centering
\normalsize
\vspace{4pt}
\caption{ViT-Base on ImageNet-1K top-1 classification.
         Rand = random candidate selection, Spat = spatial candidate
         selection, NH = number of attention heads. \textbf{Bold} indicates the best result per column. \colorbox{gray!20}{Gray} cells indicate configurations that
         outperform the \colorbox{gray!40}{baseline} ViT-Base \textsc{MHA} NH = 12 on that metric.
         }
\setlength{\tabcolsep}{4pt}
\begin{tabular}{lcccc}
\toprule
Layer Type & Test Loss & Acc (\%) & GFLOPS & \# Params. \\
\midrule
ViT Base MHA (NH=1)  & 0.913 & 79.35 & 35.131 & 86.379M \\
\rowcolor{gray!40} ViT Base MHA (NH=12) & 0.899 & 80.94 & 35.131 & 86.379M \\
\midrule[0.2pt]
ViT Base \convnn (NH=1, All, $k=4$)            & 1.131 & 74.60 & 34.423 & 86.416M \\
ViT Base \convnn (NH=1, All, $k=9$)            & 0.969 & 78.20 & 34.432 & 86.462M \\
ViT Base \convnn (NH=1, All, $k=16$)           & 0.937 & 79.14 & 34.445 & 86.526M \\
ViT Base \convnn (NH=1, Rand, $k=9$, $r=32$)   & 1.262 & 71.60 & 31.497 & 86.462M \\
ViT Base \convnn (NH=1, Spat, $k=9$, $r=128$)  & 1.076 & 75.86 & 33.205 & 86.462M \\
\midrule[0.2pt]
ViT Base \convnn (NH=12, All, $k=9$)           & \cellcolor{gray!20} 0.855 & \cellcolor{gray!20} 81.19 & 34.432 & 86.386M \\
ViT Base \convnn (NH=12, All, $k=16$)          & \cellcolor{gray!20} 0.838 & \cellcolor{gray!20} 81.60 & 34.445 & 86.391M \\
ViT Base Fast-\convnn (NH=12, All, $k=25$)   & \cellcolor{gray!20} \textbf{0.834} & \cellcolor{gray!20} \textbf{81.64} & 34.416 & 86.398M \\
ViT Base Fast-\convnn (NH=12, All, $k=36$)   & \cellcolor{gray!20} 0.845 & \cellcolor{gray!20} 81.61 & 34.416 & 86.407M \\
\midrule[0.2pt]

ViT Base \convnn (NH=12, Rand, $k=9$, $r=32$)  & 1.064 & 76.04 & 33.833 & 86.386M \\
ViT Base \convnn (NH=12, Spat, $k=9$, $r=128$) & 0.931 & 79.31 & 34.181 & 86.386M \\
\midrule[0.2pt]
ViT Base KVT (NH=1, $k=4$)                               & 1.182 & 73.38 & 35.131 & 86.379M \\
ViT Base KVT (NH=1, $k=9$)                               & 1.020 & 76.95 & 35.131 & 86.379M \\
ViT Base KVT (NH=1, $k=16$)                              & 0.972 & 78.25 & 35.131 & 86.379M \\
\midrule[0.2pt]
ViT Base KVT (NH=12, $k=9$)                              & \cellcolor{gray!20} 0.883 & 80.50 & 35.137 & 86.379M \\
ViT Base KVT (NH=12, $k=16$)                             & \cellcolor{gray!20} 0.871 & \cellcolor{gray!20} 81.04 & 35.137 & 86.379M \\
ViT Base KVT (NH=12, $k=25$)                             & \cellcolor{gray!20} 0.874 & \cellcolor{gray!20} 81.28 & 35.137 & 86.379M \\
ViT Base KVT (NH=12, $k=36$)                             & \cellcolor{gray!20} 0.880 & \cellcolor{gray!20} 81.03 & 35.137 & 86.379M \\
ViT Base KVT (NH=12, $k=100$)                            & \cellcolor{gray!20} 0.884 & \cellcolor{gray!20} 81.04 & 35.137 & 86.379M \\
\bottomrule
\end{tabular}
\label{tab:imagnet_vit_results}
\end{table}

\section{Conclusion of Large Scale Validation}
The ImageNet-1K experiments confirm that the findings established in the CIFAR proof-of-concept extend to large-scale classification, and introduce several new observations specific to the full-scale setting. 

On the convolutional side, hybrid branching with \convnn\ consistently outperforms the \textsc{Conv2D} baseline across all values of $k$ when using all-feature selection, with accuracy increasingly monotonically from 79.06\% at $k = 4$ to 79.76\% at $k = 16$, a 0.90\% gain over the 78.86\% baseline. As in the CIFAR experiments, the pure \convnn\ replacement underperforms both the hybrid branching and the \textsc{Conv2D} baseline (76.22\%), reinforcing that the combination of local and global receptive fields is essential for strong convolutional performance. Sparse random selection provides an attractive efficiency tradeoff with the configuration with $k = 4$ and $r = 48$ surpassing the \textsc{Conv2D} baseline in accuracy (79.03\% vs. 78.86\%) while using 1.4M less parameters and 31\% fewer GFLOPS, demonstrating that \convnn\ branching can match or exceed baseline performance at reduced computational cost. 

On the attention side, the ImageNet-1K experiments reveal that multi-head splitting is essential for competitive performance at this scale. Moving from $\text{NH} = 1$ to $\text{NH} = 12$ yields a nearly 3\% accuracy gain for \convnn (78.20\% to 81.19\% at $k = 9$) and an analogous improvement for KVT (76.95\% to 80.50\%), while paradoxically reducing the parameter count due to the per-head channel dimension of the depthwise aggregation kernel. With $\text{NH} = 12$, \convnn outperforms KVT-attention at every matched value of $k$, and the Fast-\convnn\ Triton variant achieves the best overall result of 81.64\% at $k = 25$, surpassing the MHA baseline by 0.70\%. The key differentiator from KVT is the learned aggregation kernel, which allows \convnn to adaptively weight neighbor contributions rather than treating all positions uniformly. KVT accuracy peaks at $k = 25$ and degrades for larger $k$, while \convnn continues to benefit from larger neighborhoods, suggesting that learnable aggregation is critical for exploiting dense global context at ImageNet scale. 

A notable divergence from the CIFAR findings concerns spatial candidate selection in the attention setting. Whereas random selection was the best-performing strategy on CIFAR, it severely underperforms on ImageNet-1K (76.04\% at $r = 32$, NH=12), while all-feature selection achieves the best accuracy. This reversal suggests a scale-dependent tradeoff: on small datasets with few classes, candidate diversity provides a regularization benefit that outweighs any loss in neighbor quality, but on a 1000-class task, the quality and coverage of the neighbor pool becomes the dominant factor, and aggressive subsampling discards too many discriminative features. 

The training dynamics analysis further reveals the \convnn's advantage over MHA and KVT is concentrated in the late stages of training, when the cosine annealing schedule reduces the learning rate to its minimum and fine-grained weight updates allow the depthwise aggregation kernel to converge to a precise per-channel weighting. This late-stage behavior, consistent with the learning rate sensitivity analysis on CIFAR, suggests that the learned aggregation kernel is not merely an architectural choice but a mechanism that specifically benefits from low learning rate fine-tuning, a property not shared by MHA or KVT's fixed uniform-weight aggregation. 

Together, these results establish \convnn\ as a principled and scalable framework that provides consistent improvements over both convolutional and attention baselines at ImageNet scale, with the appropriate configuration depending on the computational budget and architectural context. 

%% file: Chapters/Chapter6.tex

\chapter{Conclusion} 

\label{Chapter6} 

\section{Summary}
Our framework reveal both the promise and limitations of unifying convolution and attention through $k$-NN. The \convnn framework successfully demonstrates convolution and self-attention exist along a spectrum of neighbor selection strategies, with convolution representing a fixed spatial selection and attention representing global, feature-based selection. By interpolating between the two ends of the spectrum, there exists multitudes of novel operations that can be found to which we have shown with explicit results that show the benefit of interpolating between the two operations. By making convolutions more like attention or attention more like convolutions, the different combinations that can be used for the \convnn framework is vast. 

A key finding is the strong performance of all neighbor candidates in \convnn. This highlights that when employing a global receptive field, gathering the neighbor from a pool of entirety of candidates is more effective than sparsely selecting them from a subset of candidates. Experiments by varying the number of samples in random and spatial selection show that the more candidates that \convnn can pick from, the better. In certain situations like the hybrid branching layer~\ref{subsec:hybrid_branching_layer}, the candidate pool does not matter as much as it is in conjunction with the regular convolution operations that counterbalances the sparse global selection shown in the random and spatial subsampling of candidates.

Ablation studies on $k$ values reveal an important distinction between convolution and attention. While larger convolutional kernels consistently improve performance (e.g., kernel size $1 \rightarrow 2$ or $3$), the gains from increasing $k$ in \convnn are modest, eventually halting performance gains after a certain increase. 

We also observe that \convnn performs significantly better when the multi-head operation is used compared to single head. This highlights the parallelizability of the \convnn and potentially can be used in convolution-style operations where it convolves in different channels in parallel and combines with a output weights. 

The training dynamics experiments further reveal that \convnn exhibits a distinctive late-stage convergence behavior, lagging behind the standard multi-head attention in early and middle training epochs but surpassing it in the final stages. This emphasizes that the learned aggregation weights in \convnn encode a form of neighbor-channel importance that requires precise, low-magnitude gradient updates to develop, and points to a broader principle that the benefit of learnable aggregation over fixed uniform weighting is most visible under low learning rate fine-tuning conditions. 

Furthermore, we observe regularization effects, particularly in convolutional architectures. By adding the \convnn branch in the hybrid branching layer, adding the global receptive fields through \convnn acts as an implicit regularizer, forcing the network to balance spatial and feature-based aggregation. This balance between local and global prevents over-reliance on either mechanism alone, leading to a more robust feature learning and information extraction. 

\section{Limitations and Future Improvements}
Our work also highlights several limitations. The hard selection mechanism in \convnn limits gradient flow and may cause training instability in certain settings. While the Fast-\convnn makes the framework more computationally efficient, the cost of $k$-NN search remains higher than standard convolution, though lower than full self-attention. With the reordering of high-dimensional tensors combined with the $k$-NN operation, \convnn produces a higher usage in memory and takes longer for the models to train. As $K$ increases to larger values, the memory required to store the combined neighbors grow linearly and cannot realistically utilize the $K = N$ approach described in Section~\ref{sec:convnn_attn_connection}. \convnn's operational framework is useful in low-to-middle ranges of $K$ and must be careful in determining the cost and benefit of memory, speed, and performance. Extension to tasks requiring dense predictions (e.g., segmentation) may require careful consideration of how to maintain spatial correspondence during neighbor selection.

%% file: Chapters/Chapter7.tex

\chapter{Disclosure of Funding and Codebase} 

\label{Chapter7} 

\label{sec:funding}

\section{Funding Acknowledgments}
This work was supported by the following grants and fellowships: Bowdoin College Christenfeld Summer Research Fellowship (2024), Last Mile Education Google AI Award (2024), Last Mile Education NYC STEM Award (2024), John L. Roberts Fund Fall Research Grant (2025), and Bowdoin College Hastings Student AI Grant (2026). 

Large-scale experiments were conducted on the Indiana Jetstream2 (\cite{10.1145/3437359.3465565}) through allocation SEE240007 from the Advanced Cyberinfrastructure Coordination Ecosystem: Service \& Support (ACCESS) program (\cite{Boerneretal2023}), which is supported by National science Foundation grants \#2138259, \#2138286, \#2138307, \#2137603, and \#2138296.

\section{Code Availability}
All code developed from this thesis is publicly available. The repositories are organized by architectural setting. 

\begin{itemize}
    \item \textbf{CNN experiments} (VGG-11, ResNet-50 with \convnn\ branching): \\ \href{https://github.com/mingikang31/Convolutional-Nearest-Neighbor}{github.com/mingikang31/Convolutional-Nearest-Neighbor}
    \item \textbf{Vision Transformer experiments} (ViT-Tiny, ViT-Base with \convnn-Attention): \\ \href{https://github.com/mingikang31/Convolutional-Nearest-Neighbor-Attention}{github.com/mingikang31/Convolutional-Nearest-Neighbor-Attention}
    
\end{itemize}